\documentclass{article}

\PassOptionsToPackage{numbers, sort&compress}{natbib}
\usepackage[preprint]{neurips_2021}
\usepackage{times}

\usepackage[utf8]{inputenc} %
\usepackage[T1]{fontenc}    %

\usepackage{amsmath, amsthm, amssymb,amsfonts}
\usepackage{textcomp}
\usepackage{stmaryrd}
\SetSymbolFont{stmry}{bold}{U}{stmry}{m}{n}
\usepackage{upgreek}
\usepackage{bm}
\usepackage{cases}
\usepackage{mathtools}
\usepackage{arydshln}

\usepackage{epsfig}
\usepackage{graphicx}
\usepackage{wrapfig}
\usepackage[percent]{overpic}
\usepackage[belowskip=1pt,aboveskip=5pt,font=small]{caption}
\usepackage[belowskip=0pt,aboveskip=3pt,font=small]{subcaption}
\usepackage{epigraph}
\usepackage{multirow}
\usepackage{rotating}
\usepackage{booktabs}
\usepackage{makecell}

\usepackage{enumitem}
\usepackage[olditem,oldenum]{paralist}

\newcommand{\myquote}[1]{\emph{`#1'}}
\newcommand{\myapprox}{{\raise.17ex\hbox{$\scriptstyle\sim$}}}
\newcommand{\xhdr}[1]{\vspace{0pt}\noindent\textbf{#1}\xspace}

\newcommand{\reffig}[1]{Fig.~\ref{#1}}
\newcommand{\refsec}[1]{Sec.~\ref{#1}}
\newcommand{\reftab}[1]{Tab.~\ref{#1}}

\makeatletter
\newcommand\footnoteref[1]{\protected@xdef\@thefnmark{\ref{#1}}\@footnotemark}
\makeatother

\DeclareMathSymbol{@}{\mathord}{letters}{"3B}

\newcommand{\habnamefull}{Habitat 2.0\xspace}
\newcommand{\habname}{H2.0\xspace}
\newcommand{\replica}{ReplicaCAD\xspace}
\newcommand{\tasksuitenamefull}{Home Assistant Benchmark\xspace}
\newcommand{\tasksuitename}{HAB\xspace}

\newcommand{\pick}{\texttt{\footnotesize Pick}\xspace}
\newcommand{\place}{\texttt{\footnotesize Place}\xspace}

\newcommand{\opendoor}{\texttt{\footnotesize Open \hspace{-8pt} fridge\xspace}}
\newcommand{\closedoor}{\texttt{\footnotesize Close \hspace{-8pt} fridge\xspace}}
\newcommand{\opendrawer}{\texttt{\footnotesize Open \hspace{-8pt} drawer}\xspace}
\newcommand{\closedrawer}{\texttt{\footnotesize Close \hspace{-8pt} drawer}\xspace}
\newcommand{\navigate}{\texttt{\footnotesize Navigate}\xspace}

\newcommand{\stockfridge}{\texttt{\footnotesize Prepare \hspace{-10pt} Groceries}\xspace}
\newcommand{\settable}{\texttt{\footnotesize Set \hspace{-10pt} Table}\xspace}

\newcommand{\cleanhouse}{\texttt{\footnotesize Tidy \hspace{-10pt} House}\xspace}

\newcommand{\learnedee}{\textbf{Learned (Mono)}\xspace}

\newcommand{\mpvis}{\textbf{SPA}\xspace}

\newcommand{\monolithic}{\textbf{MonolithicRL}\xspace}

\newcommand{\SPAfull}{\textbf{SensePlanAct}\xspace}
\newcommand{\SPAoraclefull}{\textbf{SensePlanAct-Priviledged}\xspace}
\newcommand{\SPA}{\textbf{SPA}\xspace}
\newcommand{\SPAoracle}{\textbf{SPA-Priv}\xspace}

\newcommand{\TPSfull}{\textbf{TaskPlanning+SkillsRL}\xspace}
\newcommand{\TPS}{\textbf{TP+SRL}\xspace}
\newcommand{\TPSPAfull}{\textbf{TaskPlanning+SensePlanAct}\xspace}
\newcommand{\TPSPA}{\textbf{TP+SPA}\xspace}

\newcommand{\TPSPAoracle}{\textbf{TP+SPA-Priv}\xspace}

\DeclareMathSymbol{@}{\mathord}{letters}{"3B}

\usepackage{mysymbols}

\usepackage{url}
\usepackage{xspace}
\usepackage{comment}
\usepackage{color}
\usepackage[table]{xcolor}
\usepackage{framed}
\usepackage{fancybox}
\usepackage{silence}
\usepackage{paralist}
\usepackage{nicefrac}

\usepackage{microtype}      %

\usepackage{footnote}
\makesavenoteenv{tabular}

\DeclareCaptionLabelFormat{andtable}%
{#1~#2 \& \tablename~\thetable}

\setdefaultleftmargin{.2cm}{.2cm}{}{}{}{}
\setlist{leftmargin=.2cm}

\definecolor{citecolor}{HTML}{2779af}
\definecolor{linkcolor}{HTML}{c0392b}
\usepackage[pagebackref=false,breaklinks=true,colorlinks,bookmarks=false,citecolor=citecolor,linkcolor=linkcolor]{hyperref}
\usepackage{cleveref}

\newcommand{\zk}[1]{}

\usepackage{etoolbox}

\newif\ifarxiv
\arxivtrue

\newtoggle{arxiv}
\ifarxiv
\toggletrue{arxiv}
\else
\togglefalse{arxiv}
\fi

\title{\habnamefull:\\ Training Home Assistants to Rearrange their Habitat}

\author{%
Andrew Szot$^{2}$, 
Alex Clegg$^{1}$, 
Eric Undersander$^{1}$, 
Erik Wijmans$^{1,2}$, 
Yili Zhao$^{1}$, 
John Turner$^{1}$,\\
\textbf{Noah Maestre$^{1}$, 
Mustafa Mukadam$^{1}$, 
Devendra Chaplot$^{1}$, 
Oleksandr Maksymets$^{1}$}, \\
\textbf{Aaron Gokaslan$^{1}$, 
Vladimir Vondrus,
Sameer Dharur$^{2}$,
Franziska Meier$^{1}$, 
Wojciech Galuba$^{1}$},\\
\textbf{Angel Chang$^{4}$,
Zsolt Kira$^{2}$, 
Vladlen Koltun$^{3}$, 
Jitendra Malik$^{1,5}$, 
Manolis Savva$^{4}$, 
Dhruv Batra$^{1,2}$}\\
$^{1}$Facebook AI Research, $^{2}$Georgia Tech, $^{3}$Intel Research, $^{4}$Simon Fraser University $^{5}$UC Berkeley
}

\begin{document}

\maketitle

\vspace{-5pt}
\begin{abstract}
\vspace{-5pt}
We introduce \habnamefull (\habname), a simulation platform for training %
virtual robots in \textit{interactive} 3D environments and complex physics-enabled scenarios.
We make comprehensive contributions to all levels of the embodied AI stack -- data, simulation, and benchmark tasks. 
Specifically, we present: 
(i) \replica: an artist-authored, annotated, reconfigurable 3D dataset of apartments (matching real spaces) with articulated objects (\eg cabinets and drawers that can open/close); 
(ii) \habname: a high-performance physics-enabled 3D simulator with \textbf{speeds exceeding 25,000 simulation steps per second (850$\times$ real-time)} 
on an 8-GPU node, 
representing $100\times$ speed-ups over prior work; and,  
(iii) \tasksuitenamefull (\tasksuitename): a suite of common tasks for assistive robots 
(tidy the house, prepare groceries, set the table) %
that test a range of mobile manipulation capabilities. 
These large-scale engineering contributions allow us to %
systematically compare 
deep reinforcement learning (RL) at scale and classical sense-plan-act (SPA) pipelines in long-horizon structured tasks, 
with an emphasis on generalization to new objects, receptacles, and layouts. 
We find that 
(1) flat RL policies struggle on \tasksuitename  compared to hierarchical ones;
(2) a hierarchy with independent skills suffers from `hand-off problems', and 
(3) SPA pipelines are more brittle than RL policies.

\end{abstract}

\begin{center}
    \centering
    \includegraphics[width=0.497\textwidth]{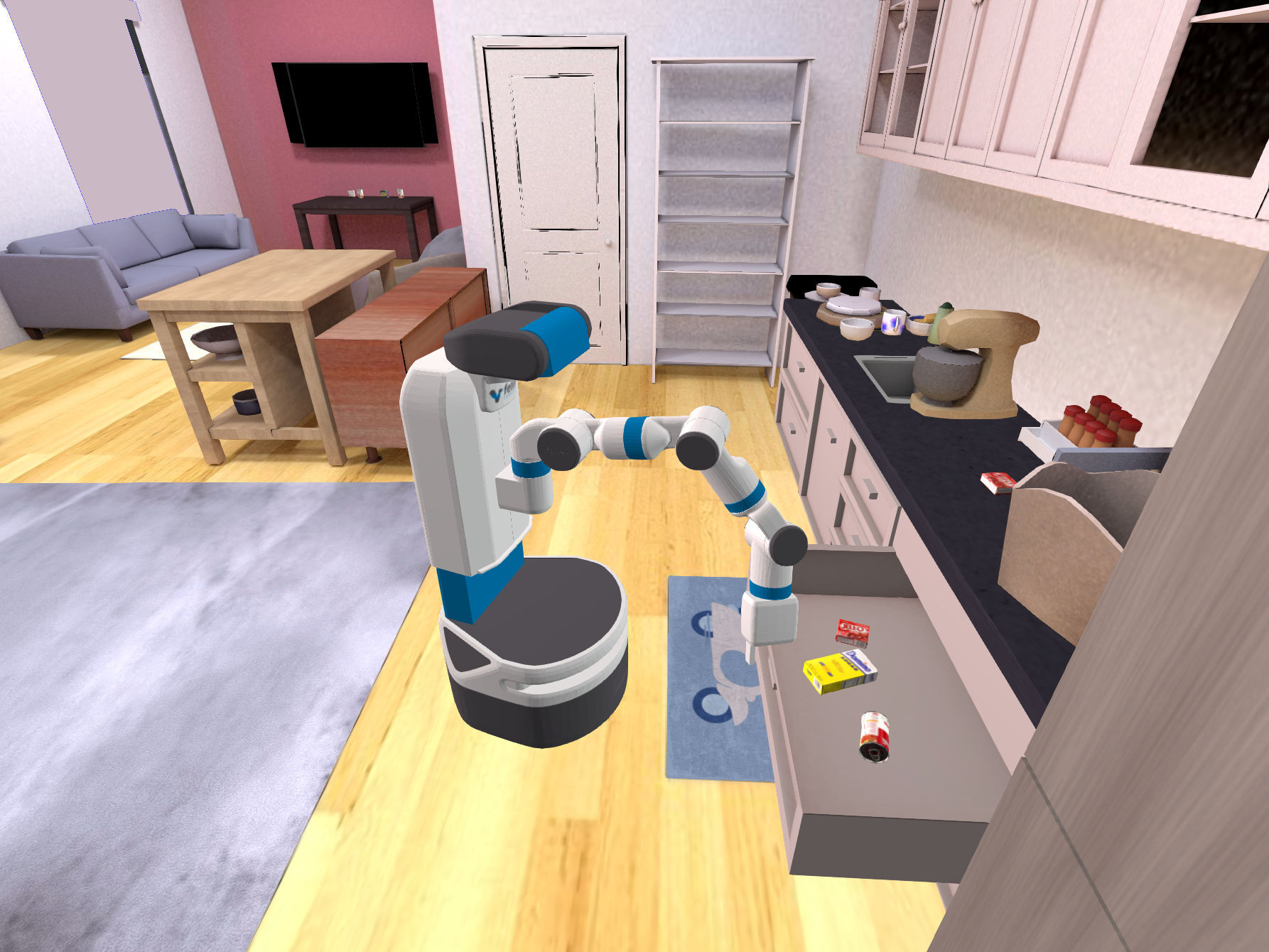}
    \includegraphics[width=0.497\textwidth]{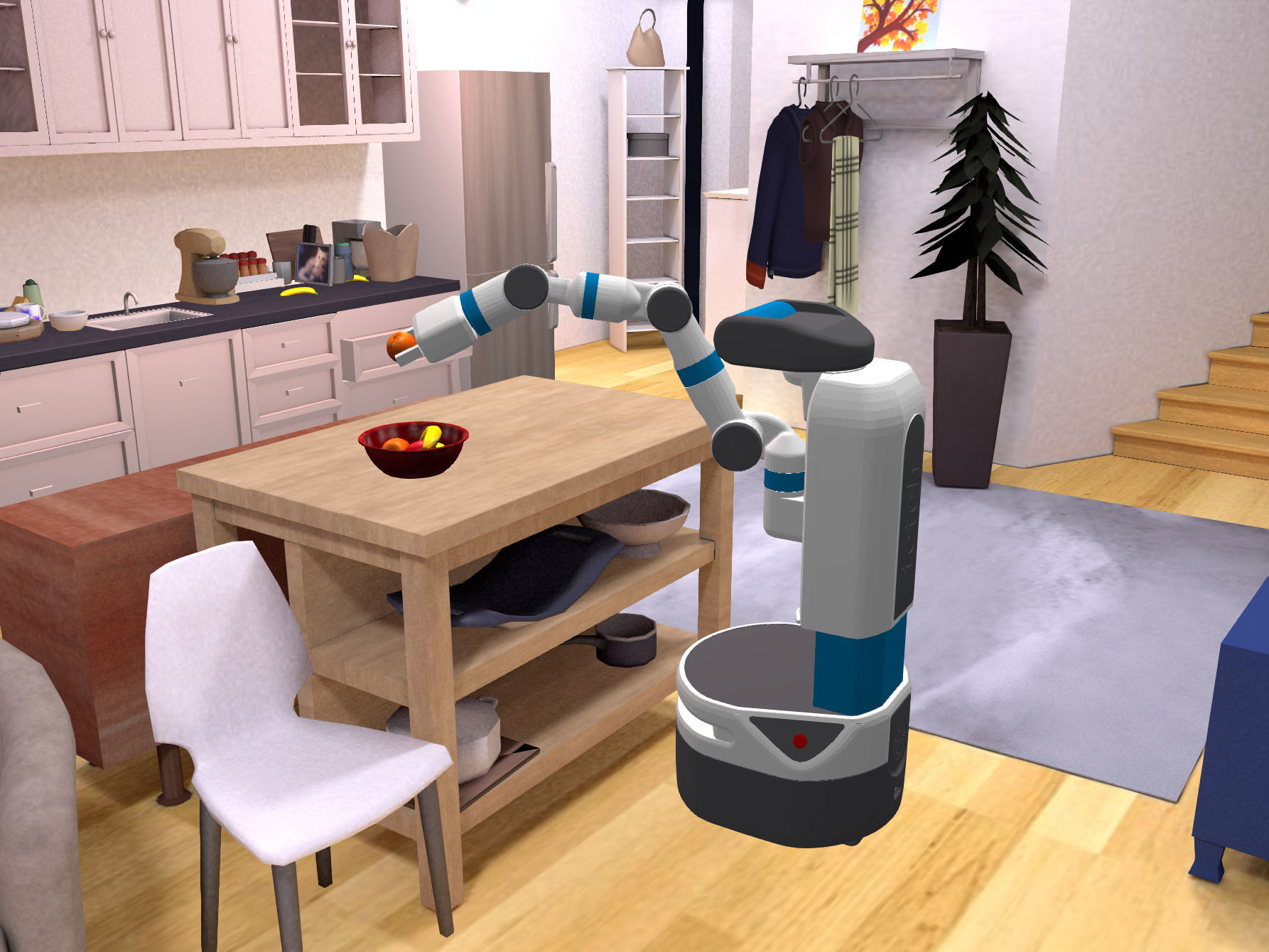}
    \captionof{figure}{
    A mobile manipulator (Fetch robot) simulated in \habnamefull performing rearrangement tasks in a \replica apartment 
    -- (left) opening a drawer before picking up an item from it, 
    and (right) placing an object into the bowl after navigating to the table. 
    Best viewed in motion at \href{https://sites.google.com/view/habitat2}{https://sites.google.com/view/habitat2}. 
    }
    \label{fig:teaser}
\end{center}

\vspace{-5pt}
\section{Introduction}
\vspace{-5pt}

Consider a home assistant robot illustrated in \figref{fig:teaser} -- a mobile manipulator (Fetch~\cite{fetchrobot}) performing tasks like stocking groceries into the fridge, clearing the table and putting dishes into the dishwasher, fetching objects on command and putting them back, \etc.
Developing such embodied intelligent systems is a goal of deep scientific and societal value.
So how should we accomplish this goal?

Training and testing such robots in hardware directly is slow, expensive, and difficult to reproduce.
We aim to advance the entire `research stack' for developing such embodied agents in simulation -- (1) data: curating house-scale interactive 3D assets (\eg kitchens with cabinets, drawers, fridges that can open/close) that support studying generalization to unseen objects, receptacles, and home layouts, (2) simulation: developing the next generation of high-performance photo-realistic 3D simulators that support rich interactive environments, (3) tasks: setting up challenging representative benchmarks to enable reproducible comparisons and systematic tracking of progress over the years.

To support this long-term research agenda, we present: 

\begin{asparaitem}
\item \textbf{\replica}: an artist-authored fully-interactive recreation of `FRL-apartment' spaces from the Replica dataset~\cite{straub2019replica} consisting of 111 unique layouts of a single apartment background with 92 authored objects including dynamic parameters, semantic class and surface annotations, and efficient collision proxies, representing 900+ person-hours of professional 3D artist effort.
\replica (illustrated in figures and videos) was created with the consent of and compensation to artists, and will be shared under a Creative Commons license for non-commercial use with attribution (CC-BY-NC).

\item \textbf{\habnamefull (\habname)}: a high-performance physics-enabled 3D simulator, representing approximately 2 years of development effort and the next generation of the Habitat project~\cite{savva2019habitat} (Habitat 1.
0).
\habname supports piecewise-rigid objects (\eg door, cabinets, and drawers that can rotate about an axis or slide), articulated robots (\eg mobile manipulators like Fetch~\cite{fetchrobot}, fixed-base arms like Franka~\cite{franka}, quadrupeds like AlienGo~\cite{aliengo}), and rigid-body mechanics (kinematics and dynamics).
The design philosophy of \habname is to prioritize performance (or speed) over the breadth of simulation capabilities.
\habname by design and choice does not support non-rigid dynamics (deformables, fluids, films, cloths, ropes), physical state transformations (cutting, drilling, welding, melting), audio or tactile sensing -- many of which are capabilities provided by other simulators~\cite{kolve2017ai2, gan2020threedworld, seita_bags_2021}.
The benefit of this focus is that we were able to design and optimize \habname to be \emph{exceedingly} fast -- simulating a Fetch robot interacting in \replica scenes at 1200 steps per second (SPS), where each `step' involves rendering 1 RGBD observation (128$\times$128 pixels) and simulating rigid-body dynamics for $\nicefrac{1}{30}$ sec.
Thus, 30 SPS would be considered `real time' and 1200 SPS is 40$\times$ real-time.
\habname also scales well -- achieving 8,200 SPS (273$\times$ real-time) multi-process on a single GPU and 26,000 SPS (850$\times$ real-time) on a single node with 8 GPUs.
For reference, existing simulators typically achieve 10-400 SPS (see \reftab{tab:comparison}).
These $100\times$ simulation-speedups correspond to cutting experimentation time from 6 months to under 2 days, unlocking experiments that were hitherto infeasible, allowing us to answer questions that were hitherto unanswerable.
As we will show, they also directly translate to training-time speed-up and accuracy improvements from training agents (for object rearrangement tasks) on more experience.

\item \textbf{\tasksuitenamefull (\tasksuitename):} a suite of common tasks for assistive robots (\cleanhouse, \stockfridge, \settable) that are specific instantiations of the generalized rearrangement problem~\cite{batra2020rearrangement}.
Specifically, a mobile manipulator (Fetch) is asked to rearrange a list of objects from initial to desired positions -- picking/placing objects from receptacles (counter, sink, sofa, table), opening/closing containers (drawers, fridges) as necessary.
The task is communicated to the robot using the GeometricGoal specification prescribed by Batra~\etal~\cite{batra2020rearrangement} -- \ie, initial and desired 3D (center-of-mass) position of each target object $i$ to be rearranged $\big(s^0_i, s^*_i\big)^N_{i=1}$.
An episode is considered successful if all target objects are placed within $15$cm of their desired positions (without considering orientation).
\footnote{
  The robot must also be compliant during execution -- an episode fails if the accumulated contact force experienced by the arm/body exceeds a threshold.
  This prevents damage to the robot and the environment.
}   
The robot operates entirely from onboard sensing -- head- and arm-mounted RGB-D cameras, proprioceptive joint-position sensors (for the arm), and egomotion sensors (for the mobile base) -- and may not access any privileged state information (no prebuilt maps or 3D models of rooms or objects, no physically-implausible sensors providing knowledge of mass, friction, articulation of containers, \etc).
Notice that an object's center-of-mass provides no information about its size or orientation.
The target object may be located inside a container (drawer, fridge), on top of supporting surfaces (shelf, table, sofa) of varying heights and sizes, and surrounded by clutter; all of which must be sensed and maneuvered.
Receptacles like drawers and fridge start closed, meaning that the agent must open and close articulated objects to succeed.
The choice of GeometricGoal is deliberate -- we aim to create the PointNav~\cite{anderson2018evaluation} equivalent for mobile manipulators.
As witnessed in the navigation literature, such a task becomes the testbed for exploring ideas \cite{wijmans2019dd, wijmans2020train, ye2020auxiliary, du2021curious, karkus2021differentiable, arpino_icra21, ramakrishnan2020occant, bansal2019combining, xiali2020relmogen} and a starting point for more semantic tasks~\cite{batra2020objectnav, ku2020room, anderson2018vision}.
The robot uses continuous end-effector control for the arm and velocity control for the base.
We deliberately focus on gross motor control (the base and arm) and not fine motor control (the gripper).
Specifically, once the end-effector is within 15cm of an object, a discrete grasp action becomes available that, if executed, snaps the object into its parallel-jaw gripper
\footnote{
  To be clear, \habname fully supports the rigid-body mechanics of grasping; 
the abstract grasping is a \emph{task-level} simplification that can be trivially undone. 
Grasping, in-hand manipulation, and goal-directed releasing of a grasp are all challenging open research problems 
\cite{murali20206, bohg2013data, hang2016hierarchical}  
that we believe must further mature in the fixed-based close-range setting before being integrated into 
a long-horizon home-scale rearrangement problem. 
}.
This follows the `abstracted grasping' recommendations in \citet{batra2020rearrangement} and is consistent with recent work~\cite{ehsani2021manipulathor}.
We conduct a systematic study of two distinctive techniques -- (1) monolithic `sensors-to-actions' policies trained with reinforcement learning (RL) at scale, and (2) classical sense-plan-act pipelines (SPA) \cite{murphy2019introduction} --   with a particular emphasis on systematic generalization to new objects, receptacles, apartment layouts  (and not just robot starting pose).
Our findings include:
\end{asparaitem}

\begin{asparaenum}
\item \textbf{Flat vs hierarchical:}
Monolithic RL policies successfully learn diverse \emph{individual} skills (pick/place, navigate, open/close drawer).  
However, crafting a combined reward function and learning scheme that elicits chaining of such skills 
for the long-horizon \tasksuitename tasks remained out of our reach. 
We saw significantly stronger results with a hierarchical approach that assumes knowledge of a perfect task planner 
(via STRIPS \cite{fikes1971strips}) to break it down into a sequence of skills. 

\item \textbf{Hierarchy cuts both ways:} 
However, a hierarchy with independent skills suffers from `hand-off problems' where a succeeding skill isn't set up for success by the preceding one -- \eg, navigating to a bad location for a subsequent manipulation, only partially opening a drawer to grab an object inside, or knocking an object out of reach that is later needed. 

\item \textbf{Brittleness of \SPAfull:}
For simple skills, SPA performs just as well as monolithic RL. 
However, it is significantly more brittle since it needs to map all obstacles in the workspace for planning.
More complex settings involving clutter, challenging receptacles, and imperfect navigation can poorly frame the target object and obstacles in the robot's camera, leading to incorrect plans. 
\end{asparaenum}

We hope our work will serve as a benchmark for many years to come.
\habname is free, open-sourced under the MIT license, and under active development.
\footnote{All code is publicly available at {\tiny\url{https://github.com/facebookresearch/habitat-lab/}}.}
We believe it will reduce the community's reliance on commercial lock-ins~\cite{todorov2012mujoco,isaac} 
and non-photorealistic simulation engines \cite{brockman2016openai,bellemare2013arcade,tassa2018deepmind}.

\begin{table*}[t]
    \captionsetup[figure]{font=small}
    \setlength{\tabcolsep}{0.5pt}
    \resizebox{\textwidth}{!}{
    \centering
    \begin{tabular}{@{}lcccccr@{}}
    \toprule 
    & \multicolumn{2}{c}{Rendering} & \multicolumn{2}{c}{Physics} & Scene  & Speed \\
    \cmidrule(lr){2-3} \cmidrule(lr){4-5}
    & Library & Supports & Library & Supports &  Complexity & (steps/sec) \\
    \midrule
    Habitat~\cite{savva2019habitat} & Magnum  & 3D scans & none & continuous navigation (navmesh) & building-scale & 3,000 \\
    AI2-THOR~\cite{kolve2017ai2} & Unity & Unity & Unity & rigid dynamics, animated interactions & room-scale & 30 - 60 \\
    ManipulaTHOR~\cite{ehsani2021manipulathor} & Unity & Unity & Unity & AI2-THOR + manipulation & room-scale & 30 - 40 \\
    ThreeDWorld~\cite{gan2020threedworld} & Unity & Unity & Unity (PhysX) + FLEX & rigid + particle dynamics & room/house-scale & 5 - 168 \\
    SAPIEN~\cite{xiang2020sapien} & OpenGL/OptiX & configurable & PhysX & rigid/articulated dynamics & object-level & 200 - 400$^\dagger$ \\
    RLBench~\cite{james2020rlbench}& CoppeliaSim (OpenGL) & Gouraud shading & CoppeliaSim (Bullet/ODE) & rigid/articulated dynamics & table-top & 1 - 60$^\dagger$ \\
    iGibson~\cite{shen2020igibson} & PyRender & PBR shading & PyBullet & rigid/articulated dynamics & house-scale & 100 \\
    \midrule
    \habnamefull (\habname) & Magnum & 3D scans + PBR shading & Bullet & rigid/articulated dynamics + navmesh & house-scale & 1,400 \\
    \bottomrule
    \end{tabular}
    }
    \caption{ 
      High-level comparison of different simulators.
      Note: Speeds were taken directly from respective publications or obtained via direct personal correspondence with the authors when not publicly available (indicated by $^\dagger$).
      Benchmarking was conducted by different teams on different hardware with different underlying 3D assets simulating different capabilities.
      Thus, these should be considered  qualitative comparisons representing what a user expects to experience on a single instance of the simulator (no parallelization).
    }
    \vspace{-15pt}
    \label{tab:comparison}
\end{table*}

\vspace{-5pt}
\section{Related Work}
\vspace{-5pt}

\xhdr{What \emph{is} a simulator?} 
Abstractly speaking, a simulator has two components: %
(1) a \emph{physics engine} that evolves the world state $s$ over time $s_t \rightarrow s_{t+1}$%
\footnote{Alternatively, $(s_t, a_t) \rightarrow s_{t+1}$ in the presence of an agent taking action $a_t$}, 
and (2) a \emph{renderer} that generates sensor observations $o$ from states: $s_t \rightarrow o_t$.
The boundary between the two is often blurred as a matter of convenience. 
Many physics engines implement minimal renderers to visualize results, 
and some rendering engines include integrations with a physics engine.
PyBullet~\cite{coumans2019pybullet}, MuJoCo~\cite{todorov2012mujoco}, DART~\cite{lee2018dart}, 
ODE~\cite{ODE}, PhysX/FleX~\cite{physx,flex}, and Chrono~\cite{chrono} 
are primarily physics engines with some level of rendering, 
while Magnum~\cite{magnum}, ORRB~\cite{orrb2019}, and PyRender~\cite{pyrender} are primarily renderers. 
Game engines like Unity~\cite{unity} and Unreal~\cite{unreal} provide tightly coupled integration of physics and rendering. 
Some simulators \cite{savva2017minos,savva2019habitat, wu2018building} involve largely static environments 
-- the agent can move but not change the state of the environment (\eg open cabinets). 
Thus, they are heavily invested in rendering with fairly lightweight physics (\eg collision checking with the agent approximated as a cylinder).

\xhdr{How are interactive simulators built today?} 
Either by relying on game engines~\cite{kolve2017ai2,yan2018chalet,puig2018virtualhome} 
or via a `homebrew' integration of existing rendering and physics libraries~\cite{gan2020threedworld,shen2020igibson,xia2020interactive,xiang2020sapien}.  
Both options have problems. 
Game engines tend to be optimized for human needs (high image-resolution, $\sim$60 FPS, persistent display) not for AI's needs~\cite{choie_pnas21} 
(10k+ FPS, low-res, `headless' deployment on a cluster). 
Reliance on them 
leads to limited control over the performance characteristics. 
On the other hand, they represent decades of knowledge and engineering effort whose value cannot be discounted. 
This is perhaps why `homebrew' efforts involve a high-level (typically Python-based) integration of 
existing libraries. 
Unfortunately but understandably, this results in simulation speeds of 10-100s of SPS, 
which is \emph{orders of magnitude} sub-optimal. 
\habname involved a deep low-level (C++) integration of 
rendering (via Magnum~\cite{magnum}) and physics (via Bullet \cite{coumans2019pybullet}),
enabling precise control of scheduling and task-aware optimizations, 
resulting in substantial performance improvements.

\xhdr{Object rearrangement.}
Task- and motion-planning \cite{garrett2020integrated} and  
mobile manipulation have a long history in AI and robotics, whose full survey is beyond the scope of this document. 
\citet{batra2020rearrangement} provide a good summary of historical background of rearrangement, a review of recent efforts, 
a general framework, and a set of recommendations that we adopt here. %
Broadly speaking, our work is distinguished from prior literature by a combination of the 
emphasis on visual perception, lack of access to state, systematic generalization, 
and the experimental setup of visually-complex and ecologically-realistic home-scale environments. 
We now situate \wrt a few recent efforts. 
\cite{garrett_icra20} study replanning in the presence of partial observability but do not consider mobile manipulation. 
\cite{xia2020interactive} tackle `interactive navigation', where the robot can bump into and push objects during navigation, but does not have an arm. 
Some works~\cite{misra2018mapping,shridhar2020alfred, weihs2021visual} abstract away gross motor control entirely by using 
symbolic interaction capabilities (\eg a `pick up X' action) or a `magic pointer' \cite{batra2020rearrangement}. 
We use abstracted grasping but not abstract manipulation.  
\cite{xiali2020relmogen} develop hierarchical methods for mobile manipulation,  combining RL policies for goal-generation and motion-planning 
for executing them. We use the opposite combination of planning and learning -- using task-planning to generate goals and RL for skills. 
\cite{ehsani2021manipulathor} is perhaps the most similar to our work.  
Their task involves moving a single object from one location to another, excluding interactions with container objects 
(opening a drawer or fridge to place an object inside). 
We will see that rearrangement of multiple objects while handling containment is a much more challenging task.
Interestingly, our experiments show evidence for the opposite conclusion reached therein -- 
monolithic end-to-end trained RL methods are outperformed by a modular approach that is trained stage-wise to handle long-horizon rearrangement tasks.

\vspace{-5pt}
\section{Replica to \replica: Creating Interactive Digital Twins of Real Spaces}
\vspace{-5pt}

We begin by describing our dataset that provides a rich set of indoor layouts for studying rearrangement tasks. 
Our starting point was Replica~\cite{straub2019replica}, a dataset of \emph{highly} photo-realistic 3D reconstructions at room and building scale. 
Unfortunately, static 3D scans are unsuitable for studying rearrangement tasks because objects in a static scan cannot be moved or manipulated. 

\begin{figure}[h]
    \centering
    \begin{subfigure}[t]{0.495\linewidth}
      \centering
      \includegraphics[width=\textwidth]{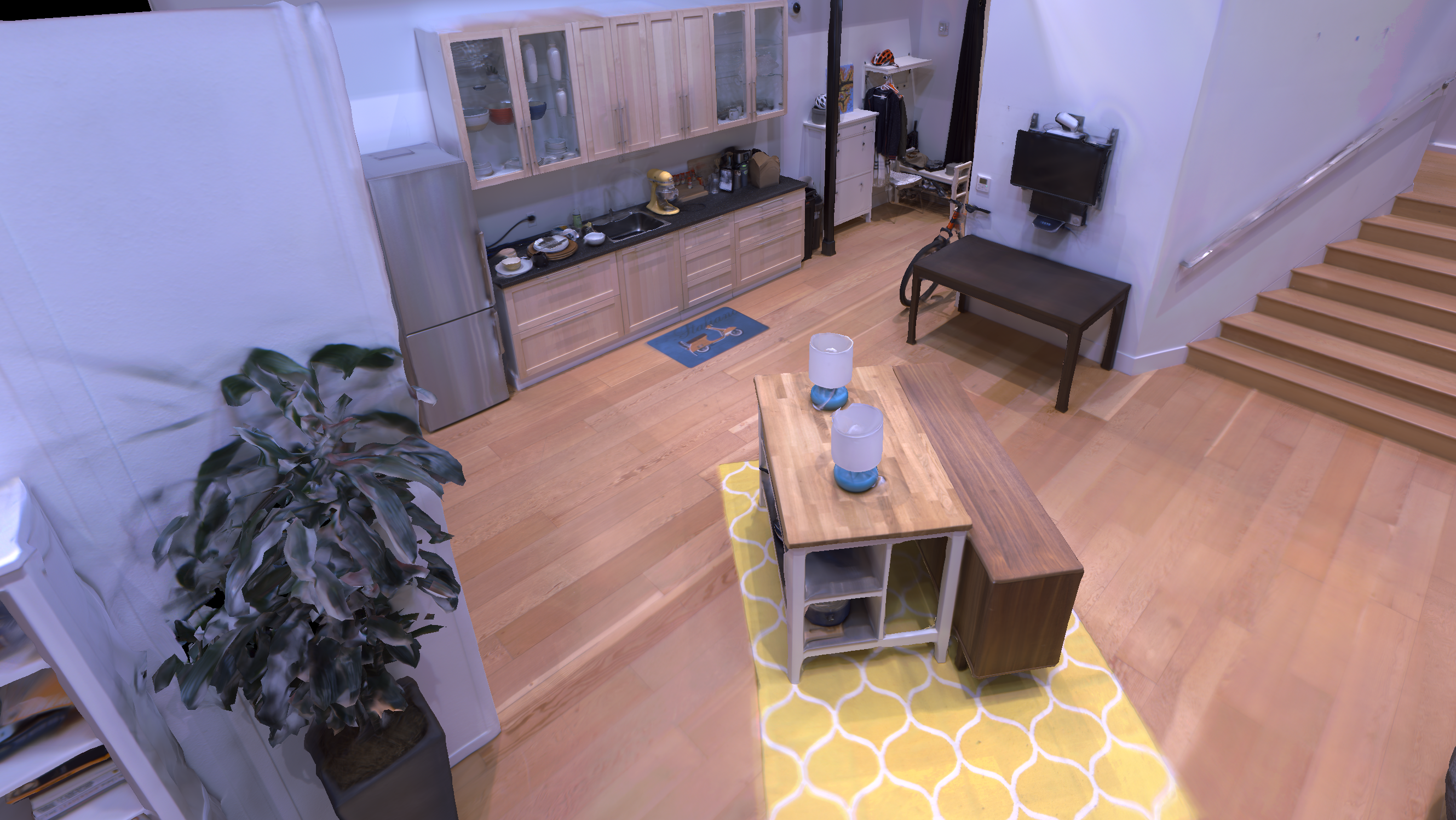}
    \end{subfigure}
    \begin{subfigure}[t]{0.495\linewidth}
      \centering
      \includegraphics[width=\textwidth]{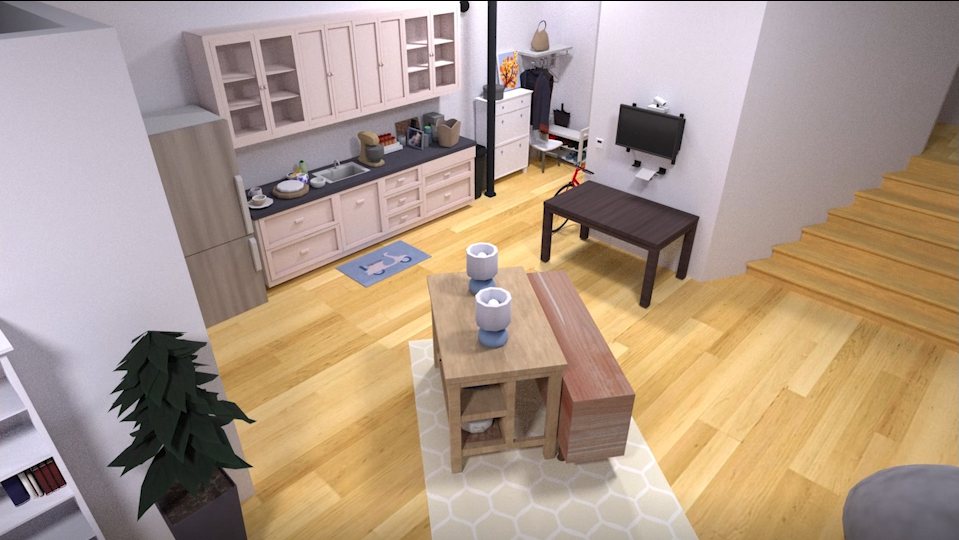}
    \end{subfigure}
    \caption{
      Left: The original Replica scene. 
      Right: the artist recreated scene \replica. 
      All objects (furniture, mugs) including articulated ones (drawers, fridge) in \replica are fully physically simulated and 
      interactive. 
    }
    \label{fig:replica_cad}
\end{figure}

\xhdr{Asset Creation.}
\replica is an artist-created, fully-interactive recreation of `FRL-apartment' spaces from the Replica dataset~\cite{straub2019replica}. 
First, a team of 3D artists authored individual 3D models (geometry, textures, and material specifications) to faithfully recreate nearly all objects 
(furniture, kitchen utensils, books, \etc; 92 in total) in all 6 rooms from the FRL-apartment spaces as well as an accompanying static backdrop (floor and walls). 
\figref{fig:replica_cad} compares a layout of \replica with the original Replica scan. 
Next, each object was prepared for rigid-body simulation by authoring physical parameters (mass, friction, restitution), 
collision proxy shapes, and semantic annotations. 
Several objects (\eg refrigerator, kitchen counter) were made `articulated' through sub-part 
segmentation (annotating fridge door, counter cabinet) and authoring of URDF files describing joint configurations 
(\eg fridge door swings around a hinge)
and dynamic properties (\eg joint type and limits). 
For each large furniture object (\eg table), we annotated surface regions 
(\eg table tops) and containment volumes (\eg drawer space) 
to enable 
programmatic placement of small objects on top of or within. %

\xhdr{Human Layout Generation.}
Next, a 3D artist authored an additional 5 semantically plausible `macro variations' of the scenes 
-- producing new scene layouts consisting only of larger furniture from the same 3D object assets. 
Each of these macro variations was further perturbed through 20 `micro variations' that re-positioned objects 
-- \eg swapping the locations of similarly sized tables or a sofa and two chairs.
This resulted in a total of 105 scene layouts that exhibit major and minor semantically-meaningful 
variations in furniture placement and scene layout, enabling controlled testing of generalization.
Illustrations of these variations can be found in \Cref{sec:supp:replica}.

\xhdr{Procedural Clutter Generation.}\label{sec:sys:task_api}%
To maximize the value of the human-authored assets %
we also develop a pipeline that allows us to generate new clutter procedurally.
Specifically, we dynamically populate 
the annotated supporting surfaces (\eg table-top, shelves in a cabinet) and containment volumes (\eg fridge interior, drawer spaces) 
with object instances from appropriate categories (\eg, plates, food items). These inserted objects can come from \replica or the YCB dataset \cite{calli2015ycb}. 
We compute physically-stable insertions of clutter offline (\ie letting an inserted bowl `settle' on a shelf) and 
then load these stable arrangements into the scene dynamically at run-time.

\replica is fully integrated with the \habname and 
a supporting configuration file structure enables simple import, instancing, and programmatic alternation of any of these interactive scenes. 
Overall, \replica represents 900+ person-hours of professional 3D artist effort so far (with augmentations in progress). 
It was was created with the consent of and compensation to artists, 
and will be shared under a Creative Commons license for non-commercial use with attribution (CC-BY-NC).
Further \replica details and statistics are in \Cref{sec:supp:replica}.

\vspace{-5pt}
\section{\habnamefull (\habname): a Lazy Simulator}
\vspace{-5pt}

\habname's design philosophy is that speed is more important than the breadth of capabilities. 
\habname achieves fast rigid-body simulation in large photo-realistic 3D scenes by being lazy and  
only simulating what is absolutely needed. 
We instantiate this principle via 3 key ideas -- 
localized physics and rendering (\secref{sec:sys:localized_phs}), 
interleaved physics and rendering (\secref{sec:sys:interleaved}), and 
simplify-and-reuse (\secref{sec:sys:simplify_reuse}). %

\subsection{Localized Physics and Rendering}
\label{sec:sys:localized_phs}
\vspace{-5pt}

Realistic indoor 3D scenes can span houses with multiple rooms (kitchen, living room), 
hundreds of objects (sofa, table, mug) and `containers' (fridge, drawer, cabinet), and thousands of parts (fridge shelf, cabinet door). 
Simulating physics for every part at all times is not only slow, it is simply unnecessary -- 
if a robot is picking a mug from a living-room table, 
why must we check for collisions between the kitchen fridge shelf and objects on it? 
We make a number of optimizations to \emph{localize} physics to the current robot interaction or part of the task -- 
(1) 
assuming that the robot is the only entity capable of applying non-gravitational forces 
and not recomputing physics updates for distant objects; 
(2) using a navigation mesh to move the robot base 
kinematically (which has been show to transfer well to real the world~\cite{kadian2020sim2real}) rather than simulating wheel-ground contact, 
(3) using the physics `sleeping' state of objects to optimize rendering by caching and re-using scene graph transformation matrices and frustum-culling results, 
and 
(4) treating all object-parts that are constrained relative to the base as static objects (\eg assuming that the walls of a cabinet will never move).

\subsection{Interleaved rendering and physics}
\label{sec:sys:interleaved}
\vspace{-5pt}

\begin{wrapfigure}{r}{0.49\textwidth}
    \includegraphics[width=1\linewidth]{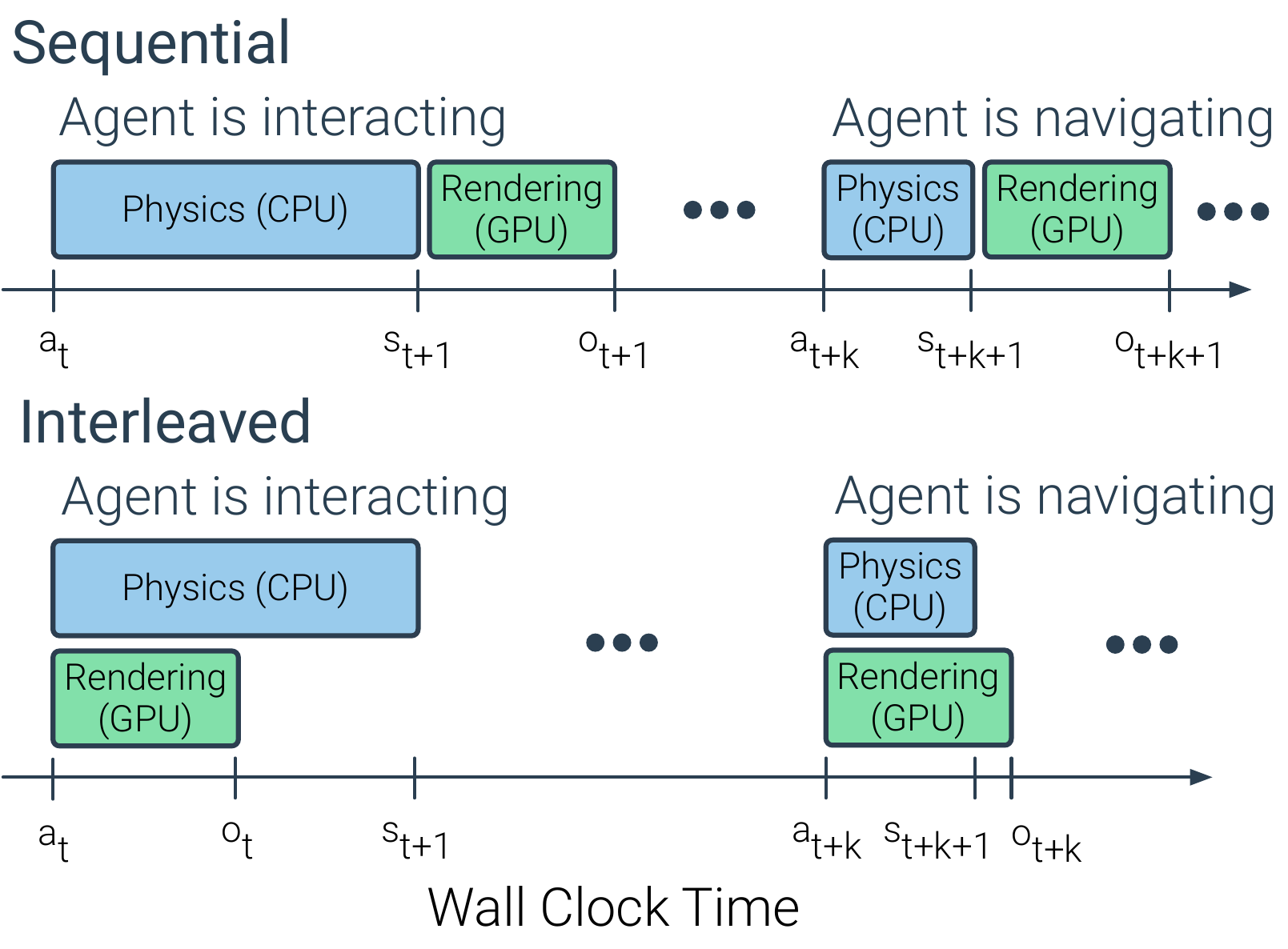}
    \vspace{-10pt}
    \caption{
      Interleaved physics and rendering.
      Top shows the normal sequential method of performing physics $(s_t, a_t) \rightarrow s_{t+1}$ then rendering $s_{t+1} \rightarrow o_{t+1}$. 
      Bottom shows \habname's interleaved physics and rendering. 
    }
    \label{fig:system}
    \vspace{-5pt}
\end{wrapfigure}

Most physics engines (\eg Bullet) run on the CPU, while rendering (\eg via Magnum) typically occurs on the GPU.
After our initial optimizations, we found each to take nearly equal compute-time. 
This represents a \emph{glaring} inefficiency -- as illustrated in \figref{fig:system}, at any given time either the CPU 
is sitting idle waiting for the GPU or vice-versa. 
Thus, interleaving them leads to significant gains. 
However, this is complicated by a sequential dependency -- 
state transitions depend on robot actions $\calT: (s_t, a_t) \rightarrow s_{t+1}$, 
robot actions depend on the sensor observations: $\pi: o_{t} \rightarrow a_{t}$, and 
observations depend on the state $\calO: s_{t} \rightarrow o_{t}$.  
Thus, 
it ostensibly appears that physics and rendering outputs ($s_{t+1}$, $o_t$)  cannot be computed in parallel from $s_t$
because computation of $a_t$ cannot begin till $o_t$ is available.  

We break this sequential dependency by changing the agent policy to be $\pi(a_t \,|\, o_{t-1})$ instead of $\pi(a_t \,|\, o_{t})$.
Thus, our agent predicts the current action $a_t$ 
not from the current observations $o_t$ but from an observation from 1 timestep ago $o_{t-1}$, essentially 
`living in the past and acting in the future'. 
This simple change means that we can generate $s_{t+1}$ on the CPU at the same time as $o_t$ is being generated on the GPU. 

This strategy not only increases simulation throughput, but also offers two other fortuitous benefits -- 
increased biological plausibility and improved sim2real transfer potential. 
The former is due to closer analogy to all sensors (biological or artificial) having a sensing latency 
(\eg, the human visual system has approximately 150ms latency~\cite{thorpe1996speed}).
The latter is due to a line of prior work~\cite{sandhasim2real,dulacarnold2020realworldrlempirical,tan2018sim} 
showing that introducing this latency in simulators improves the transfer of learned agents to reality.

\subsection{Simplify and reuse}
\label{sec:sys:simplify_reuse}
\vspace{-5pt}
Scenes with many interactive objects can pose a challenge for limited GPU memory.
To mitigate this, we apply GPU texture compression (the Basis `supercompressed' format~\cite{basis}) 
to all our 3D assets, leading to 4x to 6x (depending on the texture) reduction in GPU memory footprint. 
This allows more objects and more concurrent simulators to fit on one GPU and reduces asset import times.
Another source of slowdown are `scene resets' -- specifically, the re-loading of objects into memory as training/testing 
loops over different scenes. 
We mitigate this by pre-fetching object assets and caching them in memory, 
which can be quickly \emph{instanced} when required by a scene, thus reducing the time taken by simulator resets.
Finally, computing collisions between robots and the surrounding geometry is expensive. 
We create convex decompositions of the objects and separate these 
simplified collision meshes from the high-quality visual assets used for rendering. 
We also allow the user to specify simplified collision geometries such as bounding boxes, and per-part or merged convex hull geometry. 
Overall, this pipeline requires minimal work from the end user. 
A user specifies a set of objects, they are automatically compressed in GPU memory, cached for future prefetches, 
and convex decompositions of the object geometry are computed for fast collision calculations.

\subsection{Benchmarking}
\vspace{-5pt}

We benchmark using a Fetch robot, equipped with two RGB-D cameras 
(128$\times$128 pixels) in \replica scenes  
under two scenarios: 
\begin{inparaenum}[(1)]
\item Idle: with the robot initialized in the center of the living room somewhat far from furniture or any other object and taking random actions,  
and
\item Interact: with the robot initialized fairly close to the fridge and taking actions from a pre-computed trajectory that results in representative interaction with objects.
\end{inparaenum}
Each simulation step consists of 1 rendering pass and 4 physics-steps, 
each simulating \nicefrac{1}{120} sec for a total of \nicefrac{1}{30} sec. 
This is a fairly standard experimental configuration in robotics (with 30 FPS cameras and 120 Hz control). 
In this setting, a simulator operating at 30 steps per (wallclock) second (SPS) corresponds to `real time'. 

Benchmarking was done on machines with dual Intel Xeon Gold 6226R CPUs -- 32 cores/64 threads (32C/64T) total -- and 8 NVIDIA GeForce 2080 Ti GPUs.
For single-GPU benchmarking processes are confined to 8C/16T of one CPU, simulating an 8C/16T single GPU workstation.
For single-GPU multi-process benchmarking, 16 processes were used.
For multi-GPU benchmarking, 64 processes were used with 8 processes assigned to each GPU.
We used python-3.8 and gcc-9.3 for compiling \habname.
We report average SPS over 10 runs and a 95\% confidence-interval computed via standard error of the mean.
Note that 8 processes do not fully utilize a 2080 Ti and thus multi-process multi-GPU performance may be better on machines with more CPU cores.

\definecolor{lightgray}{gray}{0.85}
\begin{table}[h]
    \setlength{\tabcolsep}{5pt}
    \resizebox{1\textwidth}{!}{
	\begin{tabular}{@{}l rlrl rlrl rlrl @{}}
	\toprule
\multicolumn{1}{c}{} & \multicolumn{4}{c}{1 Process}    	& \multicolumn{4}{c}{1 GPU}    & \multicolumn{4}{c}{8 GPUs}                            \\ 
\cmidrule(lr){2-5} \cmidrule(lr){6-9}  \cmidrule(lr){10-13} 
				& \multicolumn{2}{c}{Idle} 	& \multicolumn{2}{c}{Interact} & \multicolumn{2}{c}{Idle} & \multicolumn{2}{c}{Interact} & \multicolumn{2}{c}{Idle} & \multicolumn{2}{c}{Interact} \\ 
\midrule
\rowcolor{lightgray}
\habname (Full) 	& 1191&{\scriptsize$\pm$36} &   510 & {\scriptsize$\pm$6} &  8186&{\scriptsize$\pm$47} & 1660&{\scriptsize$\pm$6} &  25734&{\scriptsize$\pm$301} & 7699&{\scriptsize$\pm$177}     \\
 - render opts. 		& 781&{\scriptsize$\pm$9} &  282&{\scriptsize$\pm$2} & 6709&{\scriptsize$\pm$89} & 1035&{\scriptsize$\pm$3} & 18844&{\scriptsize$\pm$285} &  5517&{\scriptsize$\pm$31} \\
 - physics opts. 		& 271&{\scriptsize$\pm$3} & 358&{\scriptsize$\pm$6} &  2290&{\scriptsize$\pm$5} & 1606&{\scriptsize$\pm$6} & 7942&{\scriptsize$\pm$50} & 6119&{\scriptsize$\pm$51} \\
 - all opts. 			& 242&{\scriptsize$\pm$2} & 224&{\scriptsize$\pm$3} & 2223&{\scriptsize$\pm$3} & 941&{\scriptsize$\pm$2} &   7192&{\scriptsize$\pm$55} &  4829&{\scriptsize$\pm$50} \\
	\bottomrule
	\end{tabular}
    }
    \vspace{2pt}
    \caption{Benchmarking \habname performance: simulation steps per second (SPS, higher better)  
    over 10 runs and a 95\% confidence-interval computed via standard error of the mean.
    We consider two scenarios: in Idle, the agent is executing random actions but not interacting with the scene, 
    while Interact uses a precomputed trajectory and thus results in representative interaction with objects.
    To put these numbers into context, see \reftab{tab:comparison}. %
    }
\label{table:benchmark}
\vspace{-15pt}
\end{table}

\tableref{table:benchmark} reports benchmarking numbers for \habname. 
We make a few observations. 
The ablations for \habname (denoted by \myquote{- render opts}, \myquote{-physics opts}, and \myquote{-all opts.}) 
show that principles followed in our system design lead to significant performance improvements.  

Our `Idle' setting is similar to the benchmarking setup of 
iGibson~\cite{shen2020igibson}, which reports 100 SPS. 
In contrast, \habname single-process \emph{with all optimizations turned off} is 240\% faster (242 vs 100 SPS). 
\habname single-process with optimizations on is $\sim$1200\% faster than iGibson (1191 vs 100 SPS). 
The comparison to iGibson is particularly illustrative since it uses the `same' physics engine (PyBullet) as \habname (Bullet).  
We can clearly see the benefit of working with the low-level C++ Bullet rather than PyBullet and the deep integration between rendering and physics. 
This required deep technical expertise and large-scale engineering over a period of 2 years. 
Fortunately, \habname will be publicly available so others do not have to repeat this work. 
A direct comparison against other simulators is not feasible due to different capabilities, assets, hardware, and experimental settings. 
But a qualitative order-of-magnitude survey is illustrative -- AI2-THOR~\cite{kolve2017ai2} achieves 60/30 SPS in idle/interact, 
SAPIEN~\cite{xiang2020sapien} achieves 200/400 SPS (personal communication), TDW~\cite{gan2020threedworld} 
achieves 5 SPS in interact, and RLBench~\cite{james2020rlbench} achieves between 1 and 60 SPS depending on the sensor suite (personal communication). 
Finally, \habname scales well -- achieving 8,186 SPS (272$\times$ real-time) multi-process on a single GPU 
and 25,734 SPS (850$\times$ real-time) on a single node with 8 GPUs.
These $100\times$ simulation-speedups correspond to cutting experimentation time from 6-month cycle to under 2 days.

\subsection{Motion Planning Integration}
\label{sec:sys:ompl_integration}
\vspace{-5pt}
Finally, \habname includes an integration with the popular Open Motion Planning Library (OMPL), giving access to a 
suite of motion planning algorithms~\cite{sucan2012open}. 
This enables easy comparison against classical sense-plan-act approaches \cite{murphy2019introduction}. These baselines are 
described in \secref{sec:exp:analysis} with details in \Cref{sec:supp:mp}.

\vspace{-5pt}
\section{Pick Task: a Base Case of Rearrangement}
\label{sec:exp:analysis}
\vspace{-5pt}

\begin{wrapfigure}{r}{0.4\textwidth}
  \centering
  \vspace{-10pt}
  \includegraphics[width=\linewidth]{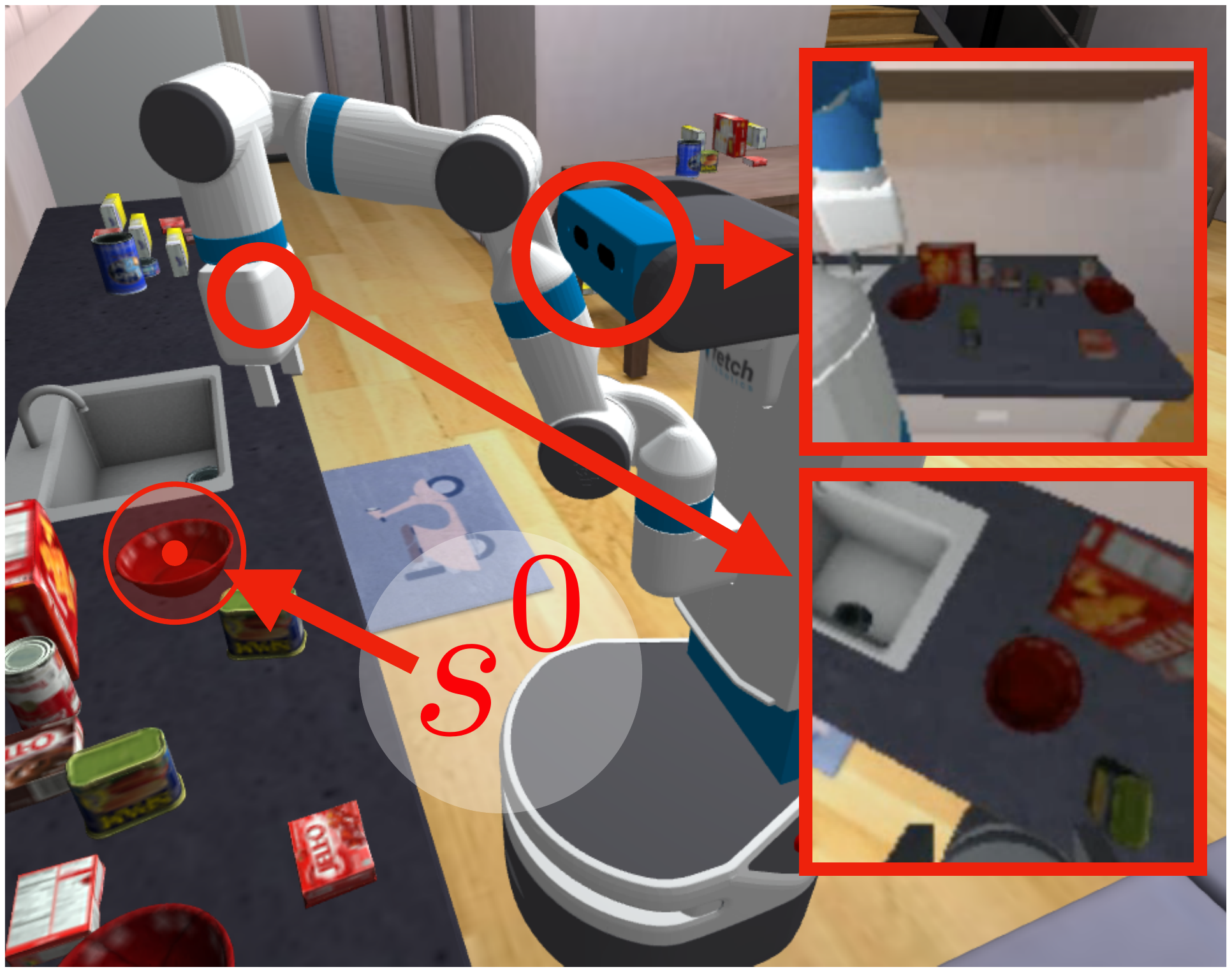}
  \caption{
    Fetch with head and arm cameras picking up a 
    bowl from the counter.
  }
  \label{fig:pick}
  \vspace{-10pt}
\end{wrapfigure}
We first carry out systematic analyses on a relatively simple manipulation task: 
picking up one object from a cluttered `receptacle'.
This forms a `base case' and an instructive starting point that we eventually expand to the more challenging 
\tasksuitenamefull (\tasksuitename) (\refsec{sec:tasks}).

\subsection{Experimental Setup}
\vspace{-5pt}

\xhdr{Task Definition: \pick$(s^0)$.} 
\figref{fig:pick} illustrates an episode in the pick task. Our agent (a Fetch robot~\cite{fetchrobot}) is spawned close to a receptacle  
(a table) that holds multiple objects (\eg cracker box, bowl). 
The task for the robot is to pick up a target object with 
center-of-mass coordinates $s^0 \in R^3$ (provided in robot's coordinate system) 
as efficiently as possible without excessive collisions. 
We study systematic generalization to new clutter layout on the receptacle, to new objects, and to new receptacles. 

\xhdr{Agent embodiment and sensing.}
Fetch~\cite{fetchrobot} is a wheeled base with a 7-DoF arm manipulator and a parallel-jaw gripper,  
equipped with two RGBD cameras ($90^\circ$ FoV, $128 \times 128$ pixels) mounted on its `head' and arm. 
It can sense its proprioceptive-state -- arm joint angles (7-dim), 
end-effector position (3-dim), and base-egomotion (6-dim, also known as GPS+Compass in the navigation literature \cite{savva2019habitat}). 
Note: the episodes in \pick are constructed such that the robot does not need to move its base. Thus, the egomotion sensor 
does not play a role in \pick but will be important in  \tasksuitename tasks (\Cref{sec:tasks}). 

\xhdr{Action space: gross motor control.}%
The agent performs end-effector control at 30Hz. 
At every step, it outputs the desired \emph{change} in end-effector position ($\delta x,\delta y, \delta z$); the desired end-effector position is fed into the inverse kinematics solver from PyBullet to derive desired states for all joints, which are used to set the joint motor targets, achieved using PD control. 
The maximum end-effector displacement per step is $1.5$cm,  
and the maximum impulse of the joint motors is 10Ns with a position gain of Kp=0.3.
In \pick, the base is fixed but in \tasksuitename, the agent also emits linear and angular velocities for the base.

\xhdr{Abstracted grasping.} 
\label{sec:agent}
The agent controls the gripper by emitting a scalar. 
If this scalar is positive and the gripper is not currently holding an object and the end-effector is within $15cm$ of an object, then 
the object closest to the end-effector is snapped into the parallel-jaw gripper.  
The grasping is perfect and objects do not slide out. 
If the scalar is negative and the gripper is currently holding an object, then the object currently held in the gripper is released and simulated as falling. 
In all other cases, nothing happens. 
For analysis of other action spaces see \Cref{supp:sec:agent}.

\xhdr{Evaluation.}
An object is considered successfully picked if the arm returns to a known `resting position' with the target object 
grasped. 
The agent fails if the accumulated contact force experienced by the arm/body exceeds a threshold of 5k Newtons. 
If the agent picks up the wrong object, the episode terminates.
Once the object is grasped, the drop action is masked out meaning the agent will never release the object. 
The episode horizon is $200$ steps.

\begin{wraptable}{r}{0.45\textwidth}
\vspace{-5pt}
\tabcolsep=0.10cm
\resizebox{0.45\textwidth}{!}{
\centering
\begin{tabular}{@{}lcccc@{}}
\toprule
\multirow{3}{*}{Method} & \multirow{3}{*}{Seen} & \multicolumn{3}{c}{Unseen} \\ 
\cmidrule(l){3-5} 
& & Layouts & Objects & Receptacles \\
\midrule
\monolithic   & $91.7$ {\tiny $\pm 1.1$} & $86.3$ {\tiny $\pm 1.4$} & $74.7$ {\tiny $\pm 1.8$} & $52.7$ {\tiny $\pm 2.0$} \\
\SPA & $70.2$ {\tiny $\pm 1.9$} & $72.7$ {\tiny $\pm 1.8$} & $72.7$ {\tiny $\pm 1.8$} & $60.3$ {\tiny $\pm 2.0$} \\
\midrule
\SPAoracle & $77.0$ {\tiny $\pm 1.7$} & $80.0$ {\tiny $\pm 1.6$} & $79.2$ {\tiny $\pm 1.7$} & $60.7$ {\tiny $\pm 2.0$} \\
\bottomrule
\end{tabular}
}
\caption{
  \pick generalization analysis: success rates with mean and standard error on $600$ episodes (and across $3$ seeds for \monolithic).
}
\label{table:generalization}
\vspace{-5pt}
\end{wraptable}

\xhdr{Methods.}
We compare two methods representing two distinctive approaches to this problem: \\[5pt]
\begin{inparaenum}
\item \monolithic: a `sensors-to-actions' policy trained end-to-end with reinforcement learning (RL). The visual input is encoded using a CNN, concatenated with embeddings of proprioceptive-sensing and goal coordinates, and fed to a recurrent actor-critic network, 
trained with DD-PPO~\cite{wijmans2019dd} for 100 Million steps of experience (see \Cref{sec:supp:mono} for details). 
This baseline 
translates our community's most-successful paradigm yet from navigation to manipulation. 

\item \SPAfull (\SPA) pipeline: 
Sensing consists of 
constructing an accumulative 3D point-cloud of the scene from depth sensors, which is then used for collision queries. 
Motion planning is done using Bidirectional RRT~\cite{lavalle2006planning} in the arm joint configuration space
(see \Cref{sec:supp:mp}). 
The controller was described in `Action Space' above and is consistent with \monolithic. 
We also create \SPAoraclefull (\SPAoracle), 
that uses \emph{privileged} information 
-- perfect knowledge of scene geometry (from the simulator) 
and a perfect controller (arm is kinematically set to desired joint poses).
The purpose of this baseline is to provide an upper-bound on the performance of \SPA.

\end{inparaenum}

\begin{figure}
  \centering
  \includegraphics[width=\textwidth]{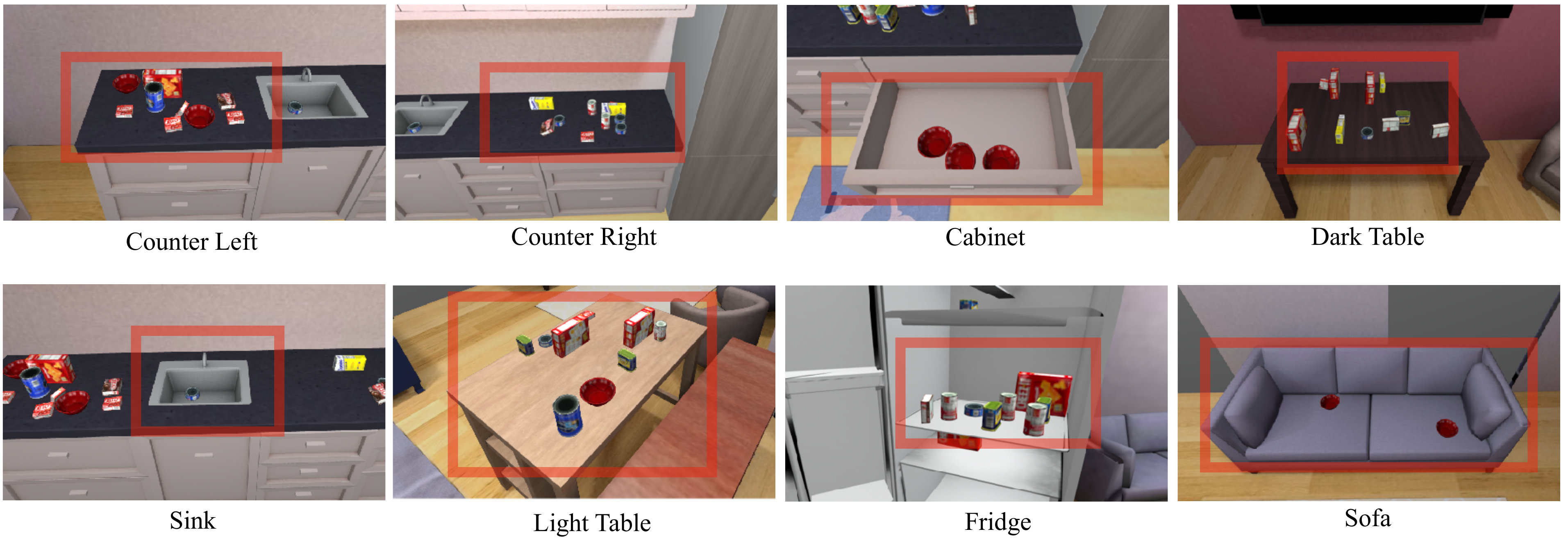}
  \caption{
    Receptacles for Pick task training. 
    One policy is trained to pick objects across all receptacles.
    Some receptacles such as the Fridge, Sink, and Cabinet are more challenging due to tight spaces and obstacle geometry. 
  }
  \label{supp:fig:receptacles}
\end{figure}

\subsection{Systematic Generalization Analysis} 
\label{sec:exps:gen_results}
\vspace{-5pt}

\begin{wrapfigure}{r}{0.5\textwidth}
  \centering
  \vspace{-10pt}
  \includegraphics[width=\linewidth]{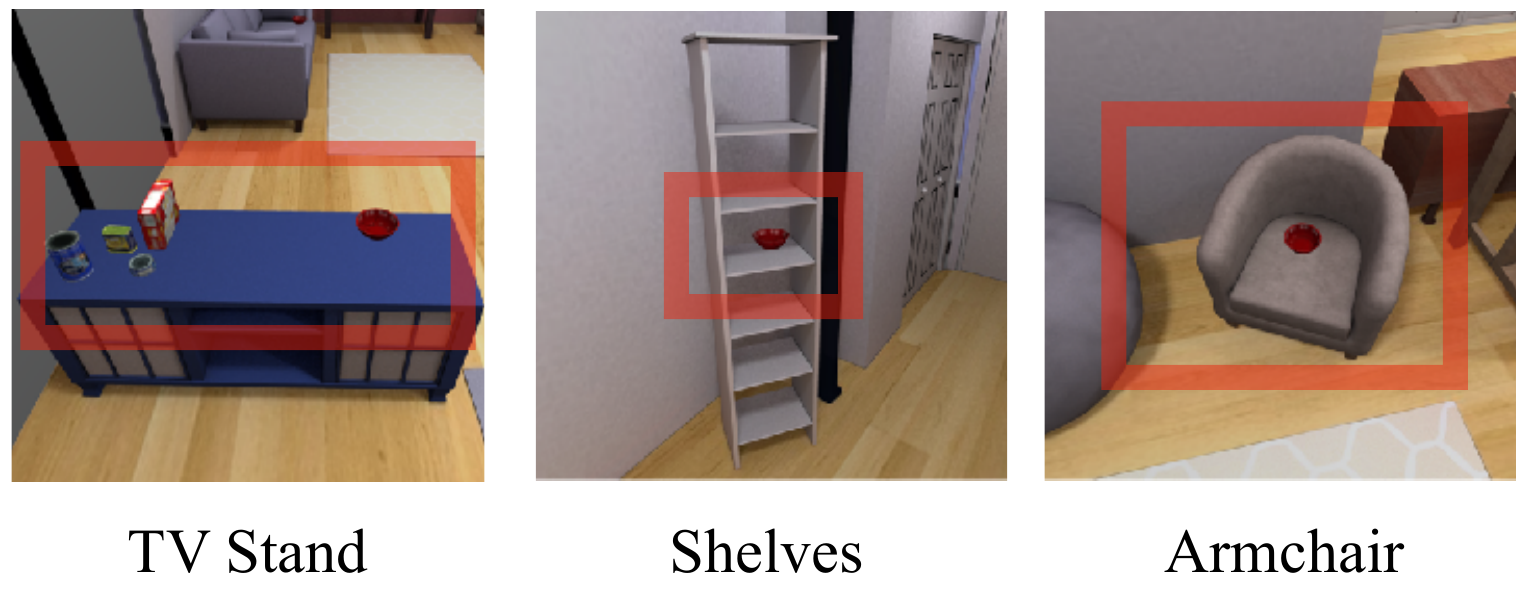}
  \caption{
    The agent is evaluated on the three unseen receptacles above. 
    These receptacles were hand chosen to test diverse solutions. 
    The chair requires avoiding the side arms, the shelf requires picking from a confined shelving space from a side angle, and finally the TV stand is visually different from those in training. 
  }
  \label{supp:fig:gen_receps}
\end{wrapfigure}

With \habname we can compare how learning based systems generalize compared to \SPA architectures.
\reftab{table:generalization} shows the results of a systematic generalization study of 
4 unseen objects, 3 unseen receptacles, and 20 unseen apartment layouts (from 1 unseen `macro variation' in \replica). 
In training the agent sees 9 objects from the kitchen and food categories of the YCB dataset (chef can, cracker box, sugar box, tomato soup can, tuna fish cap, pudding box, gelatin box, potted meat can, and bowl). 
During evaluation it is tested on 4 unseen objects (apple, orange, mug, sponge).
Likewise, the agent is trained on the counter, sink, light table, cabinet, fridge, dark table, and sofa receptacles 
(visualized in \figref{supp:fig:receptacles}) but evaluated on the unseen receptacles of tv stand, shelves, and chair (visualized in \figref{supp:fig:gen_receps}).

\monolithic generalizes fairly well from seen to unseen layouts ($91.7 \rightarrow 86.3$\%), significantly outperforming 
\SPA ($72.7$\%) and even \SPAoracle ($80.0$\%). 
However, generalization to new objects is challenging ($91.7 \rightarrow 74.7$\%) as a result of the new visual feature 
distribution and new object obstacles. 
Generalization to new receptacles is poor ($91.7 \rightarrow 52.7$\%). However, the performance drop of \SPA (and qualitative results) suggest that  
the unseen receptacles (shelf, armchair, tv stand) may be objectively more difficult to pick up objects 
from since the shelf and armchair are tight constrained areas whereas the majority of the training receptacles, such as counters and tables, have no such constraints 
(see \figref{supp:fig:gen_receps}).
We believe the performance of \monolithic will naturally improve as more 3D assets for receptacles become available; we cannot make any such claims for \SPA. 
\vspace{-3pt}

\subsection{Sensor Analysis for \monolithic: Blind agents learn to \pick}
\label{sec:sensor_ablate}
\vspace{-5pt}

We also use \habname to analyze sensor trade-offs at scale (70M steps of training).
We use the training and evaluation setting from \refsec{sec:exps:gen_results}. 

\begin{wrapfigure}{r}{0.4\textwidth}
    \centering
    \vspace{-10pt}
    \includegraphics[width=0.4\textwidth]{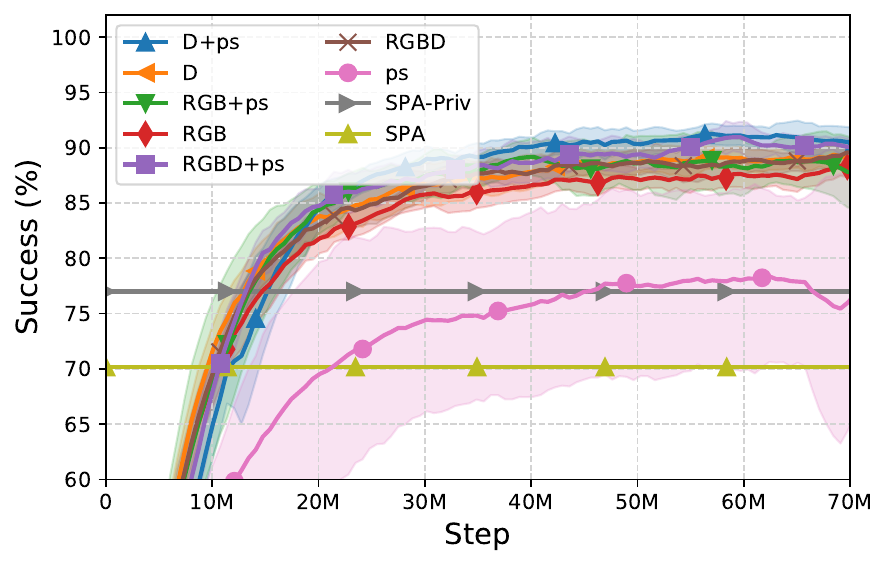}
    \caption{
     \monolithic sensor ablations: Success rates on unseen layouts ($N$=$500$) 
     vs training steps. Mean and std-dev over 3 training runs.
    }
    \label{fig:mod_pick_analysis}
    \vspace{-10pt}
\end{wrapfigure}
\Cref{fig:mod_pick_analysis} shows success rates on unseen layouts, but seen receptacles and objects types, 
vs training steps of experience for \monolithic 
equipped with different combinations of sensors \{Camera $RGB$, Depth $D$, proprioceptive-state $ps$\}. 
To properly handle sensor modality fusions, we normalize the image and state inputs using a per-channel moving average.
We note a few key findings: 
\begin{asparaenum}
\item Variations of $RGB$ and $D$ all perform similarily, but $D$+$ps$ slightly performs marginally better ($ \sim 0.5\%$ over $RGBD$+$ps$ and $ \sim 2\%$ over $RGB$+$ps$).
  This is consistent with  findings in the navigation literature~\cite{wijmans2020train} and fortuitous since depth sensors are faster to render than $RGB$.
\item Blind policies, \ie operating entirely from proprioceptive sensing are \emph{highly} effective ($78$\% success). 
This is surprising because for unseen layouts, the agent has no way to `see' the clutter; thus, we would 
expect it to collide with the clutter and trigger failure conditions. Instead, we find that the agent learns to 
`feel its way' towards the goal while moving the arm slowly so as to not incur heavy collision forces. 
Quantitatively, blind policies exceed the force threshold 2x more than sighted ones and pick the wrong object 3x more. 
  We analyze this hypothesis further in \Cref{sec:blind_policy_analysis}.
\end{asparaenum}

We also analyze different camera placements on the Fetch robot in \Cref{sec:cam_placement} and find the combination of arm and head camera to be most effective. 
For further analysis experiments, see \Cref{sec:supp:self_tracking} for qualitative evidence of self-tracking, \Cref{sec:supp:time_delay} for the effect of the time delay on performance, and \Cref{supp:sec:agent} for a comparison of different action spaces.

\vspace{-5pt}
\section{\tasksuitenamefull (\tasksuitename)}
\label{sec:tasks}
\vspace{-5pt}

We now describe our benchmark of common household assistive robotic tasks. 
We stress that these tasks \emph{illustrate} 
the capabilities of \habname but do not \emph{delineate} them -- 
a lot more is possible but not feasible to pack into a single coherent document with clear scientific takeaways. 

\subsection{Experimental Setup}
\vspace{-5pt}

\begin{figure}
  \centering
  \begin{subfigure}{\textwidth}
      \includegraphics[width=\textwidth]{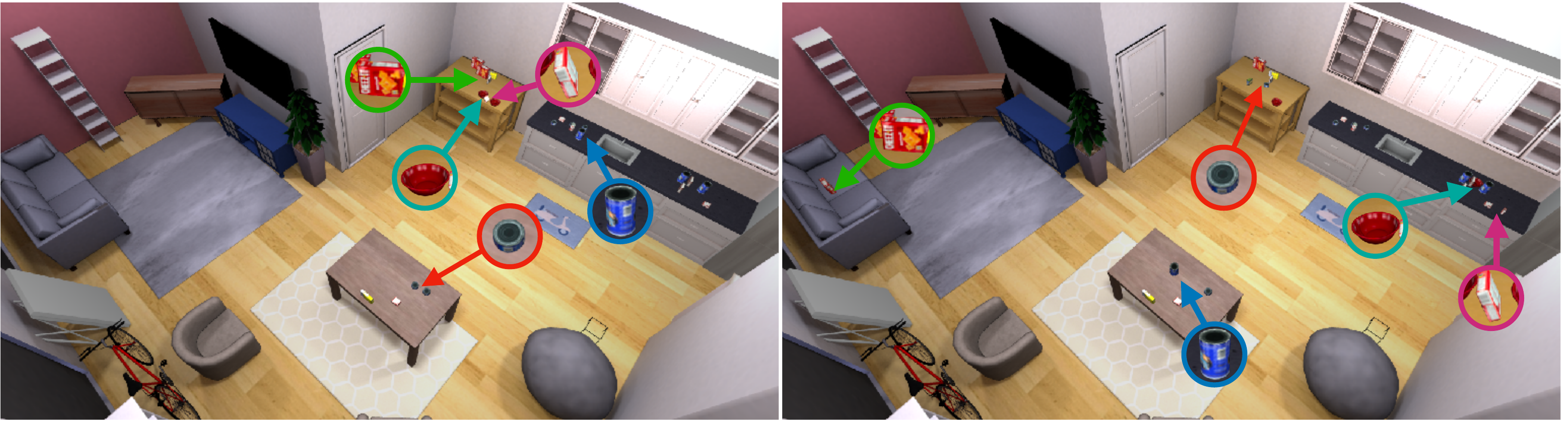}
      \caption{\cleanhouse}
      \label{supp:fig:clean_house_task}
  \end{subfigure}
  \begin{subfigure}{\textwidth}
      \includegraphics[width=\textwidth]{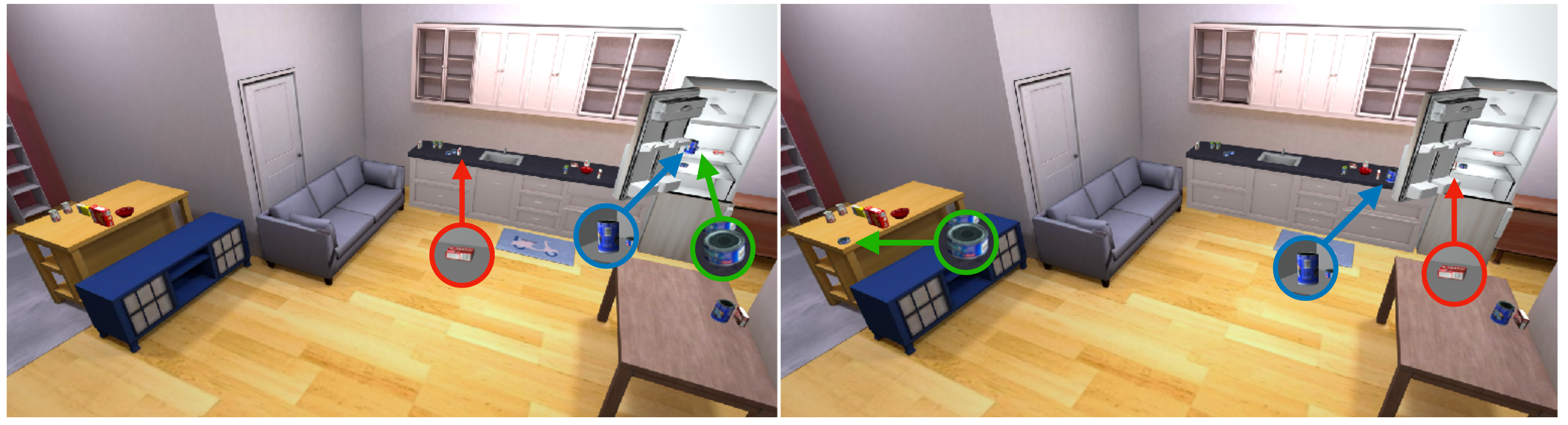}
      \caption{\stockfridge}
      \label{supp:fig:clean_house_hard_task}
  \end{subfigure}
  \begin{subfigure}{\textwidth}
      \includegraphics[width=\textwidth]{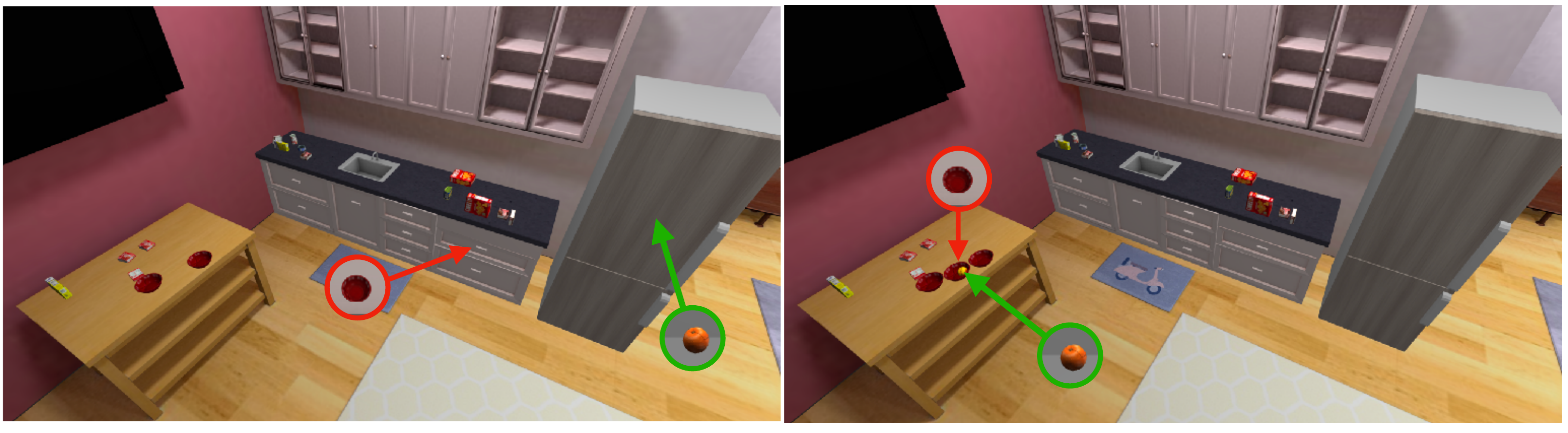}
      \caption{\settable}
      \label{supp:fig:set_table}
  \end{subfigure}
  \caption{
    Example start and goal state for \cleanhouse, \stockfridge, and \settable.
    Left column: example starting state for tasks, right column: associated goal state color coded by object. 
    Inset images and arrows denote the object start or goal position.
    Objects in \settable start in the closed drawer and fridge.
  }
\end{figure}

\xhdr{Task Definition.} 
We study three (families of) long-range tasks that correspond to common activities:
\begin{compactenum}
  \item \cleanhouse: Move 5 objects from random (unimpeded) locations back to where they belong (see \reffig{supp:fig:clean_house_task}). 
  This task requires no opening or closing and no objects are contained.
  \begin{itemize}
    \item Start: 5 target objects objects spawned in 6 possible receptacles (excluding fridge and drawer). 
    \item Goal: Each target object is assigned a goal in a different receptacle than the starting receptacle.
    \item Task length: 5000 steps.
  \end{itemize}
  \item \stockfridge: Remove 2 objects from the fridge to the counters and place one object back in the fridge (see \reffig{supp:fig:clean_house_hard_task}). 
  This task requires no opening or closing and no objects are contained.
  \begin{itemize}
    \item Start: 2 target objects in the fridge and one on the left counter. The fridge is fully opened.
    \item Goal: The goal for the target objects in the fridge are on the right counter and light table. The goal for the other target object is in the fridge. 
    \item Task length: 4000 steps
  \end{itemize}
  \item \settable: 
    Get a bowl from a drawer, a fruit from fridge, place the fruit in the bowl on the table (see \reffig{supp:fig:set_table}).
  \begin{itemize}
    \item Start: A target bowl object is in one of the drawers and a target fruit object in the middle fridge shelf. 
      Both the fridge and drawer start closed.
    \item Goal: The goal for the bowl is on the light table, the goal for the fruit is on top of the bowl. 
      Both the fridge and drawer must be closed.
    \item Task length: 4500 steps.
  \end{itemize}
\end{compactenum}
The list is in increasing order of complexity -- from no interaction with containers (\cleanhouse), to picking and placing from the fridge container (\stockfridge), to opening and closing containers (\settable). 
Note that these descriptions are provided purely for human understanding;
the robot operates entirely from a GeometricGoal specification~\cite{batra2020rearrangement} 
-- given by the initial and desired 3D (center-of-mass) position of each target object $i$ to be moved 
$\big(s^0_i, s^*_i\big)^N_{i=1}$. 
Thus, \pick$(s^0_i)$ is a special case where $N=1$ and $s^*_i$ is a constant (arm resting) location. 
For each task episode, we sample a \replica layout with YCB~\cite{calli2015ycb} objects randomly placed on feasible placement regions (see procedural clutter generation in \Cref{sec:sys:task_api}).
Each task has 5 clutter objects per receptacle.
Unless specified, objects are sampled from the `food' and `kitchen' YCB item categories in the YCB dataset.

The agent is evaluated on unseen layouts and configurations of objects, and so cannot simply memorize. 
We characterize task difficulty by the required number of rigid-body transitions (\eg, picking up a bowl, opening a drawer).
The task evaluation, agent embodiment, sensing, and action space remain unchanged from \Cref{sec:exp:analysis}, 
with the addition of base control via velocity commands.
Further details on the statistics of the rearrangement episodes, as well as the evaluation protocols are in \Cref{sec:supp:hab_task}.

\xhdr{Methods.}\label{sec:methods}
We extend the methods from \secref{sec:exp:analysis} to better handle the above long-horizon tasks with a high-level STRIPS planner using a parameterized set of skills: \pick, \place, \opendoor, \closedoor, \opendrawer, \closedrawer, and \navigate.
The full details of the planner implementation and how methods are extended are in \Cref{sec:supp:methods}.
Here, we provide a brief overview.
\begin{asparaenum}

\item \monolithic: Essentially unchanged from \secref{sec:exp:analysis}, with the exception of 
accepting a list of start and goal coordinates $\big(s^0_i, s^*_i\big)^N_{i=1}$, as opposed to just $s^0_1$. 

\item \TPSfull (\TPS): a hierarchical approach that assumes knowledge of a perfect task planner 
(implemented with STRIPS~\cite{fikes1971strips}) and the initial object containment needed by the task planner to break down a task into a sequence of parameterized skills: \navigate, \pick, \place, \opendoor, \closedoor, \opendrawer, \closedrawer.
Each skill is functionally identical to \monolithic in \secref{sec:exp:analysis} -- 
taking as input a single 3D position, either $s^0_i$ or $s^*_i$. 
For instance, in the \settable task, let ($a^0, a^*$) and ($b^0, b^*$)
denote the start and goal positions of the apple and bowl, respectively. 
The task planner converts this task into: 
\vspace{-3pt}
\begin{equation*}
\resizebox{0.9\textwidth}{!}{
$
\overbrace{\navigate(b^0), \opendrawer(b^0)}^{\text{Open Drawer}}, 
\overbrace{\pick(b^0), \navigate(b^*), \place(b^*)}^{\text{Transport Bowl}}, 
\overbrace{\navigate(b^0), \closedrawer(b^0)}^{\text{Close Drawer}},
$
}
\end{equation*}\vspace{-10pt}
\begin{equation*}
\resizebox{1\textwidth}{!}{
$
\underbrace{\navigate(a^0), \opendoor(a^0), \navigate(a^{0})}_{\text{Open Fridge}}, 
\underbrace{\pick(a^{0}), \navigate(a^*), \place(a^*)}_{\text{Transport Apple}}, 
\underbrace{\navigate(a^0), \closedoor(a^0)}_{\text{Close Fridge}}. 
$
}
\end{equation*}
Simply listing out this sequence highlights the challenging nature of these tasks. 

\item \TPSPAfull (\TPSPA): Same task planner as above, with each skill implemented via 
\SPA from \secref{sec:exp:analysis} except for \navigate where the same learned navigation policy from \TPSPA is used. 
\TPSPAoracle is analogously defined. 
Crafting an \SPA pipeline for opening/closing unknown articulated containers is an open unsolved problem in robotics -- involving 
detecting and tracking articulation \cite{schmidt2014dart, hartanto2020hand} without models, 
constrained full-body planning \cite{berenson2011task, burget2013whole, kingston2018sampling} without hand engineering constraints, 
and designing controllers to handle continuous contact \cite{meeussen2010autonomous, jain2010pulling} -- 
making it out of scope for this work. Thus, we do not report \TPSPA on \settable. 

\end{asparaenum}

\subsection{Results and Findings} 
\label{sec:hab_rearrang_results} 

\Cref{fig:all_full_task} shows progressive success rates for different methods on all tasks. 
Due to the difficulty of the full task, for analysis, the X-axis lists the sequence of agent-environment interactions (pick, place, open, close) required to accomplish the task, same as that used by the task-planner.\footnote{This sequence from the task plan is useful for experimental analysis and debugging, but does not represent the only way to solve the task and should be disposed in future once methods improve on the full task.} 
The number of interactions is a proxy for task difficulty and the plot is analogous to precision-recall curves 
(with the ideal curve being a straight line at 100\%). 
Furthermore, since navigation is often executed between successive skills, we include versions of the task planning methods with an oracle navigation skill. 
We make the following observations:  

\begin{figure}[t!]
    \centering
      \includegraphics[width=\textwidth]{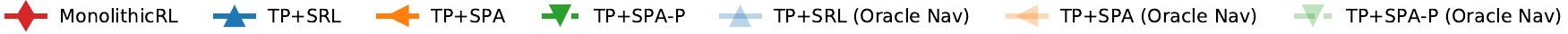}
    \begin{subfigure}{0.34\textwidth}
        \includegraphics[width=\textwidth]{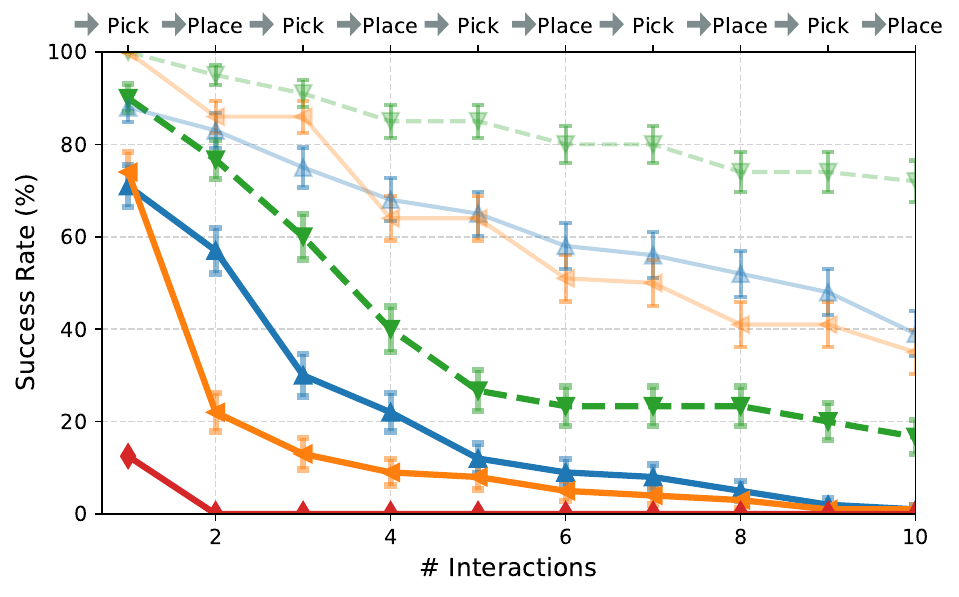}
        \caption{\cleanhouse}
        \label{fig:all_full_task:clean_house}
    \end{subfigure}
    \begin{subfigure}{0.30\textwidth}
        \includegraphics[width=\textwidth]{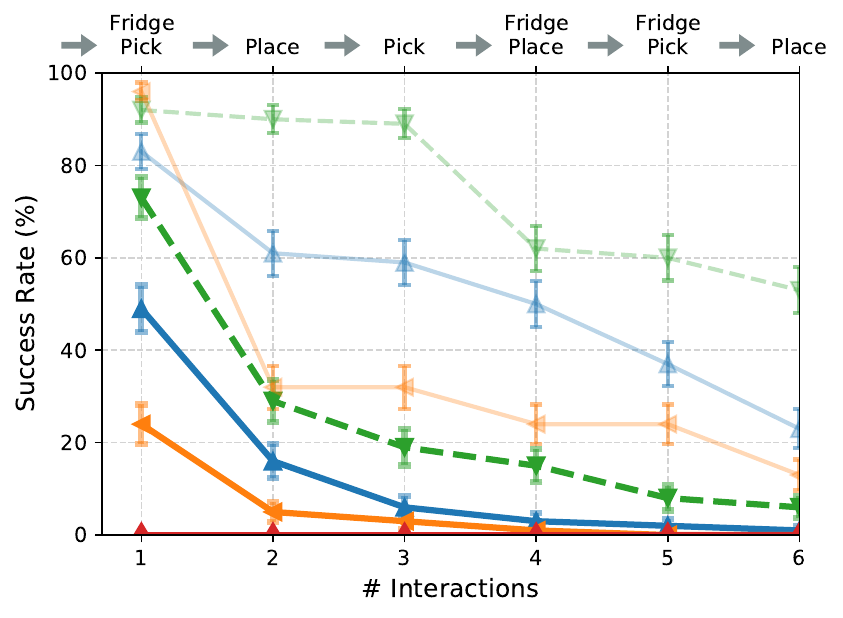}
        \caption{\stockfridge}
        \label{fig:all_full_task:stock_fridge}
    \end{subfigure}
    \begin{subfigure}{0.34\textwidth}
    	\includegraphics[width=\textwidth]{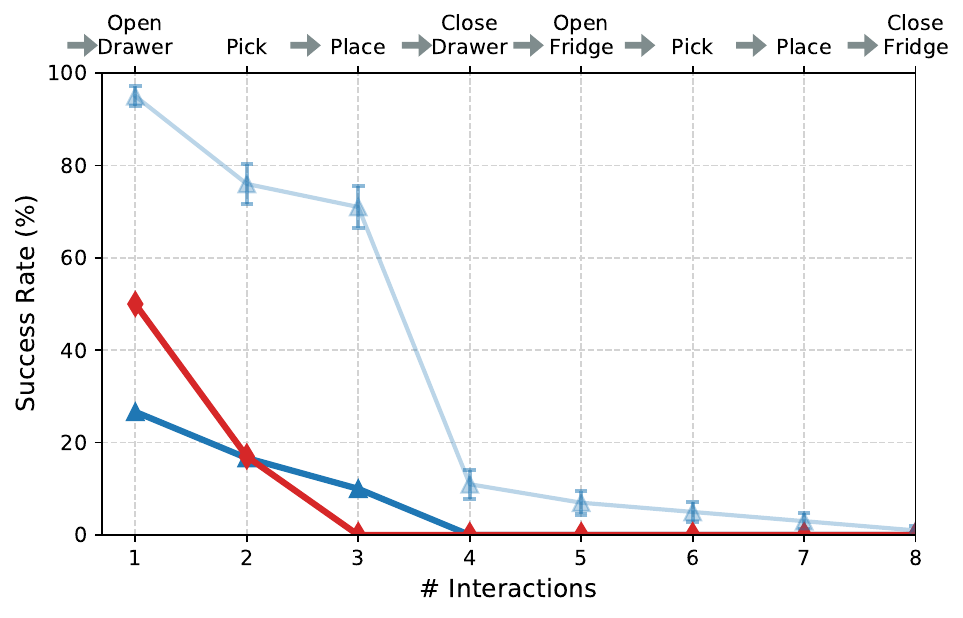}
      \caption{\settable}
      \label{fig:all_full_task:set_table}
    \end{subfigure}
    \caption{
      Success rates for \tasksuitenamefull tasks. 
      Due to the difficulty of full \tasksuitename tasks, we analyze performance as completing a part of the overall task. 
      For the TP methods that use an explicit navigation skill, we indicate with an arrow in the interaction names where navigation occurs and include versions for learned and oracle navigation. 
      Results are on unseen layouts with mean and standard error computed for 100 episodes. 
    }
    \vspace{-10pt}
    \label{fig:all_full_task}
\end{figure}
\begin{asparaenum}
\item \monolithic performs abysmally. We were able to train \emph{individual} skills with 
RL to reasonable degrees of success (see \Cref{sec:supp:skill_exps}). However, crafting a \emph{combined} reward function and learning scheme 
that elicits chaining of such skills for a long-horizon task, without any architectural inductive bias about the task structure,     
remained out of our reach despite prolonged effort.  
\item Learning a navigation policy to chain together skills is challenging as illustrated by the performance drop between learned and oracle navigation. 
In navigation for the sake of navigation (PointNav~\cite{anderson2018evaluation}), 
the agent is provided coordinates of the reachable goal location. 
In navigation for manipulation (\navigate), the agent is provided coordinates of a target object's center-of-mass 
but needs to navigate to an unspecified non-unique \emph{suitable} location from where the object is manipulable. 

\item Compounding errors hurt performance of task planning methods. 
  Even with the relatively easier skills in \cleanhouse in \Cref{fig:all_full_task:clean_house} all methods with oracle navigation gradually decrease in performance as the number of required interactions increases. 

\item Sense-plan-act variants scale poorly to increasing task complexity.  
  In the easiest setting, \cleanhouse with oracle navigation (\Cref{fig:all_full_task:clean_house}), \TPSPA performs better than \TPS. 
  However, this trend is reversed with learned navigation since \TPSPA methods, 
  which rely on egocentric perception for planning, are not necessarily correctly positioned to sense the workspace. 
  In the more complex task of \stockfridge (\Cref{fig:all_full_task:stock_fridge}), \TPS outperforms \TPSPA both with and without oracle navigation 
  due to the perception challenge of the tight and cluttered fridge.
  \TPSPA fails to find a goal configuration 3x more often and fails to find a plan in the allowed time 3x more often in \stockfridge than \cleanhouse.
\end{asparaenum}

See \Cref{sec:supp:hab_exp} for individual skill success rates, learning curves, and \SPA failure statistics. 

\vspace{-5pt}
\section{Societal Impacts, Limitations, and Conclusion}
\label{sec:conclusion}
\vspace{-5pt}

\replica was modeled upon apartments in one country (USA). 
Different cultures and regions may have different layouts of furniture, types of furniture, and types of objects not represented in \replica; 
and this lack of representation can 
have negative social implications for the assistants developed. 
While \habname is a fast simulator, we find that the performance of the overall simulation+training loop is bottlenecked 
by factors like synchronization of parallel environments and reloading of assets upon episode reset. 
An exciting and complementary future direction is holistically reorganizing the rendering+physics+RL interplay as studied by 
 \cite{dalton2020accelerating, stooke2019rlpyt, espeholt2018impala, espeholt2019seed, petrenko2020sample, shacklett2021large}. 
Concretely, as illustrated in \Cref{fig:system}, there is idle GPU time when rendering is faster than physics, because inference waits for both 
$o_t$ and $s_{t+1}$ to be ready despite not needing $s_{t+1}$. 
This is done to maintain compatibility with existing RL training systems, which expect 
the reward $r_t$ to be returned when the agent takes an action $a_t$, 
but $r_t$ is typically a function of $s_t$, $a_t$, and $s_{t+1}$. 
Holistically reorganizing the rendering+physics+RL interplay is an exciting open problem for future work.

We presented the \replica dataset, the \habnamefull platform and a home assistant benchmark.
\habname is a fully interactive, high-performance 3D simulator that enables efficient experimentation 
involving embodied AI agents rearranging richly interactive 3D environments. 
Coupled with the \replica data these improvements allow us to investigate the performance of RL policies against classical MP approaches for the suite of challenging rearrangement tasks we defined.
We hope that the \habnamefull platform will catalyze work on embodied AI for interactive environments.

{\small
\bibliographystyle{unsrtnat}
\setlength{\bibsep}{0pt}
\bibliography{main}
}
\newpage

\newpage 
\appendix

\section{\replica Further Details}
\label{sec:supp:replica} 

\begin{figure}[h]
    \centering
    \begin{subfigure}[t]{\linewidth}
    	\centering
    	\includegraphics[width=\textwidth]{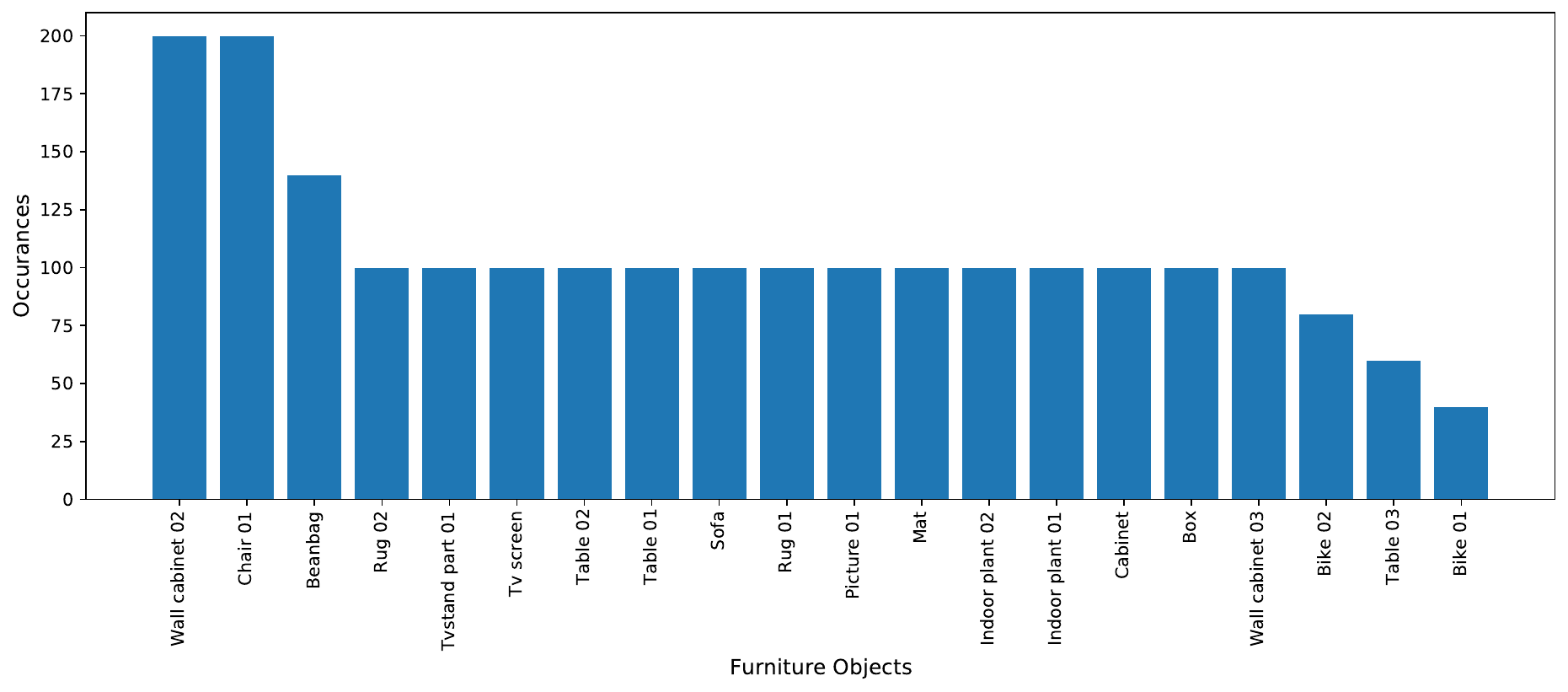}
    \end{subfigure} \\    
    \caption{
      Number of occurrences for each furniture type across the 100 micro-variations out of the total 111 \replica scenes.
    }
    \label{fig:furn_counts}
\end{figure}

The 20 micro-variations of the 5 macro-variations of the scene were created with the rule of swapping at least two furniture pieces and perturbing the positions of a subset of the other furniture pieces.
The occurrences of various furniture objects in these 100 micro-variations are illustrated in \reffig{fig:furn_counts}.
Several furniture objects such as `Beanbag' and `Chair' occur more frequently with multiple instances in a some scenes while others such as `Table 03' occur less frequently. 

\begin{figure}
    \centering
    \begin{subfigure}[t]{\linewidth}
    	\centering
    	\includegraphics[width=\textwidth]{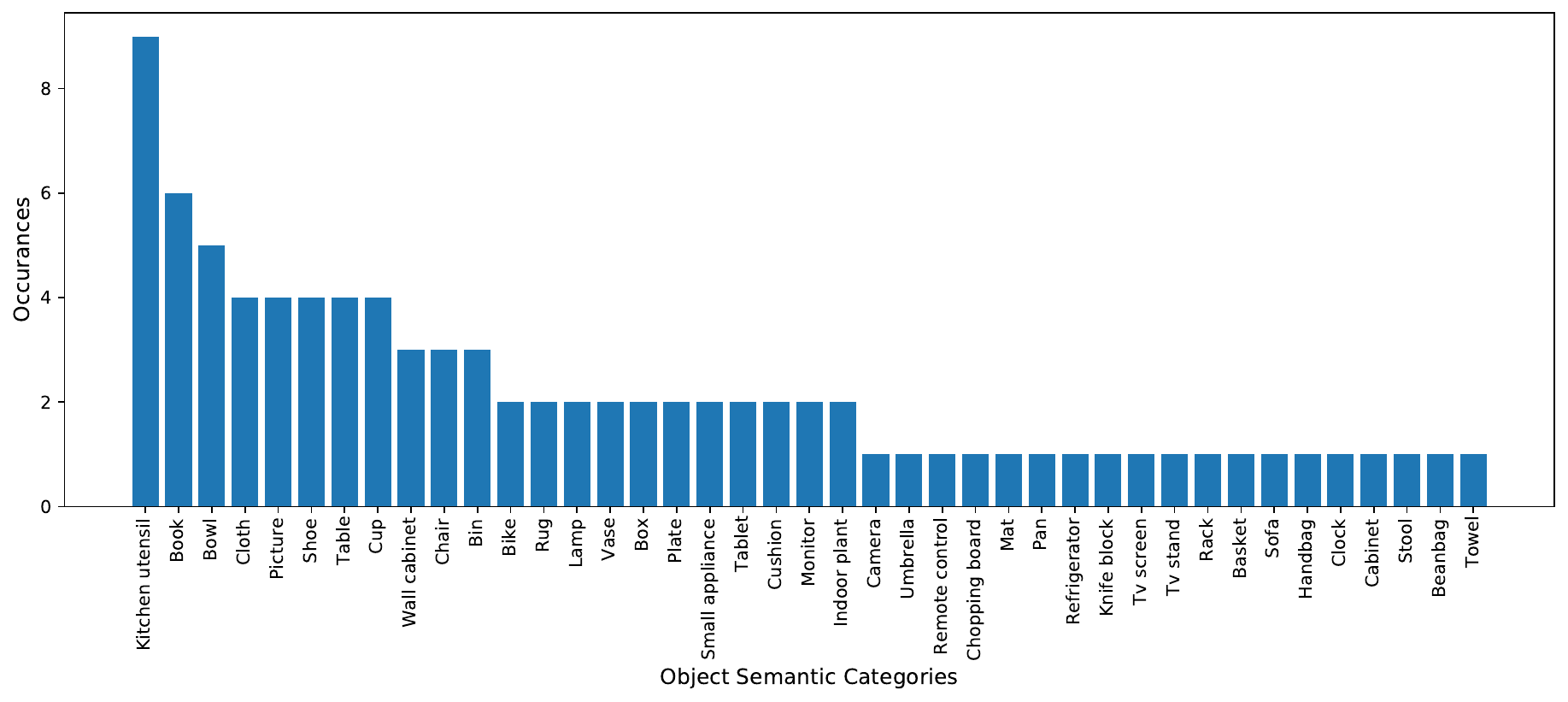}
    \end{subfigure} \\    
    \caption{
      Histogram of objects belonging to each semantic category out of the 92 overall objects.
    }
    \label{fig:obj_counts}
\end{figure}

We also analyze the object categories of all objects in the original 6 `FRL-apartment' space recreations. 
We map each of the 92 objects to a semantic category and list the counts per semantic category in a histogram in \reffig{fig:obj_counts}.
Since these spaces have a large kitchen area, there is a larger ratio of kitchen objects such as `Kitchen utensil' and 'Bowl'.

\begin{figure}
    \centering
    \centering
    \includegraphics[width=0.19\textwidth]{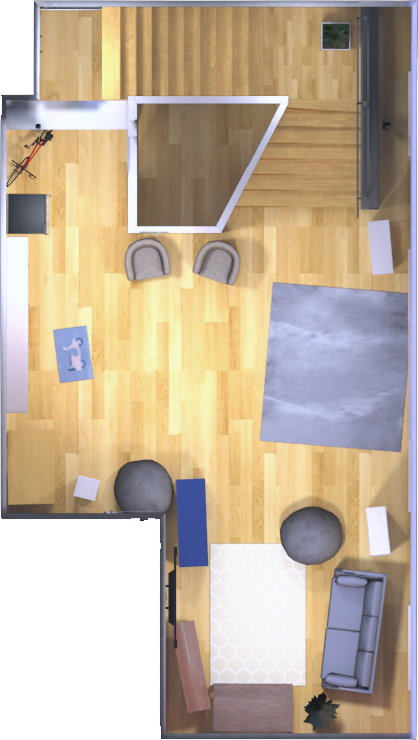}
    \includegraphics[width=0.19\textwidth]{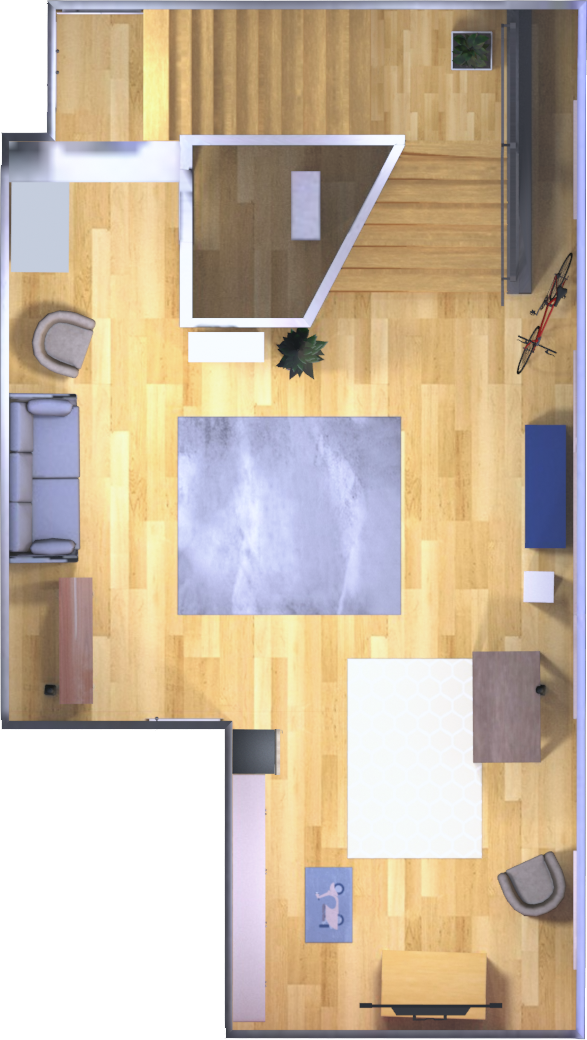}
    \includegraphics[width=0.19\textwidth]{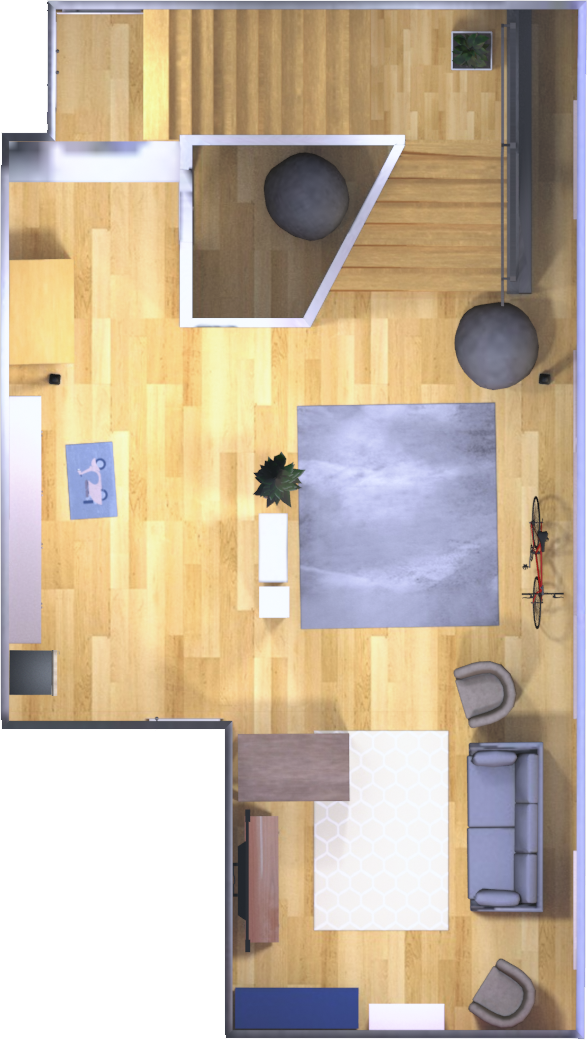}
    \includegraphics[width=0.19\textwidth]{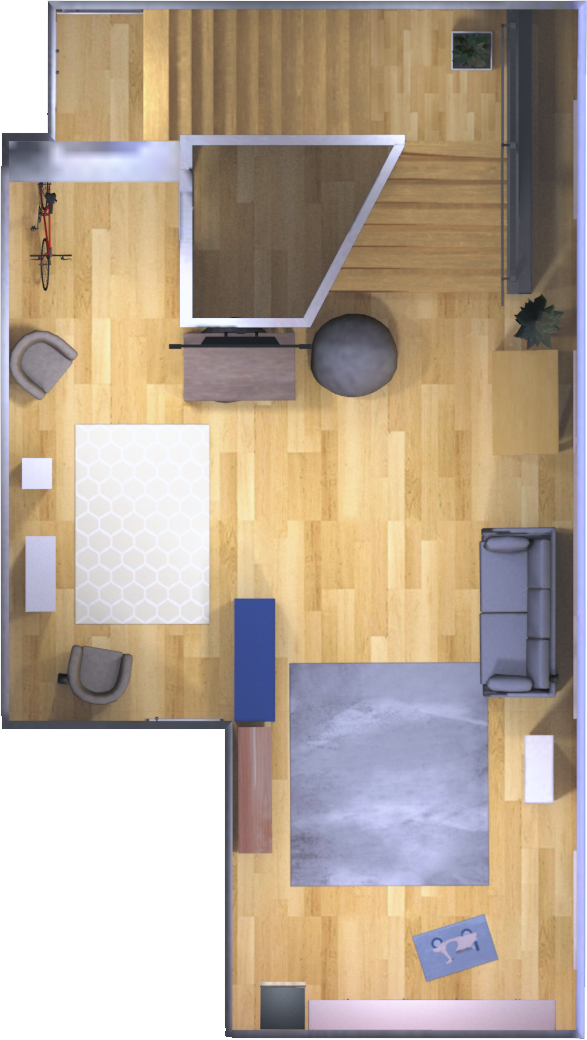}
    \includegraphics[width=0.19\textwidth]{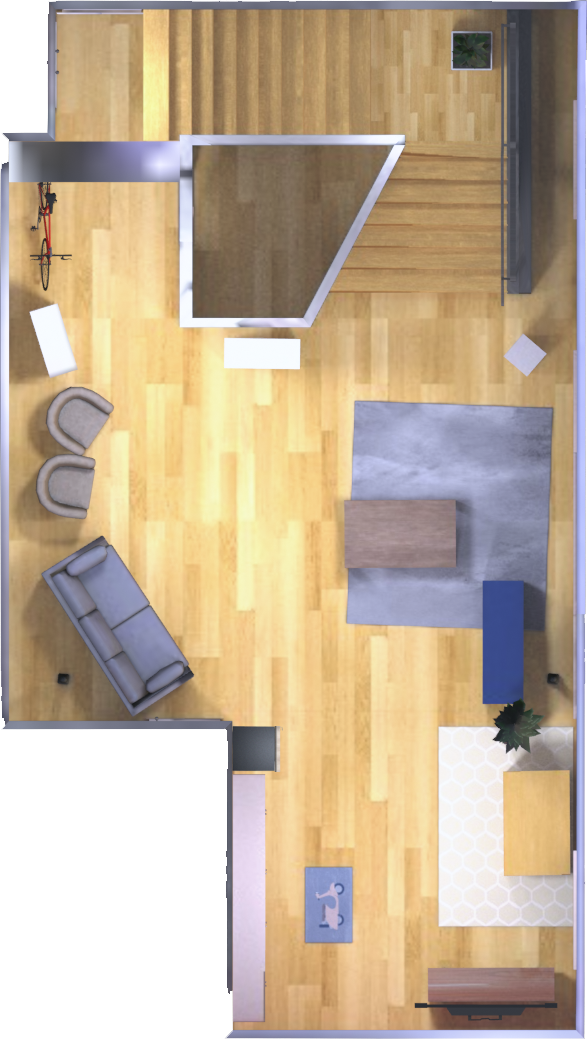}
    \caption{
      The 5 \replica `macro variations' of semantically plausible configurations of furniture in the apartment space. 
      Objects are procedurally added on furniture and surfaces using the annotated supporting surface and containment volume information provided by \replica.
    }
    \label{fig:replica_macro}
\end{figure}

Top down views of the 5 `macro variations' of the scenes are shown in \reffig{fig:replica_macro}.
These variations are 5 semantically plausible configurations of furniture in the space generated by a 3D artist. 
Each surface is annotated with a bounding box, enabling procedural placement of objects on the surfaces.
For each of these 5 variations, we generate 20 additional variations, giving 105 scene layouts.

\section{\monolithic Details}
\label{sec:supp:mono}

\subsection{Architecture}
\begin{figure}
    \centering
    \begin{subfigure}[t]{0.9\linewidth}
    	\centering
    	\includegraphics[width=\textwidth]{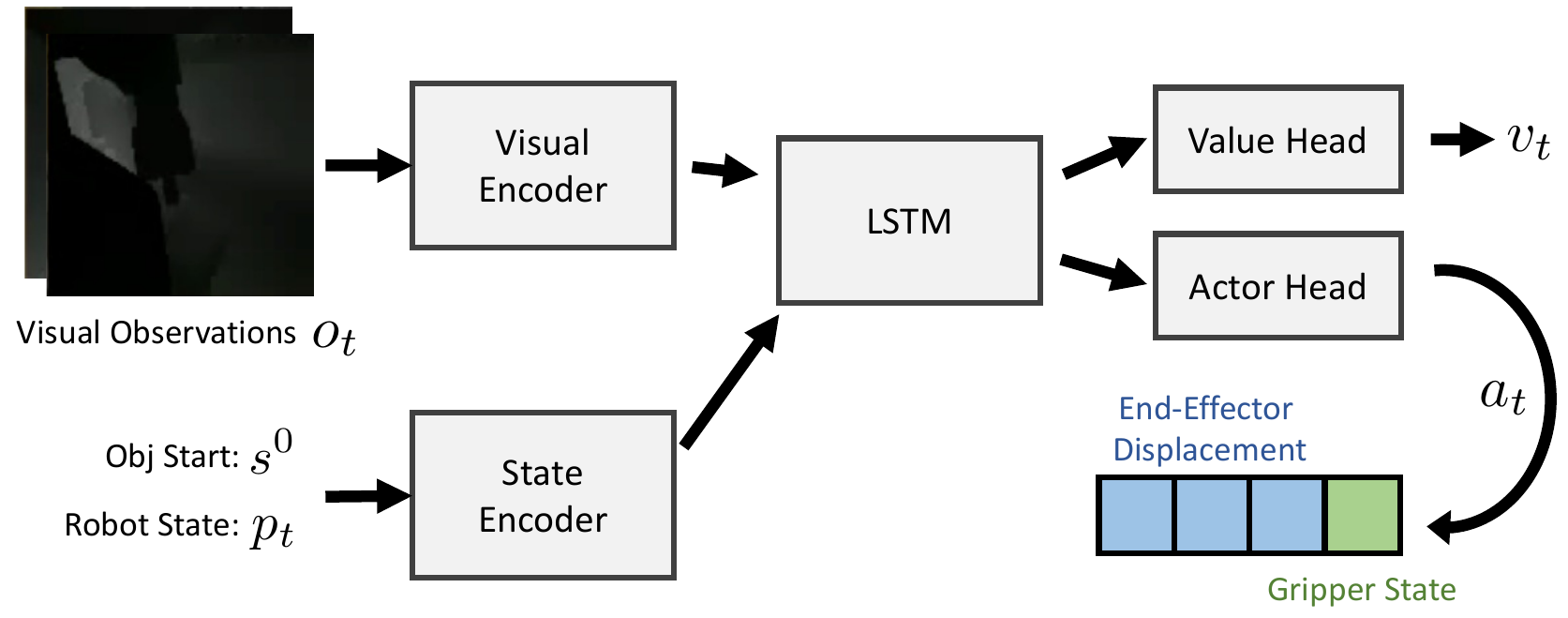}
    \end{subfigure} \\    
    \caption{
      \learnedee policy architecture. The policy maps egocentric visual observations $o_t$, the task-specification in the form of a geometric object goal $s^0$, and the robot proprioceptive state $p_t$ into an action $a_t$ which controls the arm and gripper. A value output is also learned for the PPO update. 
    }
    \label{fig:mono_method}
\end{figure}

The \monolithic architecture consists of a visual encoder which takes as input the egocentric visual observation and a state encoder neural network which takes as input the object start position and the current proprioceptive robot state. 
Both the image and state inputs are normalized using a per-channel moving average. 
$RGB$ and $D$ input modalities are fused by stacking them on top of each other.
These two encodings are passed into an LSTM module which are then processed by an actor head to produce the end-effector and gripper state actions and a value head to produce a value estimate. 
The agent architecture is illustrated in \figref{fig:mono_method}. 

\subsection{Training}
\label{sec:supp:mono_training} 
The agent is trained with the following reward function
\begin{align*}
  r_t = 20\mathbb{I}_{success} + 5\mathbb{I}_{pickup} + 20\Delta_{arm}^{o} \mathbb{I}_{!holding} + 20\Delta_{arm}^{r}\mathbb{I}_{holding} - \max(0.001 C_t, 1.0)
\end{align*} 
Where $\mathbb{I}_{holding}$ is the indicator if the robot is holding an object, 
$\mathbb{I}_{success}$ is the indicator for success, 
$\mathbb{I}_{pickup}$ is the indicator if the agent just picked up the object,  
$\Delta_{arm}^{o}$ is the change in Euclidean distance between the end-effector and target object (if $d_t$ is the distance between the two at timestep $t$, then $\Delta_{arm}^o = d_{t-1} - d_t$), 
and $\Delta_{arm}^{r}$ is the change in distance between the arm and arm resting position. 
$C_t$ is the collision force in Newtons at time $t$.

We train using the DDPPO algorithm \cite{wijmans2019dd} with 16 concurrent processes per GPU across 4 GPUs for 64 processes in total with a preemption threshold of 60\%. 
For the PPO \cite{schulman2017proximal} hyperparameters, we use a value loss coefficient of $0.1$, entropy loss coefficient of $0.0001$, $2$ mini-batches, $2$ epochs over the data per update, and a clipping parameter of $0.2$
We use the Adam \cite{kingma2014adam} with a learning rate of $0.0001$. 
We also clip gradient norms above a magnitude of $0.5$. 
We train for 100M steps of experience and linearly decay the learning rate over the course of training. 
We train on machines using the following NVIDIA GPUs: Titan Xp, 2080 Ti, RTX 6000.

\section{Motion Planning}
\label{sec:supp:mp}

\begin{figure*}[t!]
    \centering
    \begin{subfigure}[t]{0.8\linewidth}
    	\centering
    	\includegraphics[width=\textwidth]{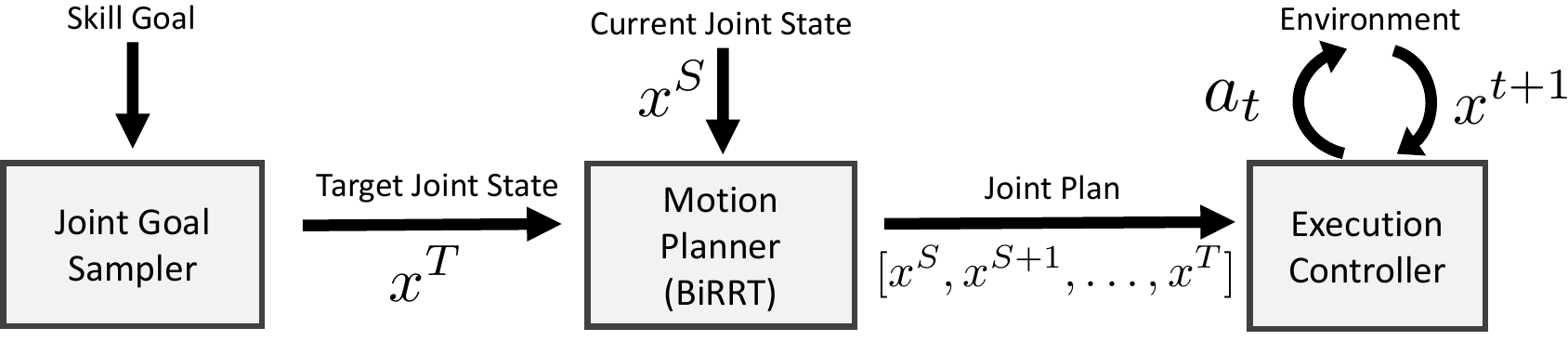}
    \end{subfigure} \\    
    \caption{
      The three stages of our robotics pipeline for \SPAoracle and \SPA.  Starting from a high-level objective such as picking a certain object, the ``Joint Goal Sampler" produces the necessary goal for the motion planner to plan to based on random sampling and inverse-kinematics.  The motion planner then plans a path in joint space from the current joint angles to the desired joint angles.  The executor then translates the motion planner into torque actions for the robot motors.
    }
    \label{fig:mp}
\end{figure*}

In this section, we provide details on our motion planning based sub-task policies that can be composed together to solve the overall task analogous to the Learned policy. 
These approaches employ a more traditional non-learning based robotics pipeline \cite{coleman2014reducing}.
Our pipeline consists of three stages: joint goal sampling, motion planning, and execution as illustrated in Figure~\ref{fig:mp}. 

We exclusively use the sampling-based algorithm RRTConnect~\cite{kuffner2000rrt} (bidirectional rapidly-exploring random tree) as the motion planner given that it is one of the state-of-the-art methods that the robotics literature frequently builds on and compares to~\cite{kuwata2008motion,ratliff2009chomp,schulman2014motion,hernandez2016team,mukadam2018continuous,ichter2018learning,hou2020posterior,islam2020alternative,pantic2021mesh} and for which a well maintained open source implementation is available in the OMPL library~\cite{sucan2012open} (open motion planning library). Since it does not employ learning, it also serves as a stand-in for a more traditional non-learning based robotics pipeline.

Our aim with the current baselines is to demonstrate a strong starting point and our hope is that it drives adoption within the robotics community to develop and benchmark their algorithms, learning based or otherwise on this platform.
For instance, work in the area of motion planning has made several advancements with new sampling techniques~\cite{gammell2014informed,gammell2015batch} and optimization based methods~\cite{ratliff2009chomp,schulman2014motion,kappler2018real,mukadam2018continuous}, but largely operated on the assumption of a reliable perception stack. However, difficulty in obtaining maintained open source implementations that are not tied to a specific hardware or have complex dependencies like ROS~\cite{quigley2009ros} have also posed challenges in bringing the vision and robotics communities together under a common set of tasks. More recent work has however begun utilizing learning and transitioning towards hybrid methods, for example learning distributions for sampling~\cite{ichter2018learning}, using reinforcement~\cite{faust2018prm} or differentiating through the optimization~\cite{bhardwaj2020differentiable}. 

We implement two variants that defer in how they handle perception: one that uses privileged information from the simulator (\SPAoracle) and one that uses egocentric sensor observations (\SPA). 
\mpvis uses depth sensor to obtain a 3D point cloud in the workspace of the robot at the measurement instance which is used for collision checking. 
Since the arm can get in the way of the depth measurement, the arm optionally lowers so the head camera on the Fetch robot can sense the entire workspace. 
If it is not possible to lower the camera (as in the case of holding an object), the detected points consisting of the robot's arm are filtered out and detected points from prior robot positions and orientations are accumulated (which is possible since we have perfect localization). 
\SPAoracle on the other hand directly accesses the ground truth scene geometry for collision checking. 
\SPAoracle plans in an identical Habitat simulator instance as the current scene by directly setting the state and checking for collisions using the duplicate Habitat simulator instance. 
When the robot is holding an object, \SPAoracle updates the position of the held object based on the current joint states for collision checking in planning. 
The full, not simplified robot model is used for collision checking. 

A wrapper exposes a Habitat or PyBullet simulator instance to OMPL to perform the motion planning. 
Specifically, this exposes a collision check function based on a set of Fetch robot arm joint angles. 
Sampling is constrained to the valid joint angles of the Fetch robot. 

Motion planning is used as a component in performing skills. 
At a high-level many skills repeat the same steps. 
First, determine the specific goal as a target joint state of the robot arm for the planner based off the desired interaction of the skill.
This could be a grasp point for picking an object up, a valid position of the arm to drop an object, a position of the arm which can grasp the handle, etc.  
A combination of IK, random sampling and collision checks, are used to solve this step. 
Next, a planning algorithm from OMPL is invoked to find a valid sequence of arm joint states to achieve the goal. 
Finally, the robot executes the plan by consecutively setting joint motor position targets based on the planned joint positions. 

\begin{itemize}
    \item \pick: First, sample a grasp position on the object. 
      Sample points in the graspable sphere around the object, in our experiments a sphere with a radius of $15$cm. 
      Filter out points which are closer to another object than the desired object to pick. 
      Use IK to find a joint pose which reaches the sampled point. 
      Next, check if the robot in the calculated joint pose is collision free and if so return this as the desired joint state target. 
    This grasp planning produces a desired arm joint state, now use BiRRT to solve the planning problem. 
    After the plan is executed with kinematic control for \SPAoracle or PD control for \SPA, execute the grasp action. 
    After grasping the object, the robot plans a path back to the arm resting position, using the stored joint states of the resting arm as the target.
    \item \place: The same as Pick, but now sample a goal position as a joint state which has the object at the target. For \SPAoracle, this uses the exact object model. For \SPA this uses a heuristic distance of the gripper to the desired object placement. 
\end{itemize}

We use a 30 second timeout for the planning. 
A step size of 0.1 radians is used for the step size in the RRTConnect algorithm. 
All planning is run on a machine using a Intel(R) Core(TM) i9-9900X CPU @ 3.50GHz. 

\section{Pick Task Further Analysis Experiments}

\subsection{Blind Policy Analysis}
\label{sec:blind_policy_analysis} 
\begin{wrapfigure}{r}{0.4\textwidth}
  \vspace{-10pt}
  \includegraphics[width=0.4\textwidth]{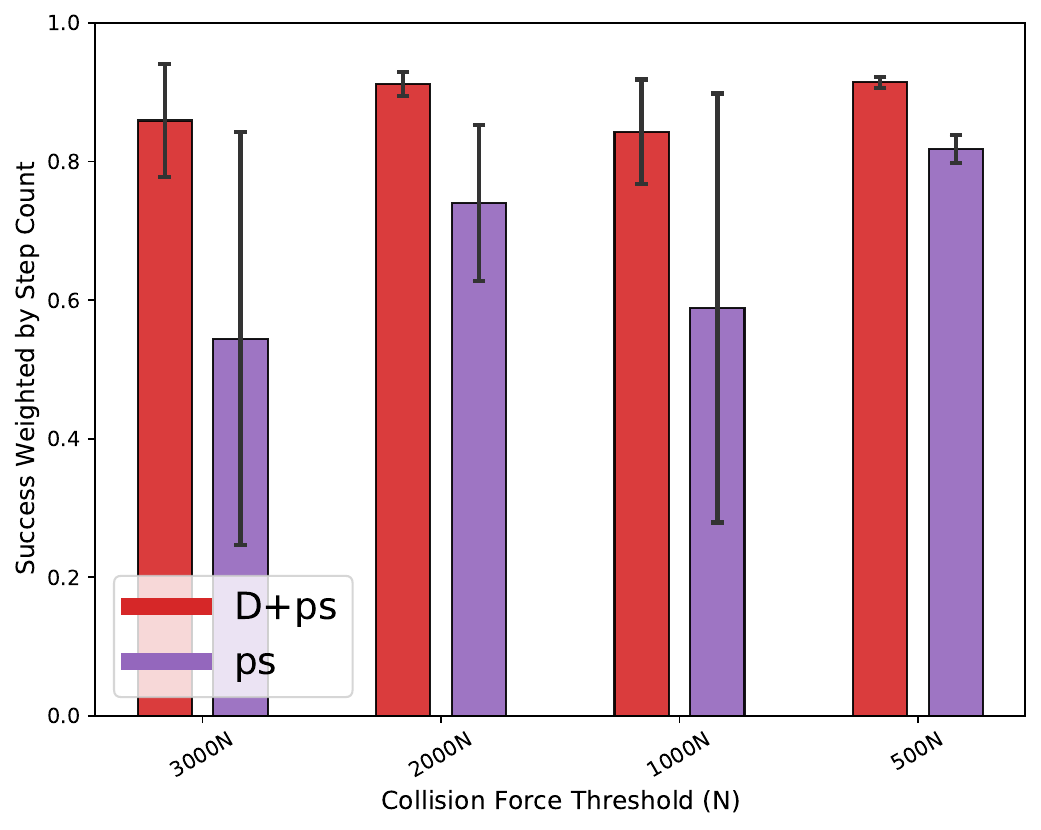}
  \caption{
    Path efficiency for sighted vs blind policies vs amount of collision allowed ($N$=$3$).%
  }
  \label{fig:mod_coll_pick_analysis}
  \vspace{-10pt}
\end{wrapfigure}

To further investigate the hypothesis that the blind `feels its way' to the goal, we analyze how \emph{efficient} the two are at picking up objects, 
using the Success weighted by Completion Time (SCT) metric~\cite{yokoyama2021success}. 
Specifically, $SCT = \text{Success} \cdot (\nicefrac{\text{time taken by agent}}{\text{time taken by oracle}})$. 
We use an upper-bound on the oracle-time: $\nicefrac{\text{2*Euclidean distance(end-effector, goal)}}{\text{maximum speed of end-effector}}$. 
For ease of analysis, we use a simplified Pick setting with only the `left counter' receptacle.
The robot starts in front of the counter receptacle facing the wall.
$N(0,50)$cm is added to both the $x,y$ position of the base, $ \mathcal{N}(0,0.15)$ radians is added to the base orientation, $ \mathcal{N}(0,5)$cm is added to the $x,y,z$ of the starting end-effector position.
5 objects are randomly placed on the counter from the `food' or `kitchen' item categories of the YCB dataset.
One of the objects is randomly selected to be the target object.

\Cref{fig:mod_coll_pick_analysis} shows the SCT (on unseen layouts) as a function of the collision-force threshold used during training for policies trained for 100M steps. 
We find that sighted policies (Depth) are remarkably efficient across the board, achieving over $80\%$ SCT. 
Since we use a crude upper-bound on the oracle time it is unclear if a greater SCT is possible.
The sighted policies may be discovering nearly maximally efficient trajectories, 
which would be consistent with known results in navigation~\cite{wijmans2019dd}. 
The collision threshold is not related to performance, since the collision threshold is also used in training and will affect training. %
Very low collision thresholds result in conservative policies which avoid any hard collisions with objects and succeed more. 
Blind policies are significantly less efficient and \emph{improve} in efficiency as the allowed collision threshold is reduced

\subsection{Camera Placement: Arm cameras are most useful; Suggestive evidence for self-tracking}
\label{sec:cam_placement} 
One advantage of fast simulation is that it lets us study 
robot designs that may be expensive or even \emph{impossible} 
to construct in hardware. 
We use the same experimental settings as \secref{sec:sensor_ablate}, training the policies to pick objects from 8 receptacles (receptacles depicted in \figref{supp:fig:receptacles}). 
`Arm' and `Head' placements were already described in \secref{sec:agent}.  
`3rdPoV' is a physically-implausible camera placement with a view from over the robot's shoulder 
(commonly used in video games and image-based RL papers \eg \cite{kostrikov_iclr2021}). 
`Invisible Arm' is a physically-impossible setting where the robot's arm is physically present 
and interacts with the scene but is not visible in the cameras. %
\begin{figure}
    \centering
    \includegraphics[width=0.45\textwidth]{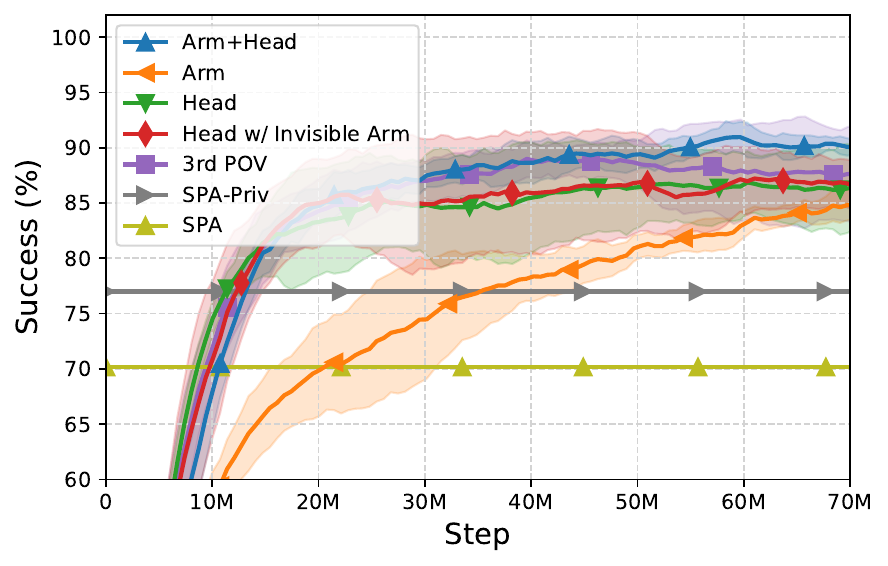}
    \caption{
    Camera placement analysis: Success rates on unseen layouts ($N$=$500$) 
     vs training steps. Mean and std-dev over 3 training runs. 
    }
    \label{fig:pick_cam_analysis}
\end{figure}

\figref{fig:pick_cam_analysis} shows performance on unseen layouts (vs training steps) for different camera placements on Fetch. 
While all camera placements perform generally well ($80$-$90$\% success), 
The combination of head and arm camera performs best (at $92\%$ success).  
The arm only camera performs the worst, being slower to learn and only ultimately achieving $85\%$ success rate.

\subsection{Emergence of Self-Tracking}
\label{sec:supp:self_tracking}
\begin{wrapfigure}{r}{0.43\textwidth}
    \includegraphics[width=0.45\textwidth]{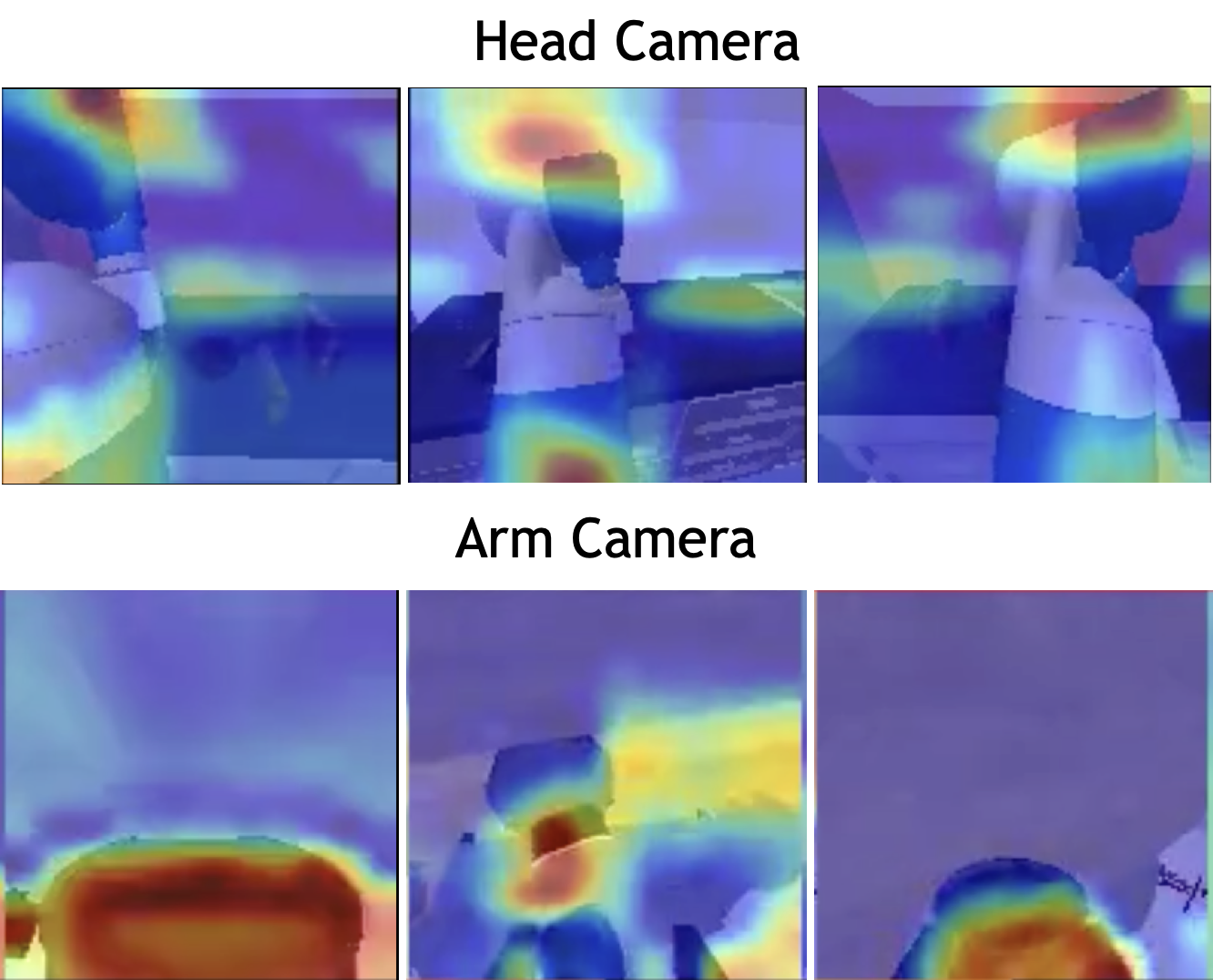}
    \caption{
    Grad-CAM saliency maps for three different scenes from cameras mounted on the Head and the Arm. 
    Notice that the arm-joints are considered particularly salient in both cases across scenes.  
    }
    \label{fig:self_tracking}
\end{wrapfigure}

In order to qualitatively analyze the performance of the Pick policies, we visually interpret the saliency of the trained policy via Grad-CAM maps~\cite{selvaraju_ICCV17} computed with respect to the actions of the robot.
To generate these Grad-CAM heatmaps, we follow the protocol laid down in Grad-CAM~\cite{selvaraju_ICCV17} and compute the gradient of each of the four continuous action values (‘displacement’ in three directions and ‘grab’) with respect to the activations of the final convolutional layer of the visual encoder. 
Subsequently, we average the heatmaps for each of the ‘displacement’ actions to give us an overall sense of saliency for the robot’s displacement based on the input at each step, and perform the required post-processing to overlay this on top of the input frame. 
\reffig{fig:self_tracking} shows the overall displacement maps for the robot in three different scenes and demonstrates the emergence of self-tracking behavior.
In different scenes from cameras mounted on the Head as well as the Arm, we find a consistent trend that the maps highlight arm joints suggesting that the agent has learned to track the arm.

Caveat: we stress that saliency maps and the act of 
drawing inferences from them are fraught with a host of 
problems (see \cite{Atrey2020Exploratory, SanityNIPS2018} for excellent discussions). 
This analysis should be considered 
a speculative starting point for further investigation and not a finding it itself.

\subsection{Effect of Time-Delay on Performance}
\label{sec:supp:time_delay}
\begin{figure}
    \centering
    \includegraphics[width=0.5\linewidth]{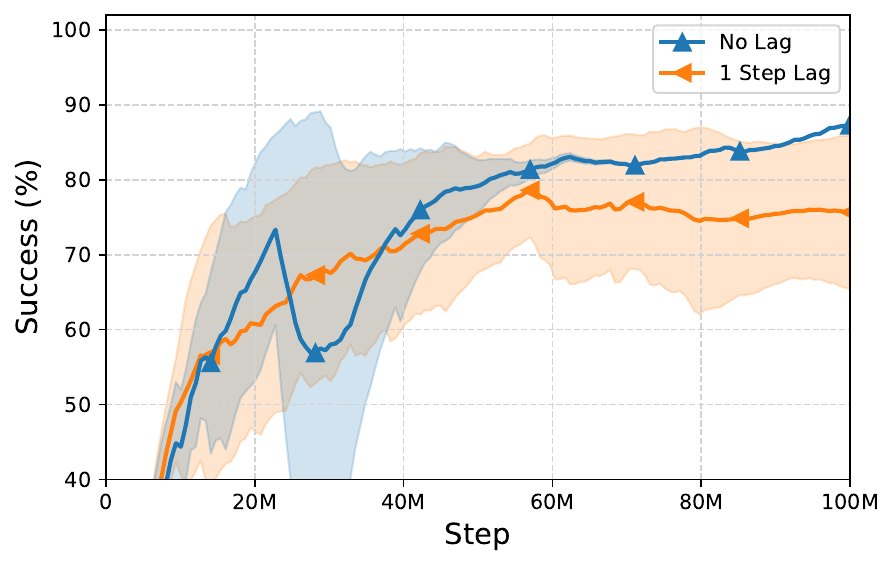}
    \caption{
      Effect of the time-delay on performance on the picking skill. Averages and standard deviations across 3 seeds. 
      1-step has high-variance results which could be reduced with more seeds.
    }
    \label{supp:fig:lag_analysis}
\end{figure}

We studied the effect the time delay in \reffig{supp:fig:lag_analysis} in the same experimental setting as \Cref{sec:blind_policy_analysis} and find that the time delay has a minimal impact on performance. 
The 1-step delayed time has large variance which could be reduced through more seeds.

\subsection{Action Space Analysis}
\label{supp:sec:agent}
\begin{figure}
    \centering
    \includegraphics[width=0.5\linewidth]{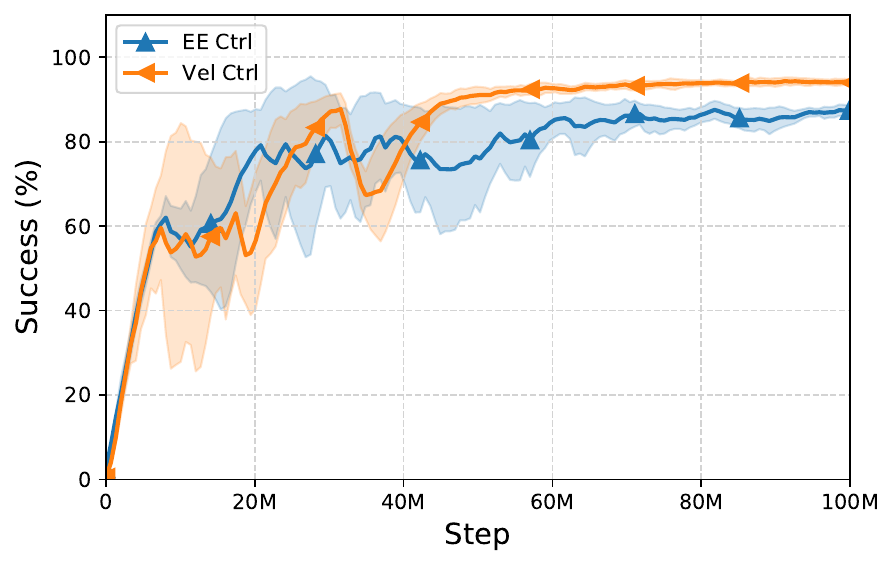}
    \caption{
      Comparison of end-effector and velocity control for the picking skill. 
      Averages and standard deviations across 3 seeds.
      Both end-effector and velocity control are able to solve the task.
    }
    \label{supp:fig:ctrl_analysis}
\end{figure}

Action spaces other than end-effector control are possible in \habname. 
We compare end-effector versus velocity control in the Pick skill in \Cref{supp:fig:ctrl_analysis} in the same experimental setting as \Cref{sec:blind_policy_analysis}. 
For velocity control, the policy outputs a 7 dimension vector representing the relative displacement of the position target for the PD controller. 
Despite, this higher dimension action space, velocity control learns just as well as end-effector control for the picking skill. 

\section{\tasksuitenamefull Experimental Setup Details}
\label{sec:supp:hab_task}

\begin{figure}
    \centering
    \begin{subfigure}[t]{0.32\textwidth}
        \includegraphics[width=\textwidth]{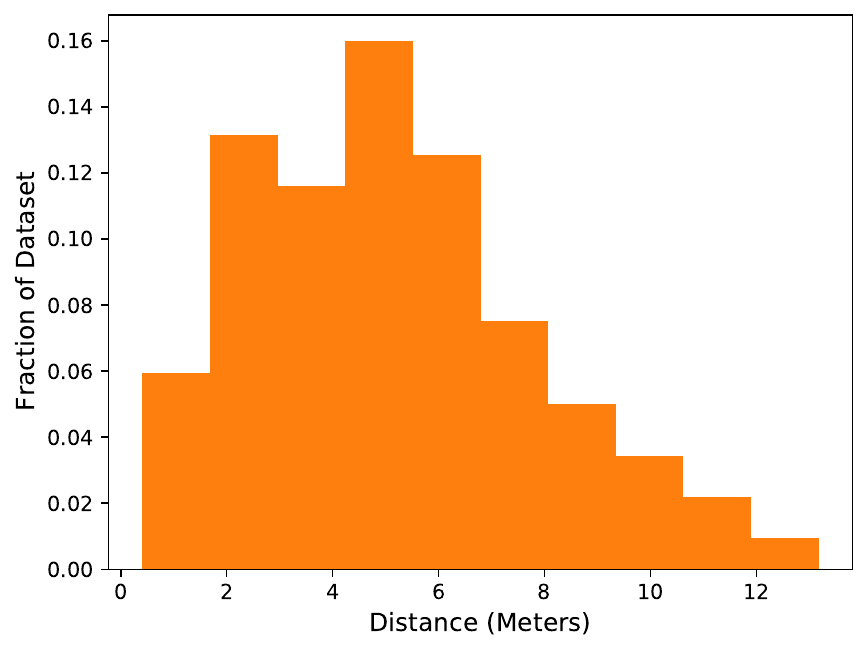}
        \caption{Start Distance}
    \end{subfigure}
    \begin{subfigure}[t]{0.32\textwidth}
        \includegraphics[width=\textwidth]{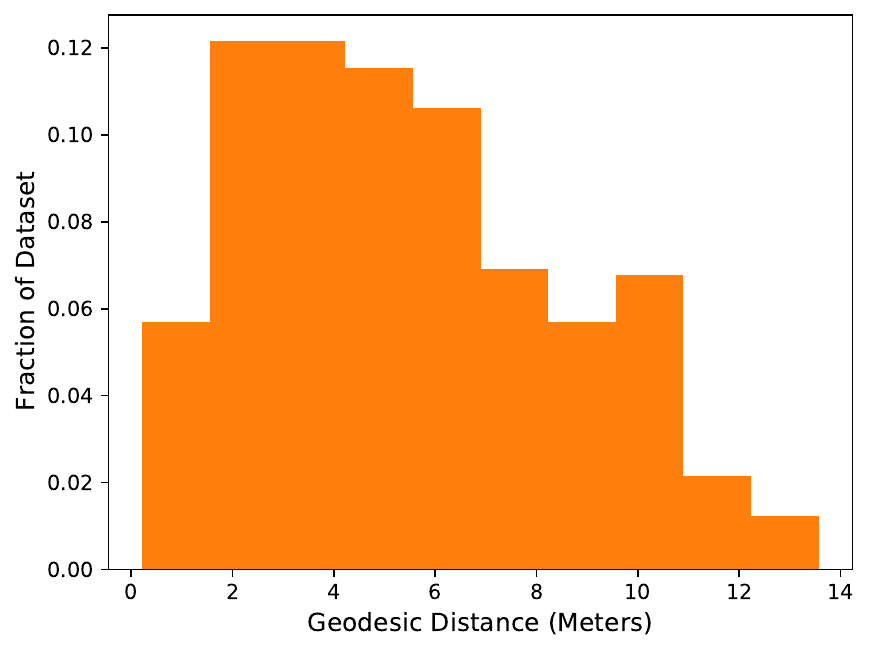}
        \caption{Start Distance (Geodesic)}
    \end{subfigure}
    \begin{subfigure}[t]{0.32\textwidth}
        \includegraphics[width=\textwidth]{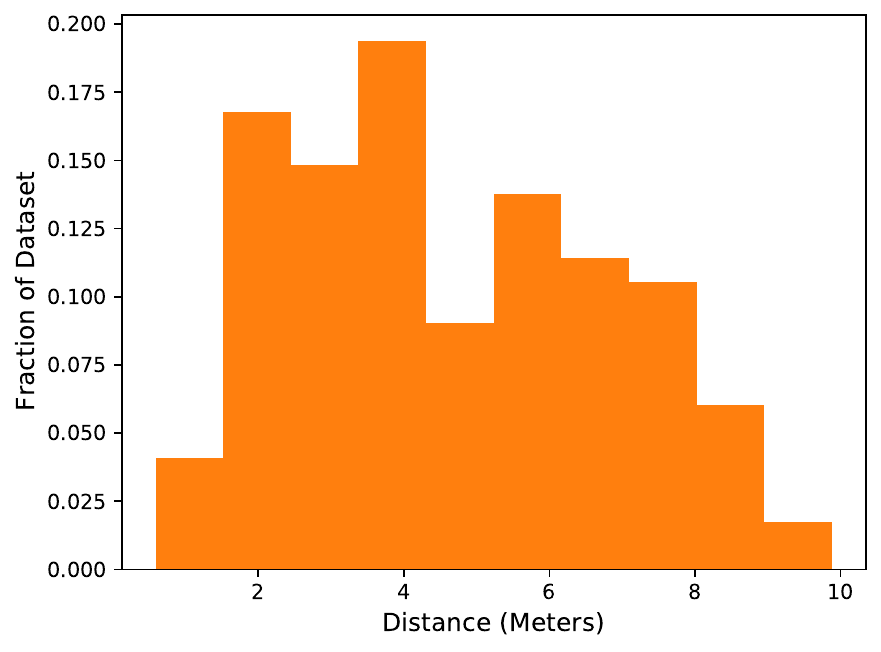}
        \caption{Goal Distance}
    \end{subfigure}
    \caption{ \cleanhouse Rearrangement dataset statistics.  }
    \label{fig:rearrang_dataset:clean_house}
\end{figure}

\begin{figure}
    \centering
    \begin{subfigure}[t]{0.32\textwidth}
        \includegraphics[width=\textwidth]{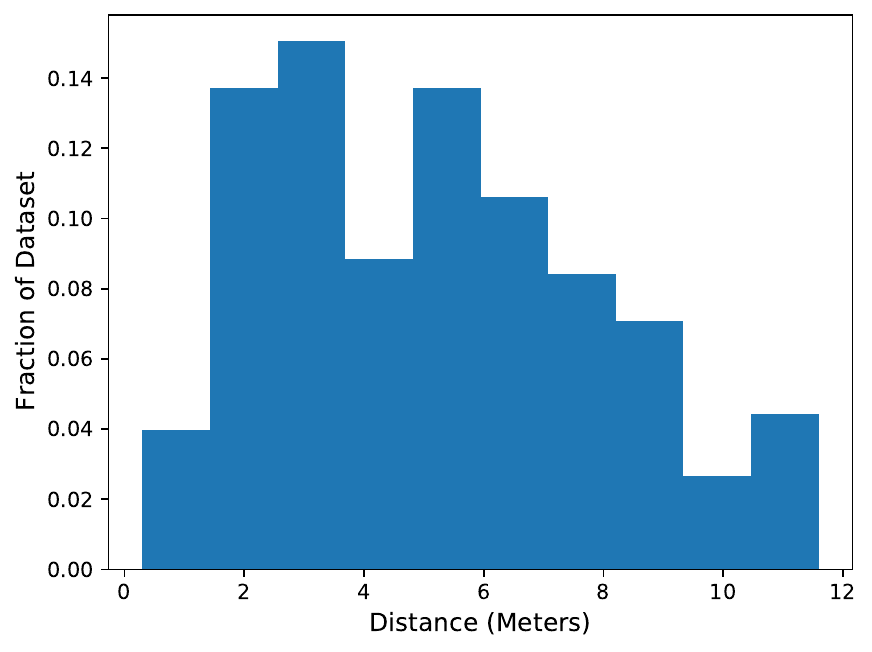}
        \caption{Start Distance}
    \end{subfigure}
    \begin{subfigure}[t]{0.32\textwidth}
        \includegraphics[width=\textwidth]{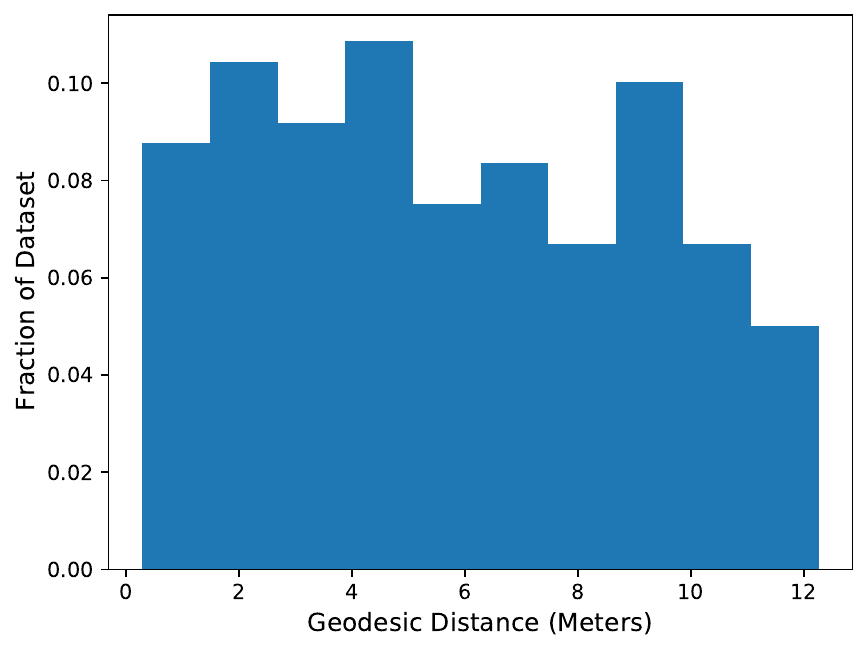}
        \caption{Start Distance (Geodesic)}
    \end{subfigure}
    \begin{subfigure}[t]{0.32\textwidth}
        \includegraphics[width=\textwidth]{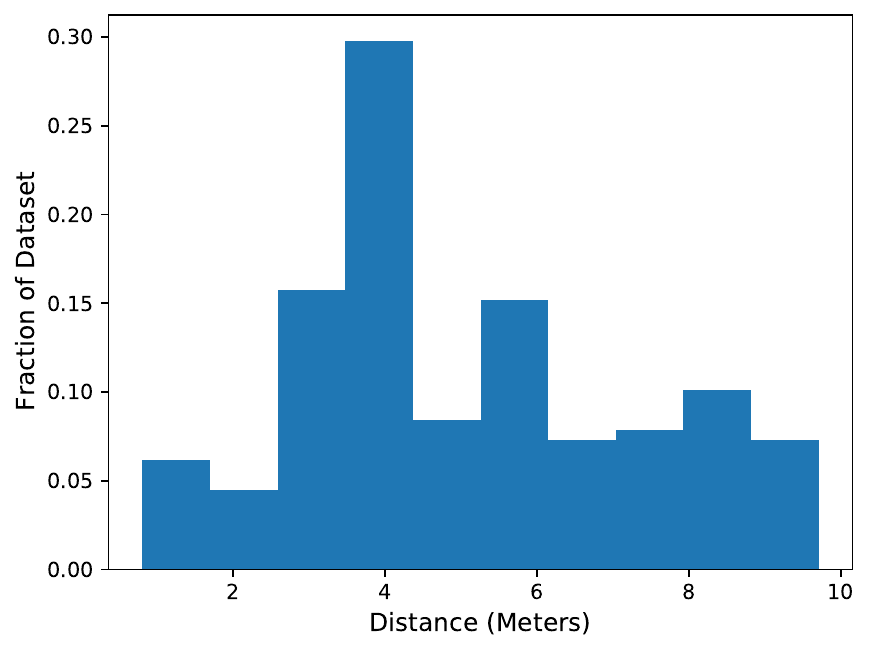}
        \caption{Goal Distance}
    \end{subfigure}
    \caption{
      \settable Rearrangement dataset statistics.
    }
    \label{fig:rearrang_dataset:set_table_fix}
\end{figure}

\begin{figure}
    \centering
    \begin{subfigure}[t]{0.32\textwidth}
        \includegraphics[width=\textwidth]{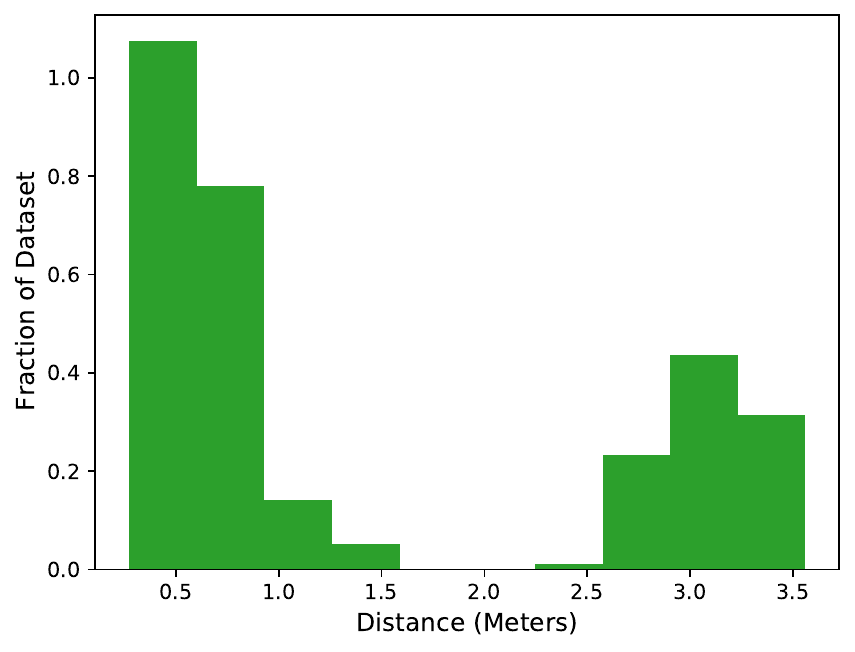}
        \caption{Start Distance}
    \end{subfigure}
    \begin{subfigure}[t]{0.32\textwidth}
        \includegraphics[width=\textwidth]{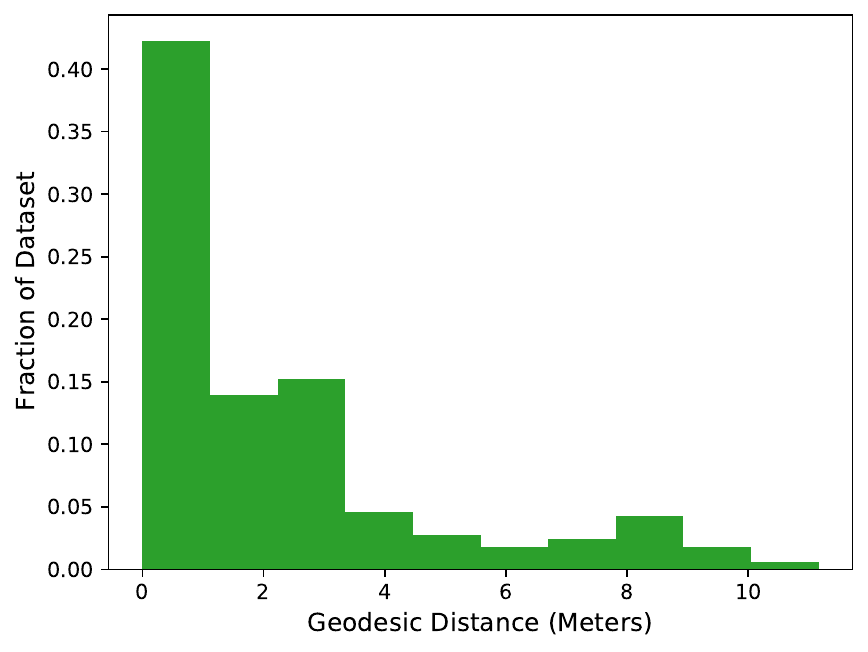}
        \caption{Start Distance (Geodesic)}
    \end{subfigure}
    \begin{subfigure}[t]{0.32\textwidth}
        \includegraphics[width=\textwidth]{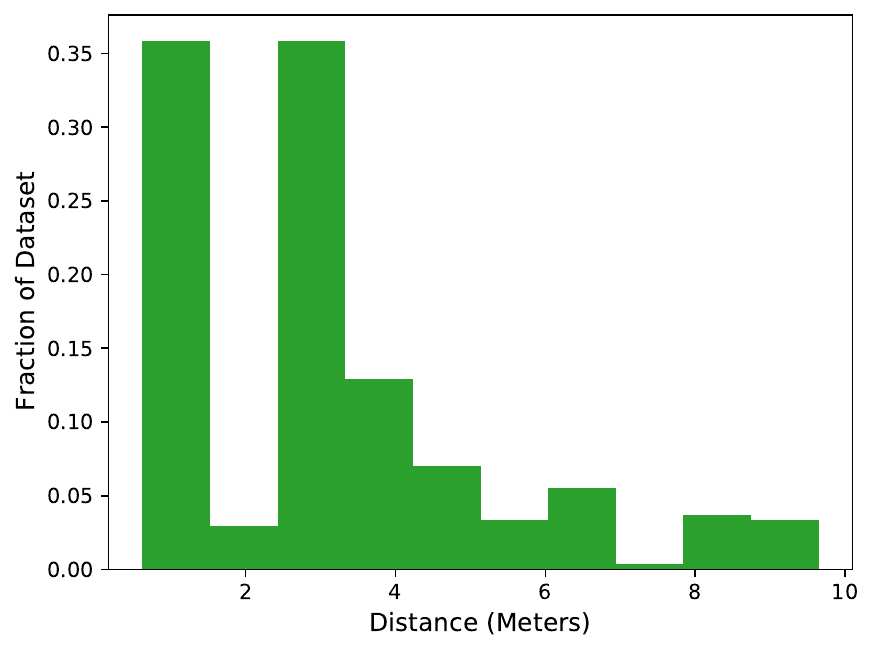}
        \caption{Goal Distance}
    \end{subfigure}
    \caption{
      \stockfridge Rearrangement dataset statistics.
    }
    \label{fig:rearrang_dataset:stock_fridge}
\end{figure}

\subsection{Evaluation}
\label{sec:supp:task_evaluation}
For each task, 100 evaluation episodes are generated. 
These evaluation episodes have unseen micro-variations of the furniture not seen during any training for the learned methods. 
Object positions are randomized between episodes and the robot spawns at a random position in the scene. 
See \Cref{fig:rearrang_dataset:clean_house,fig:rearrang_dataset:stock_fridge,fig:rearrang_dataset:set_table_fix} for rearrangement dataset statistics for the \tasksuitenamefull task definitions.

For each task, success is evaluated based on if all target objects were placed within $15$cm of the goal position for that object, object orientation is not considered. 
To make evaluation easier, there was no collision threshold applied for full task evaluation. 

\subsection{Partial Evaluation}
\label{sec:supp:partial_eval}
Since our tasks are very challenging, we also feature partial evaluation of the tasks up to only a part of the overall rearrangements needed to solve the task. 
These partial task solving rearrangements are listed below, note each rearrangement builds upon the previous rearrangements and the robot must complete each of the previous rearrangements as well.
\begin{itemize}
  \item \cleanhouse: (1) pick object 1, (2) place object 1, (3) pick object 2, etc. 
    Each of the 10 interactions is picking and placing a successive target object.
  \item \stockfridge: (1) pick first fridge object, (2) place first fridge object on counter, (3) pick second fridge object, (4) place second fridge object on table, (5) pick counter object, (6) place counter object in fridge.
    Like \cleanhouse, each of the interactions is picking and placing an object. 
  \item \settable: (1) open the drawer, (2) pick the bowl from the drawer, (3) place the bowl on the table, (4) close the drawer, (5) open the fridge, (6) pick the apple from the fridge, (7) place the apple in the bowl, (8) close the fridge.
\end{itemize}

\section{\tasksuitenamefull Baseline Method Details}
\label{sec:supp:methods}

\subsection{Planner Details}
\label{sec:supp:planner}
\begin{figure*}[t!]
    \centering
    \begin{subfigure}[t]{0.99\linewidth}
    	\centering
      \begin{overpic}[trim=50 0 0 0, clip, width=0.32\textwidth]{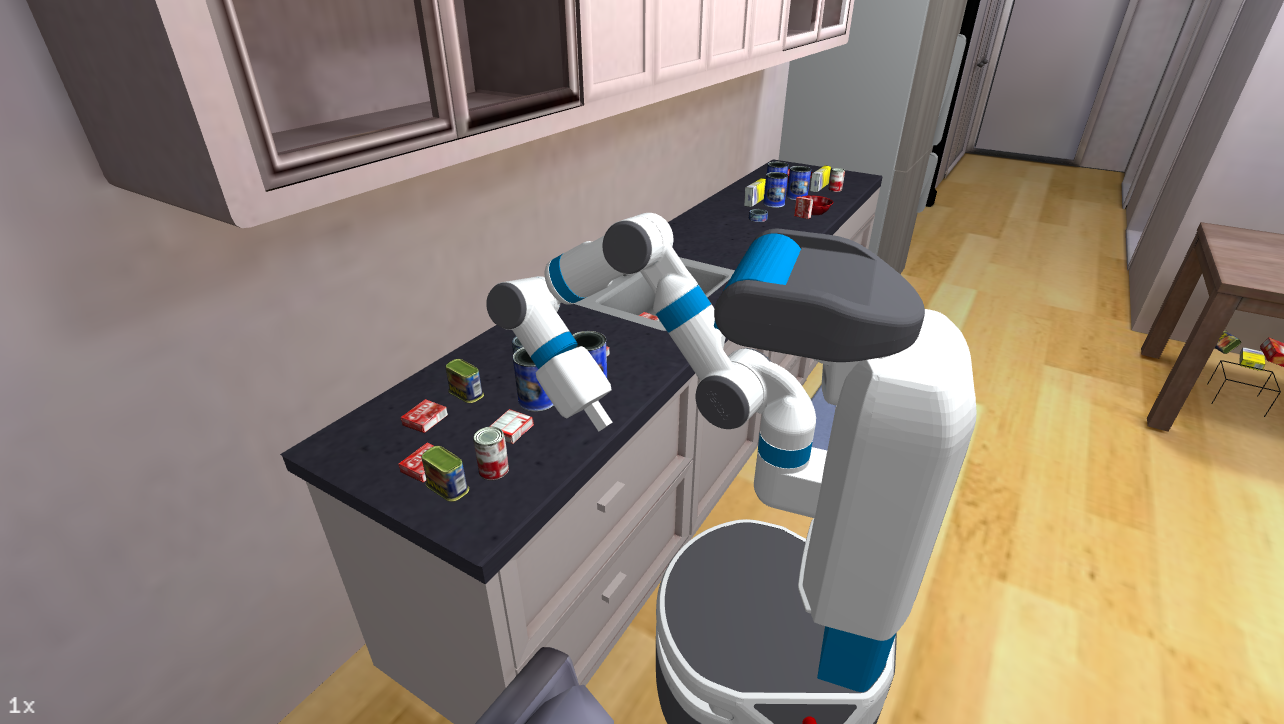}
       \put (0,40) {\color{red} \huge$s^0$}
      \end{overpic}
    	\includegraphics[trim=50 0 0 0, clip, width=0.32\textwidth]{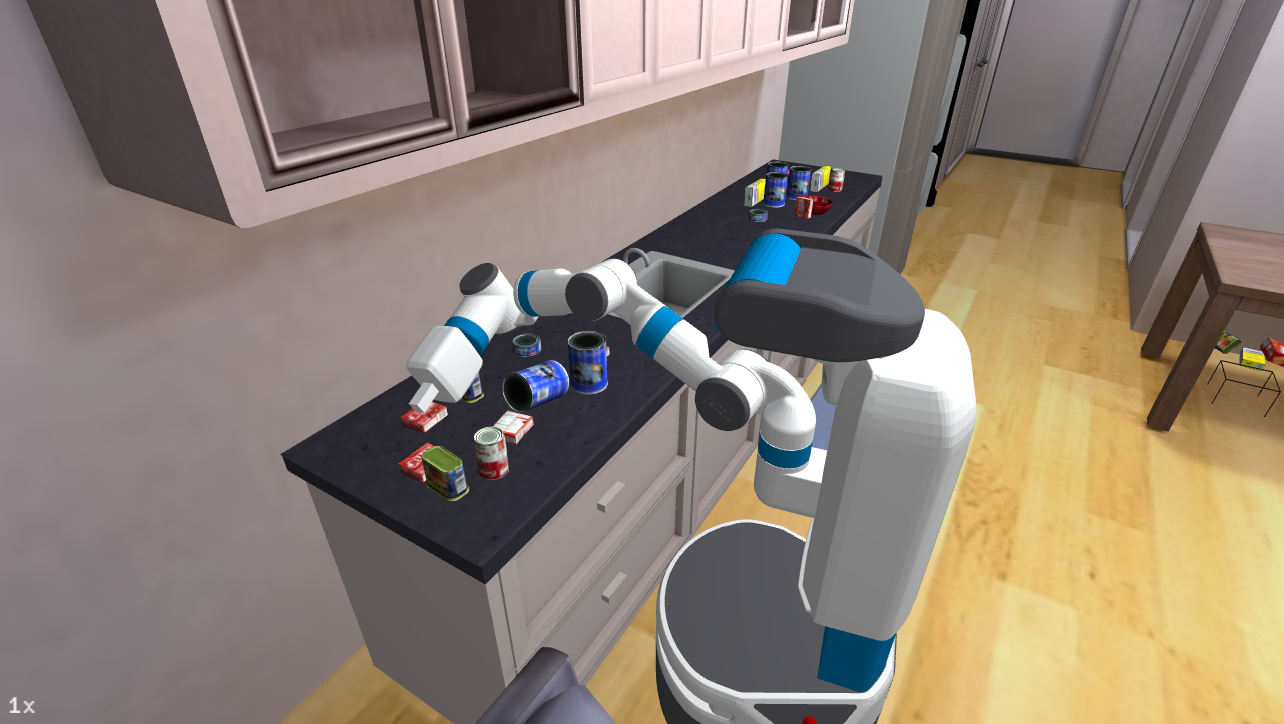}
      \begin{overpic}[trim=50 0 0 0, clip, width=0.32\textwidth]{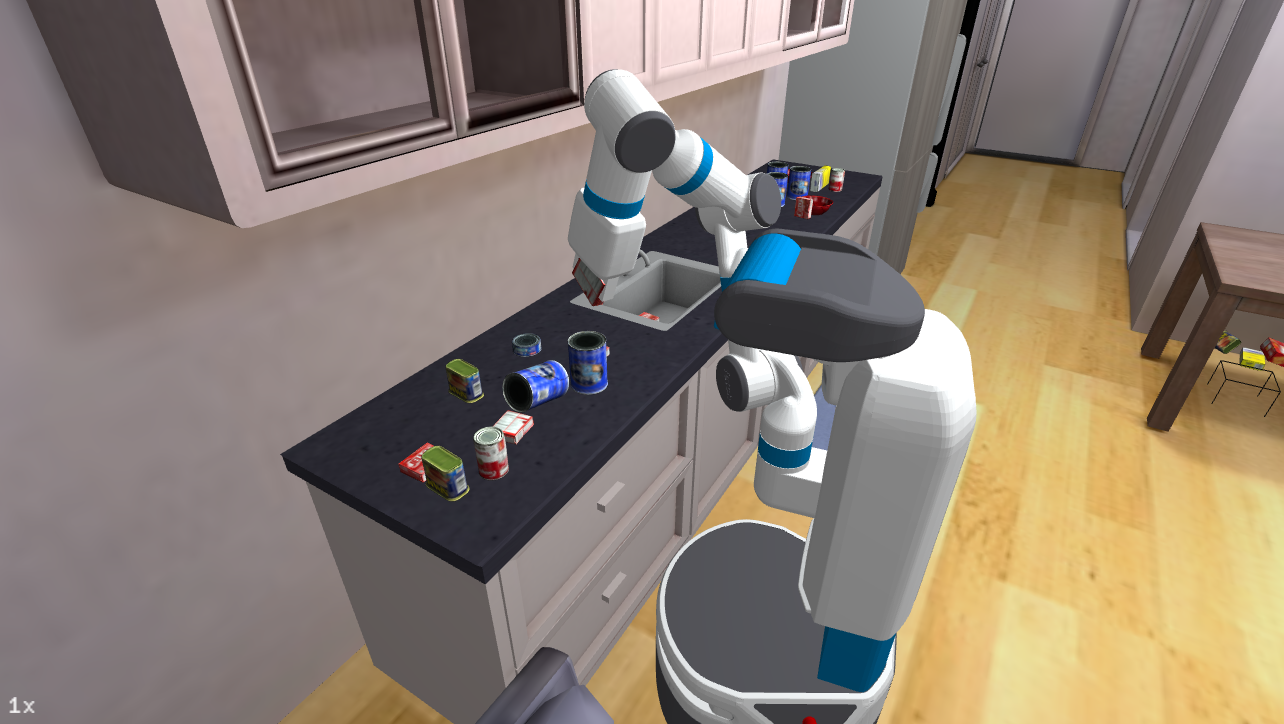}
        \put (0,40) {\color{red} \huge$s^*$}
      \end{overpic}
      \caption{Pick}
      \label{fig:supp:pick}
    \end{subfigure} \\    
    \begin{subfigure}[t]{0.99\linewidth}
    	\centering
      \begin{overpic}[trim=50 0 0 0, clip, width=0.32\textwidth]{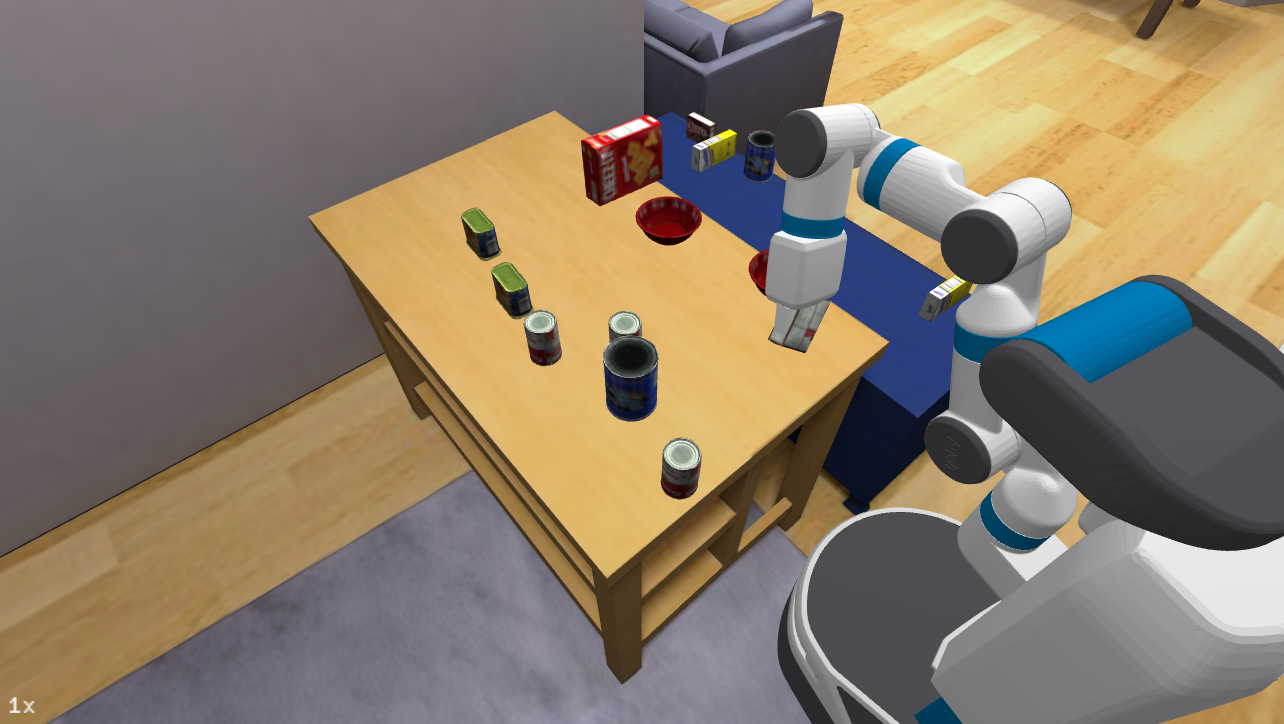}
       \put (0,40) {\color{red} \huge$s^0$}
      \end{overpic}
    	\includegraphics[trim=50 0 0 0, clip, width=0.32\textwidth]{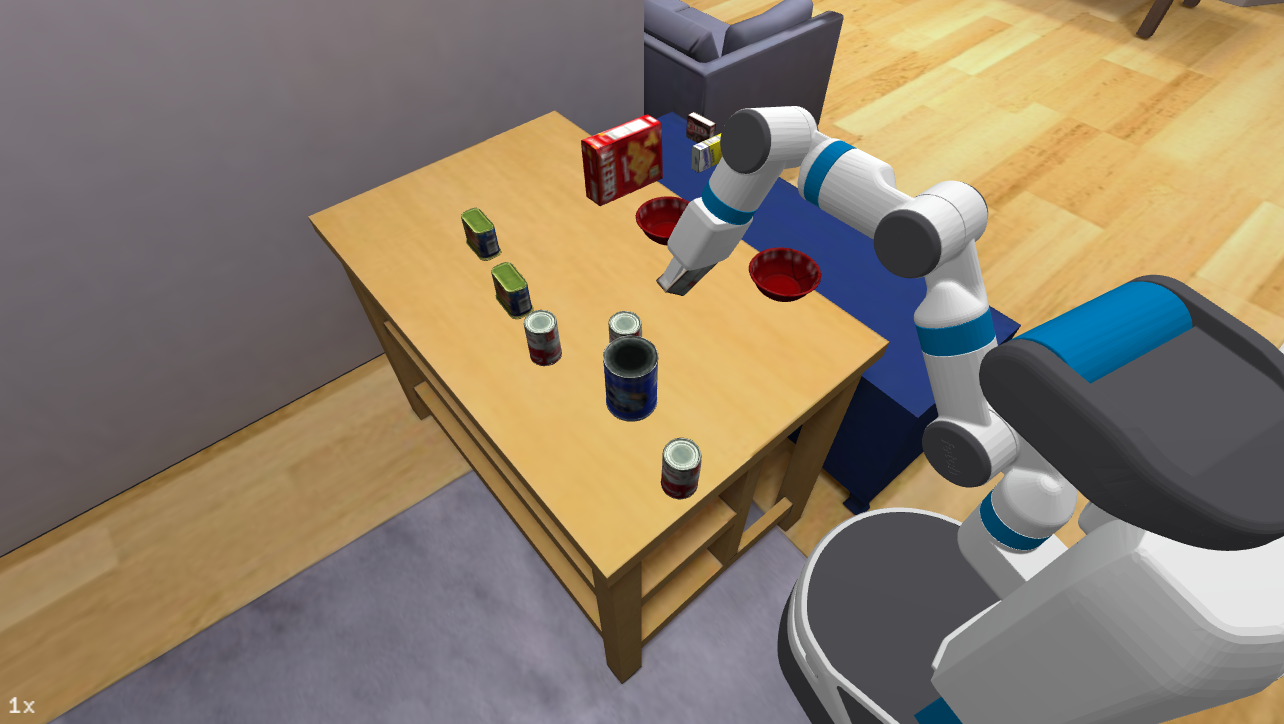}
      \begin{overpic}[trim=50 0 0 0, clip, width=0.32\textwidth]{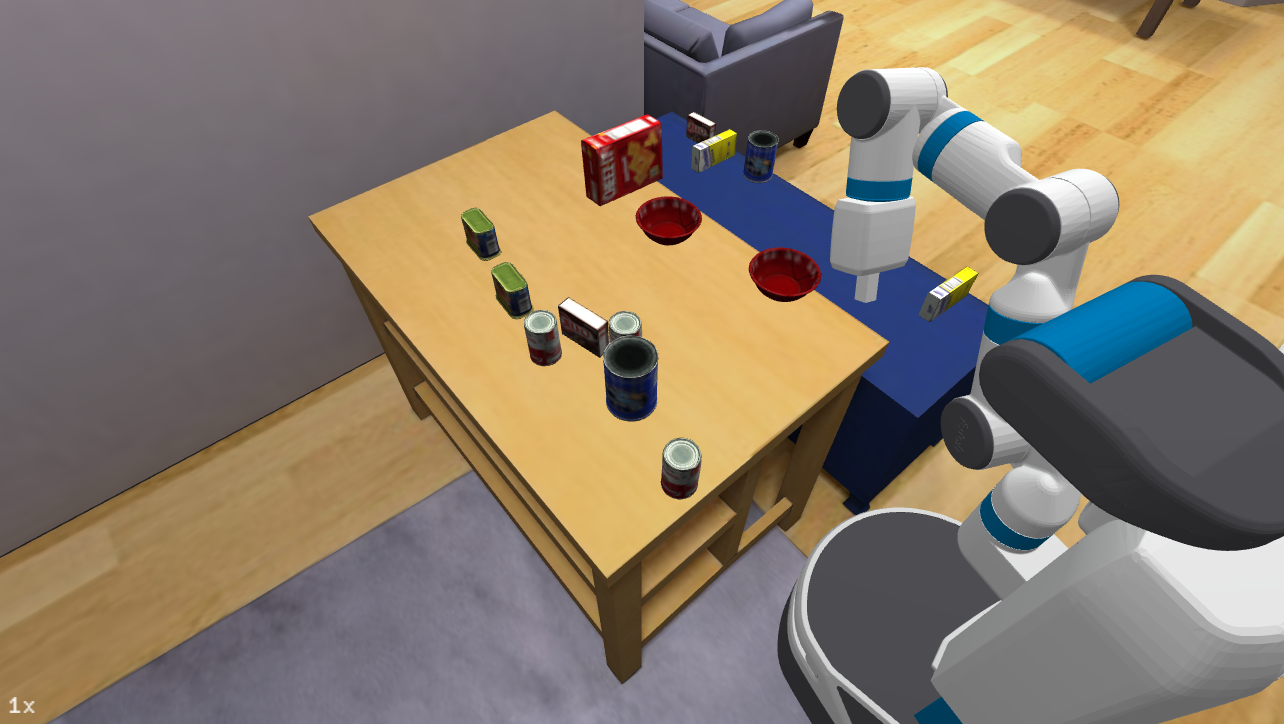}
        \put (0,40) {\color{red} \huge$s^*$}
      \end{overpic}
      \caption{Place}
      \label{fig:supp:place}
    \end{subfigure} \\    
    \begin{subfigure}[t]{0.99\linewidth}
    	\centering
      \begin{overpic}[trim=50 0 0 0, clip, width=0.32\textwidth]{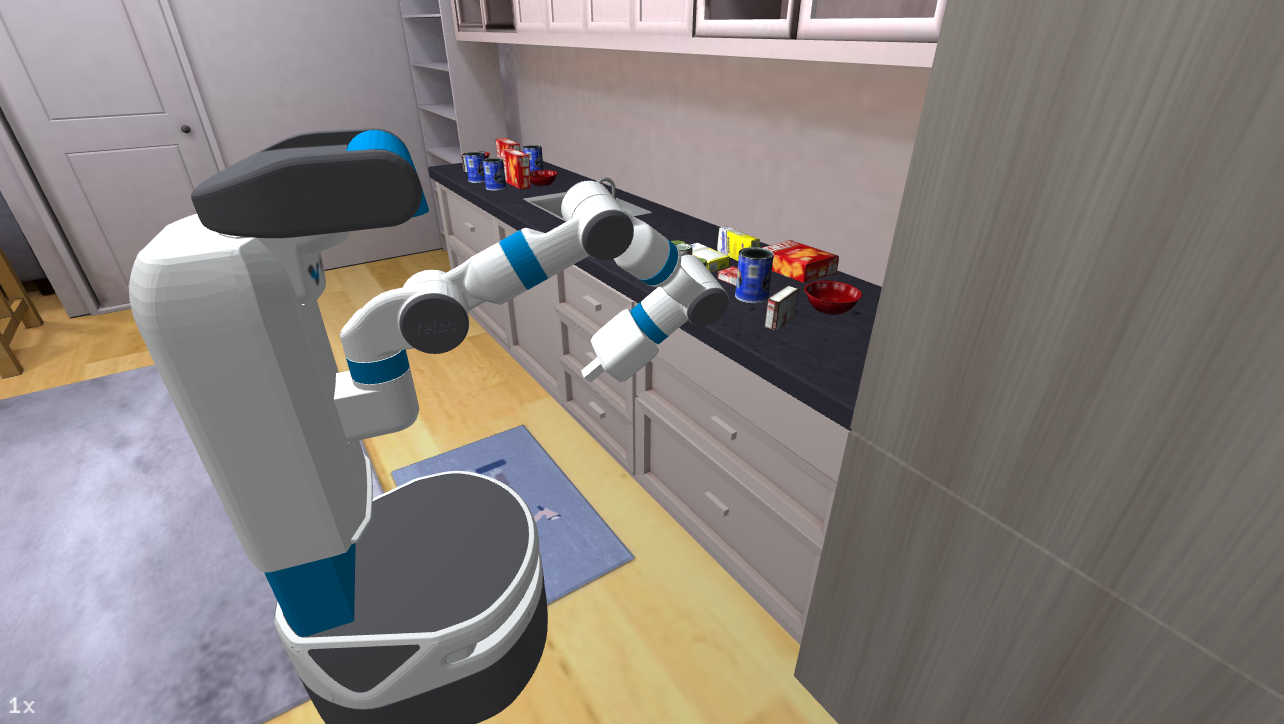}
       \put (0,40) {\color{red} \huge$s^0$}
      \end{overpic}
    	\includegraphics[trim=50 0 0 0, clip, width=0.32\textwidth]{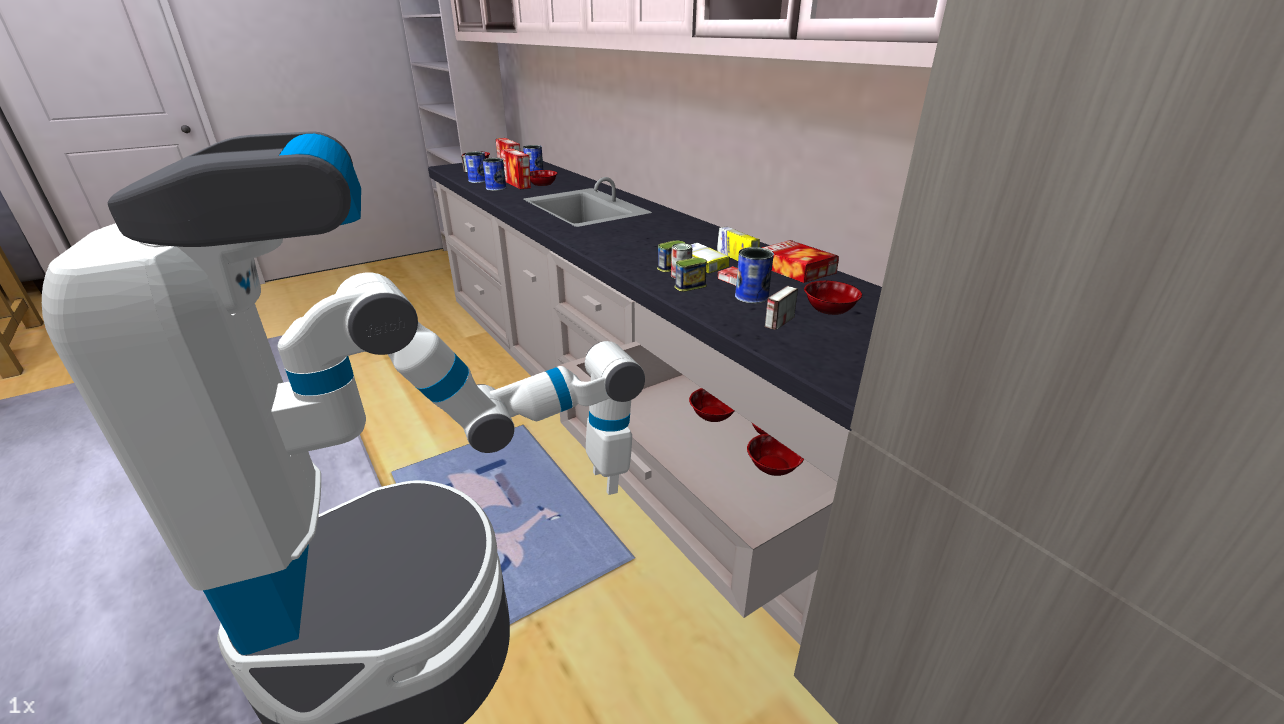}
      \begin{overpic}[trim=50 0 0 0, clip, width=0.32\textwidth]{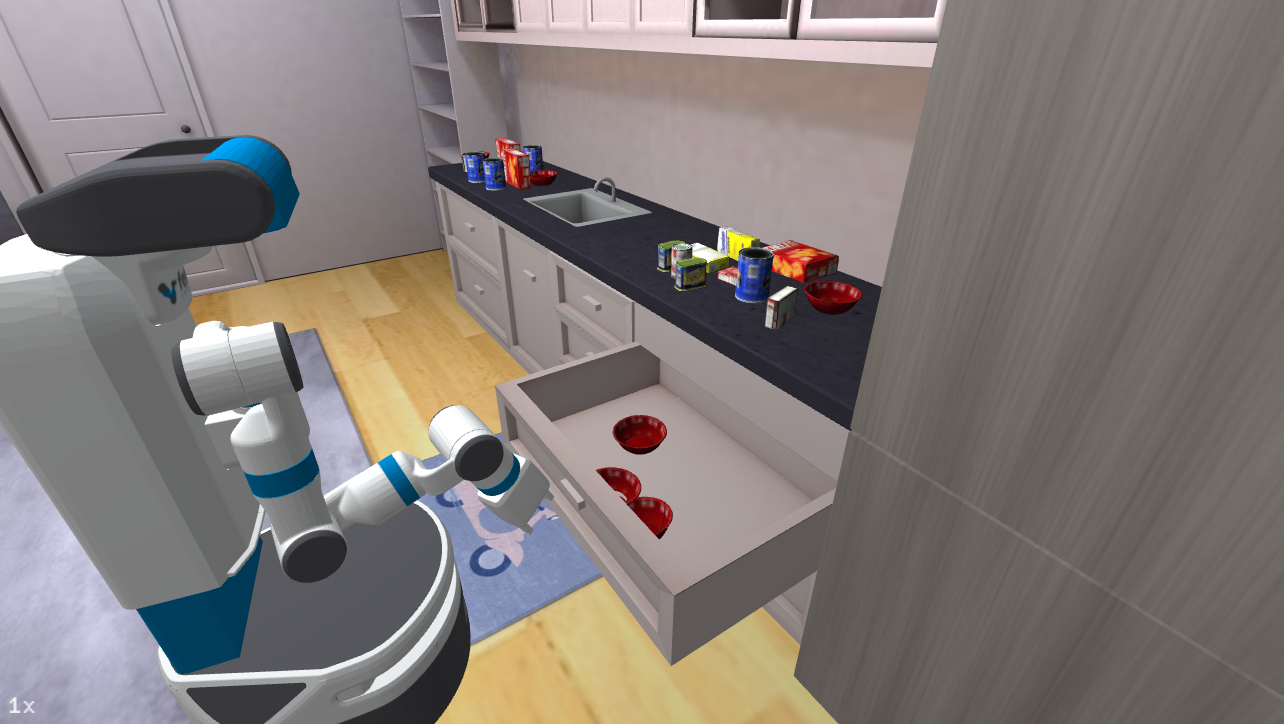}
        \put (0,40) {\color{red} \huge$s^*$}
      \end{overpic}
      \caption{Open (Drawer)}
      \label{fig:supp:open_drawer}
    \end{subfigure} \\    
    \begin{subfigure}[t]{0.99\linewidth}
    	\centering
      \begin{overpic}[trim=50 0 0 0, clip, width=0.32\textwidth]{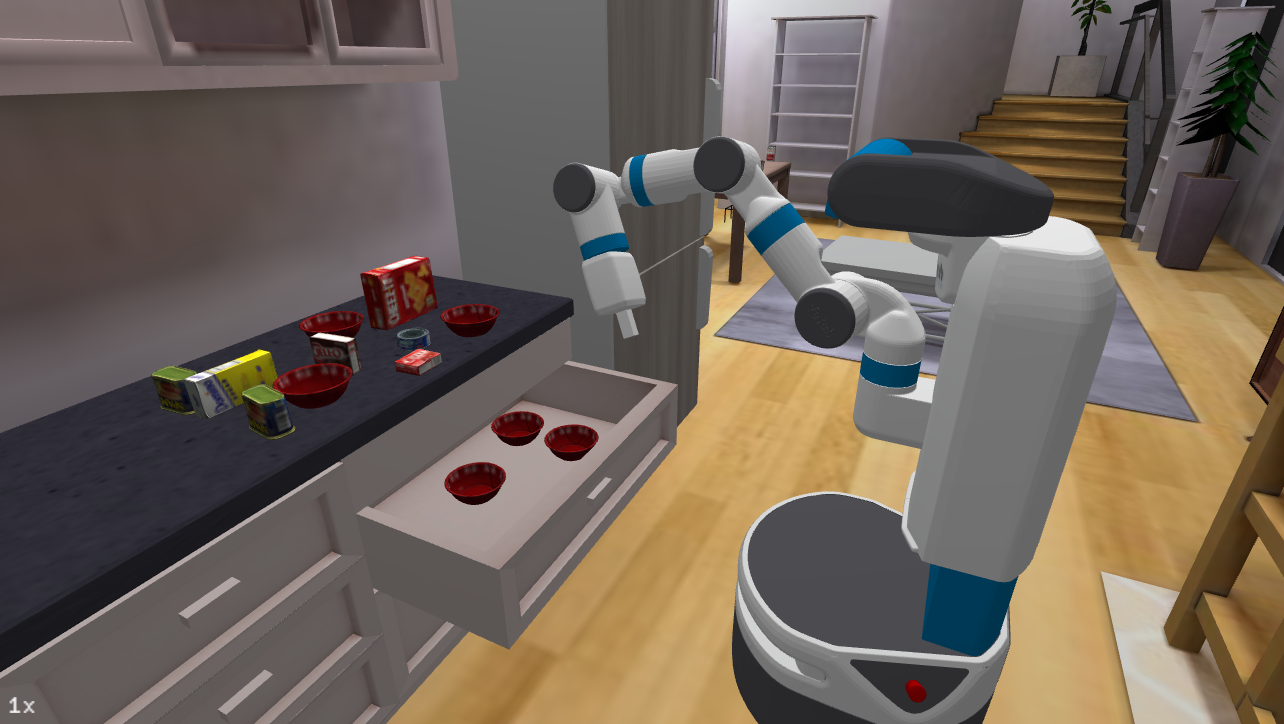}
       \put (0,40) {\color{red} \huge$s^0$}
      \end{overpic}
    	\includegraphics[trim=50 0 0 0, clip, width=0.32\textwidth]{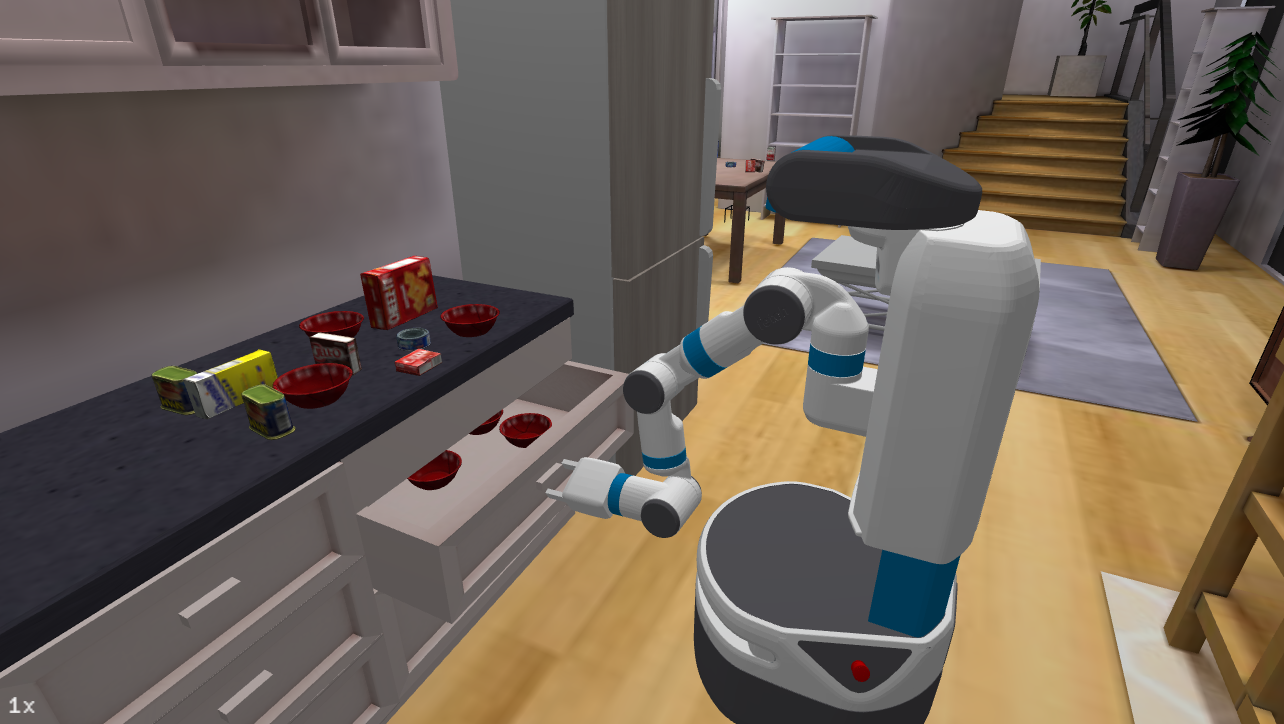}
      \begin{overpic}[trim=50 0 0 0, clip, width=0.32\textwidth]{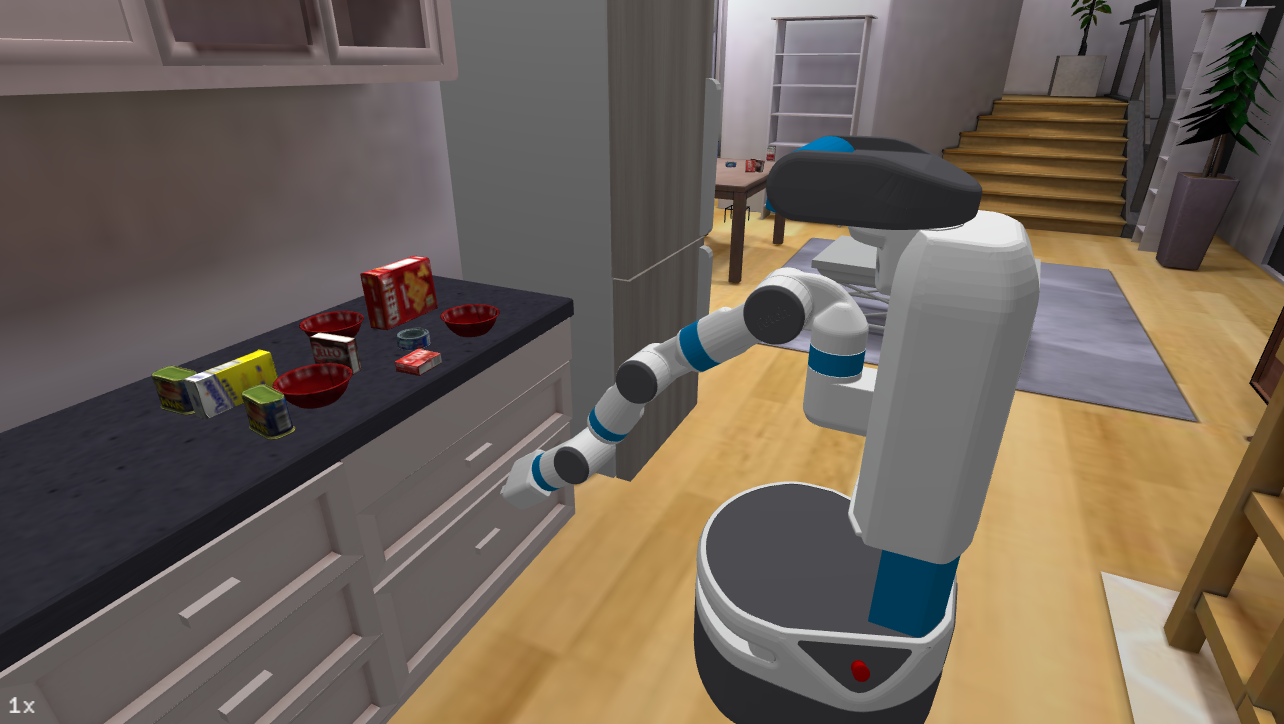}
        \put (0,40) {\color{red} \huge$s^*$}
      \end{overpic}
      \caption{Close (Drawer)}
      \label{fig:supp:close_drawer}
    \end{subfigure} \\    
    \begin{subfigure}[t]{0.99\linewidth}
    	\centering
      \begin{overpic}[trim=50 0 0 0, clip, width=0.32\textwidth]{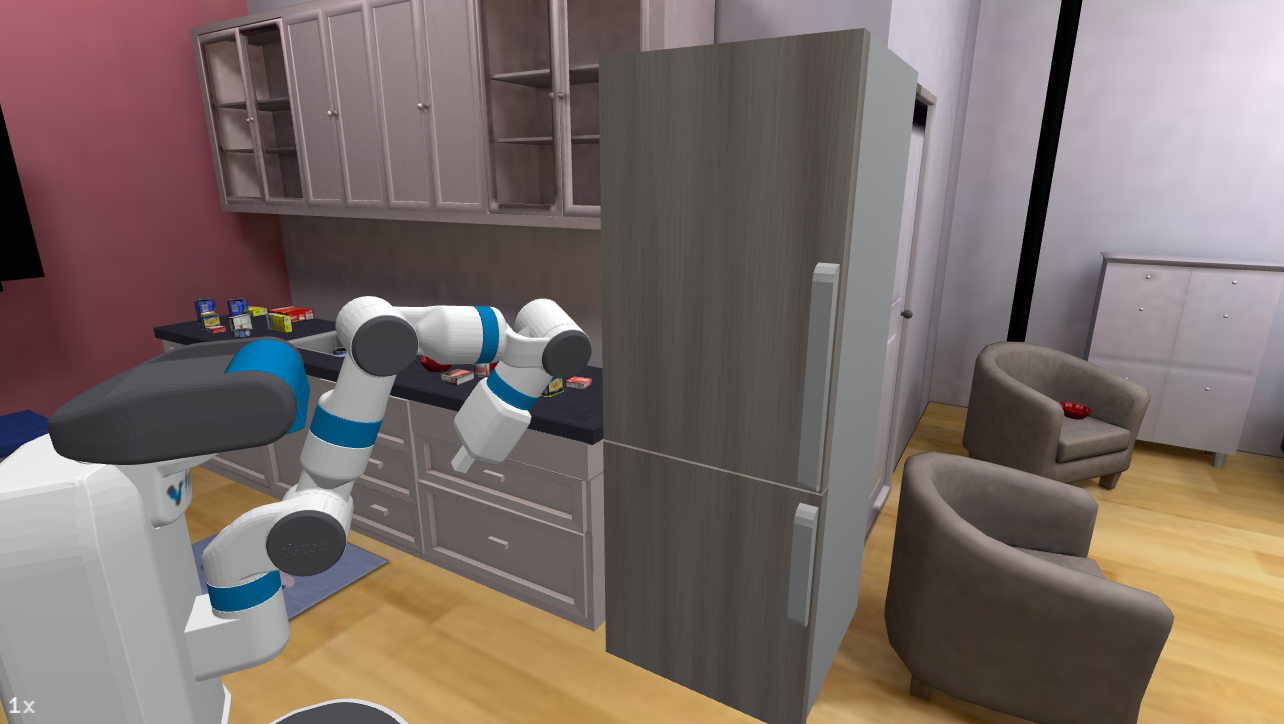}
       \put (0,40) {\color{red} \huge$s^0$}
      \end{overpic}
      \includegraphics[trim=50 0 0 0, clip, width=0.32\textwidth]{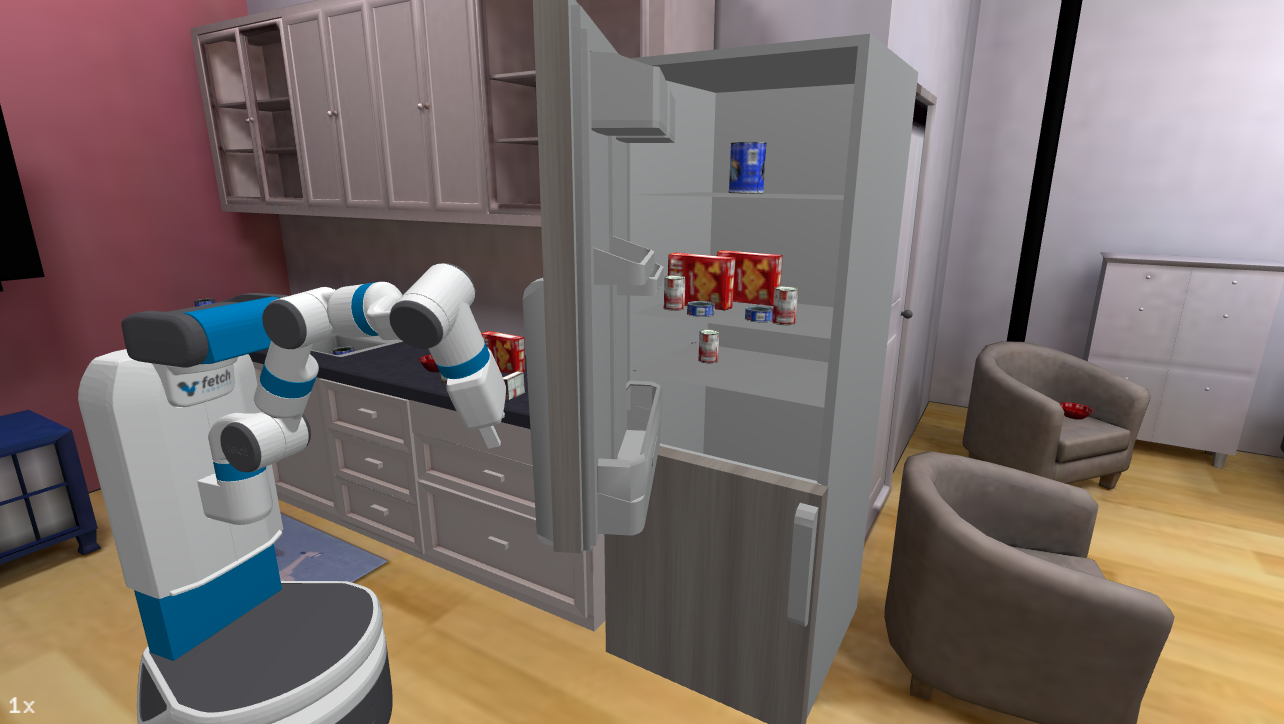}
      \begin{overpic}[trim=50 0 0 0, clip, width=0.32\textwidth]{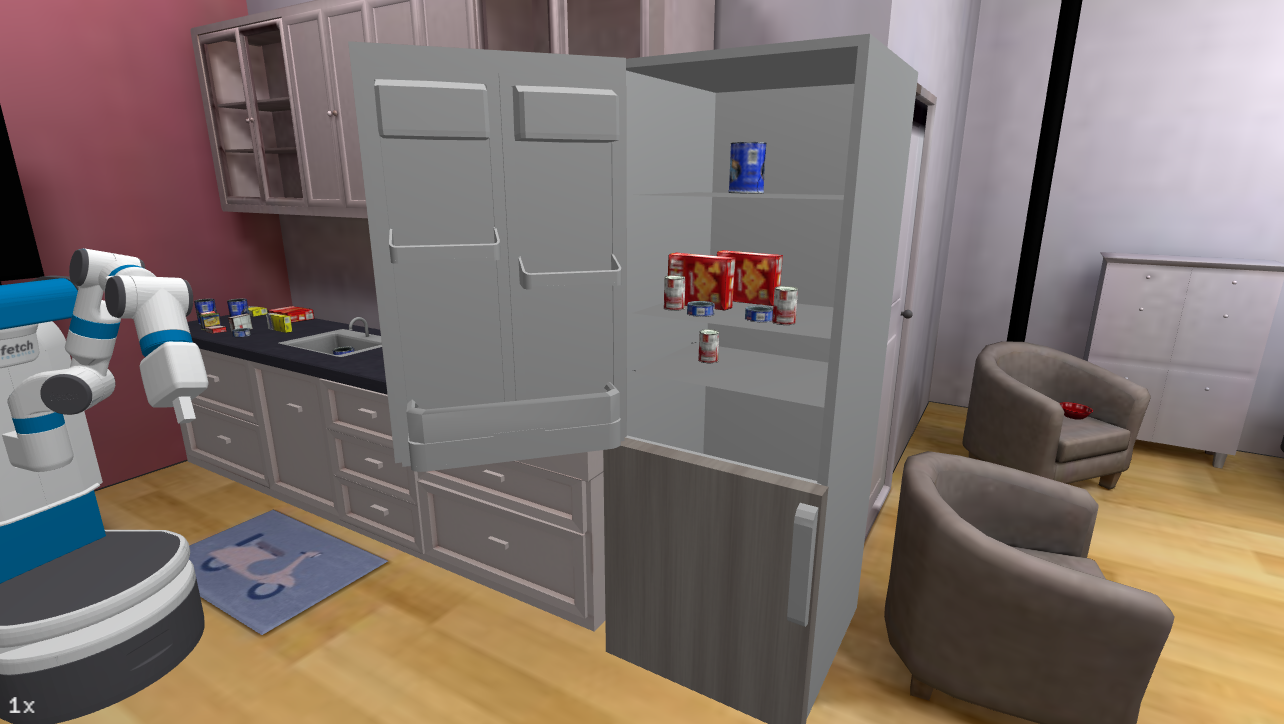}
        \put (0,40) {\color{red} \huge$s^*$}
      \end{overpic}
      \caption{Open (Fridge)}
      \label{fig:supp:open_fridge}
    \end{subfigure} \\    
    \begin{subfigure}[t]{0.99\linewidth}
    	\centering
      \begin{overpic}[trim=50 0 0 0, clip, width=0.32\textwidth]{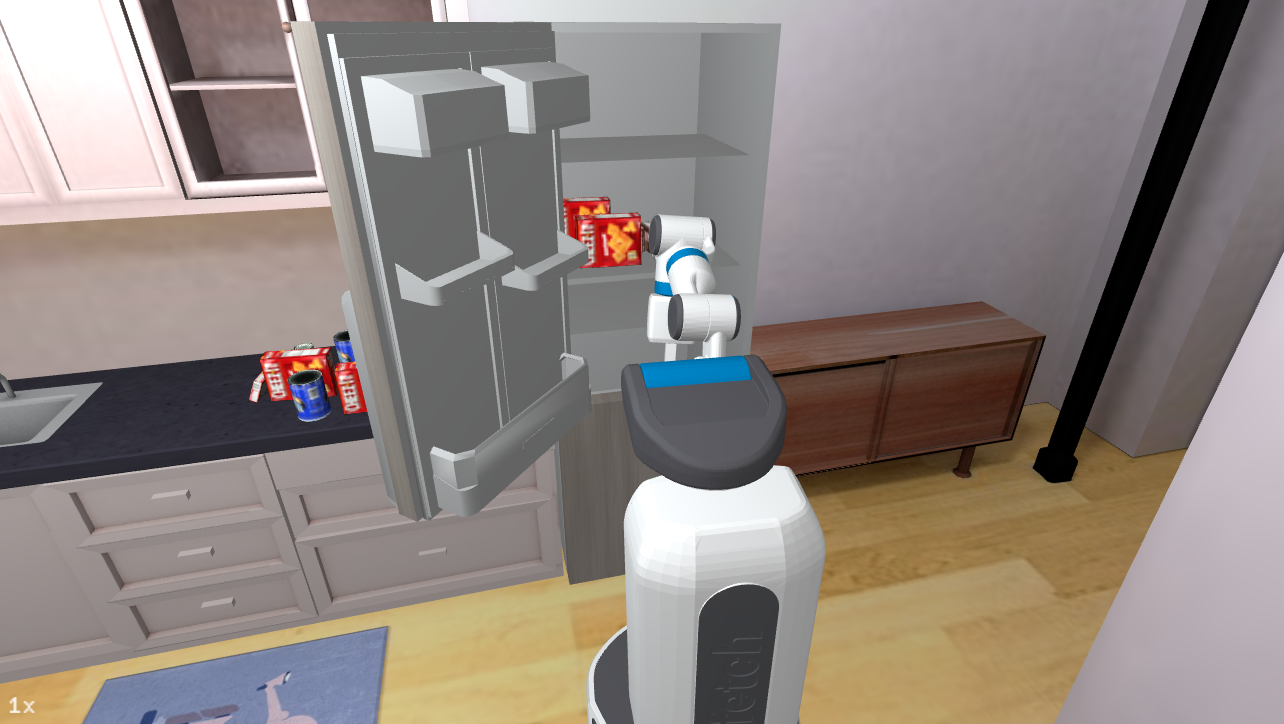}
       \put (0,40) {\color{red} \huge$s^0$}
      \end{overpic}
    	\includegraphics[trim=50 0 0 0, clip, width=0.32\textwidth]{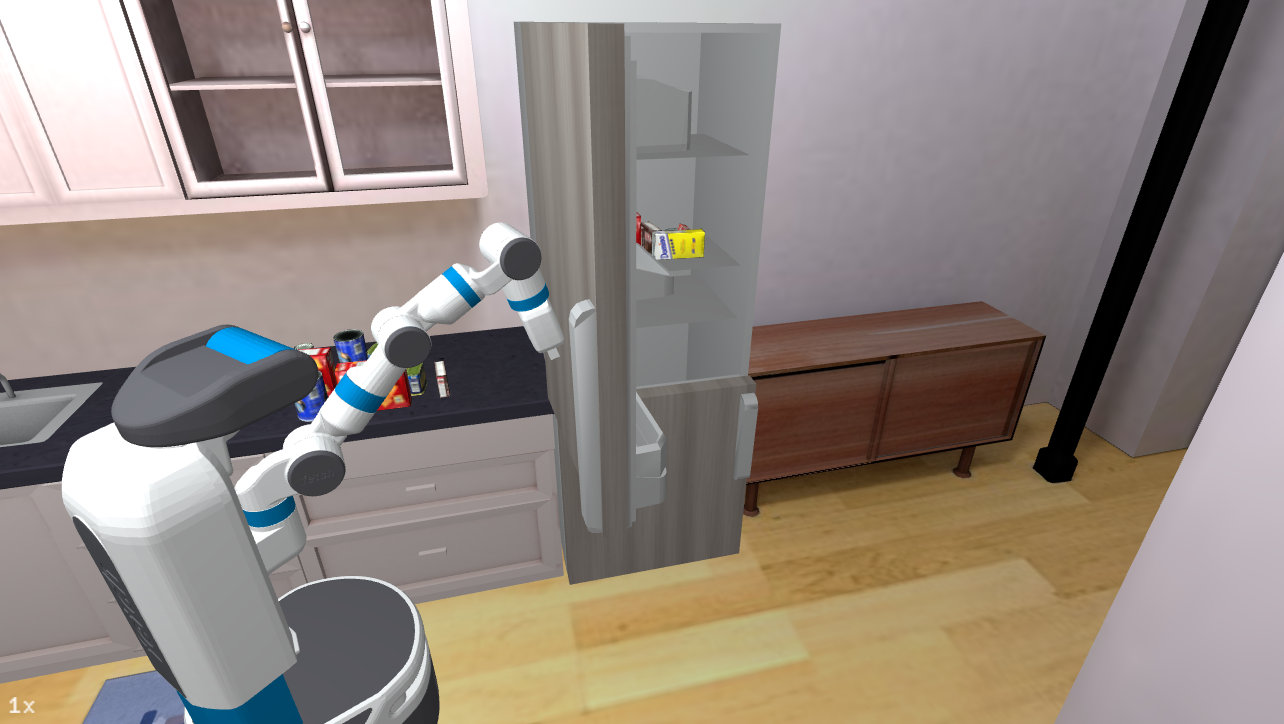}
      \begin{overpic}[trim=50 0 0 0, clip, width=0.32\textwidth]{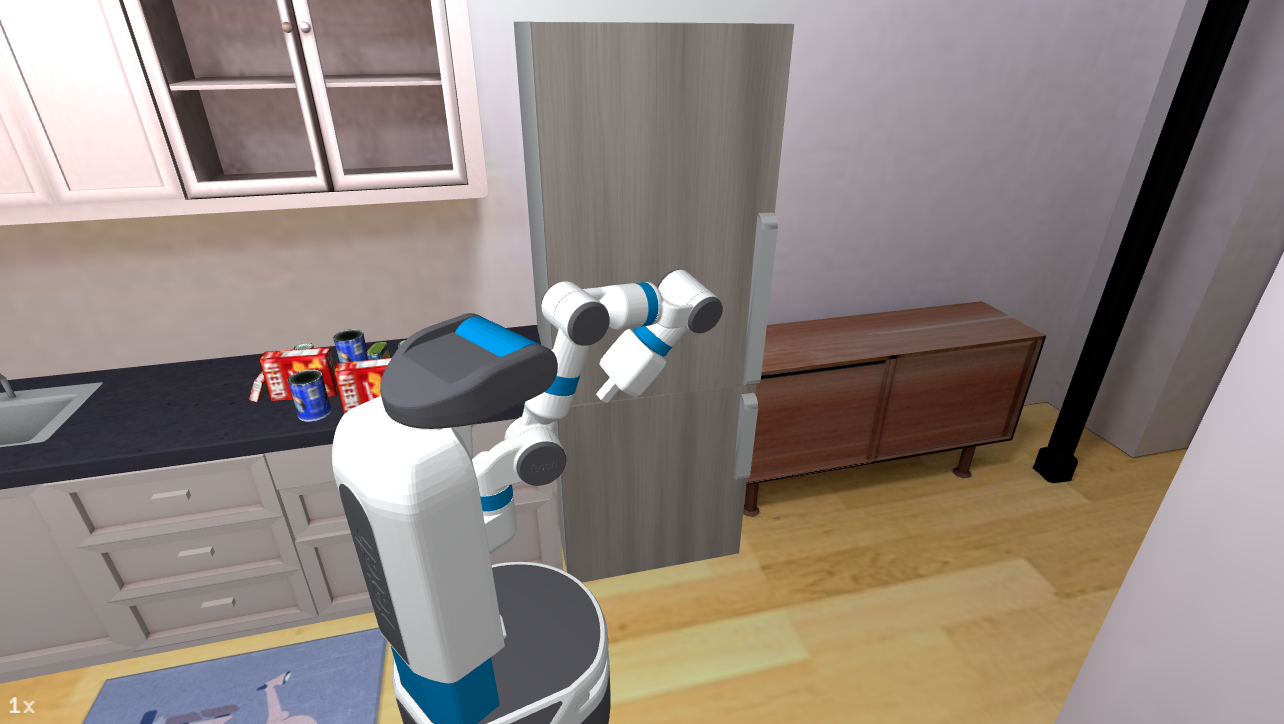}
        \put (0,40) {\color{red} \huge$s^*$}
      \end{overpic}
      \caption{Close (Fridge)}
      \label{fig:supp:close_fridge}
    \end{subfigure} \\    
    \begin{subfigure}[t]{0.99\linewidth}
    	\centering
      \begin{overpic}[trim=50 0 0 0, clip, width=0.32\textwidth]{figures/method/close_fridge/start.png}
       \put (0,40) {\color{red} \huge$s^0$}
      \end{overpic}
    	\includegraphics[trim=50 0 0 0, clip, width=0.32\textwidth]{figures/method/close_fridge/mid.png}
      \begin{overpic}[trim=50 0 0 0, clip, width=0.32\textwidth]{figures/method/close_fridge/end.png}
        \put (0,40) {\color{red} \huge$s^*$}
      \end{overpic}
      \caption{Navigate}
      \label{fig:supp:ac_nav}
    \end{subfigure} \\    
    \caption{
      Overview of all the high level planner actions with the pre-conditions (right), post-conditions (left), and an intermediate state when executing the action. 
    }
\end{figure*}

All three of the hierarchical methods, \TPS, \SPA, and \SPAoracle utilize a STRIPS high-level planner. 
A PDDL style domain file defines a set of predicates and actions. We define the following predicates 
\begin{itemize}
  \item \emph{in(X,Y)}: Is object $X$ in container $Y$? 
  \item \emph{holding(X)}: Is the robot holding object $X$? 
  \item \emph{at(X,Y)}: Is entity $X$ within interacting distance of $Y$? 
  \item \emph{is\_closed(X)}: Is articulated object $X$ in the closed state (separately defined for each articulated object)? 
  \item \emph{is\_open(X)}: Is articulated object $X$ in the open state?
\end{itemize}
And the following actions where each action is also linked to an underlying skill. 
\begin{itemize}
  \item \emph{pick(X)}: Pick object X (\Cref{fig:supp:pick}):
    \begin{itemize}
      \item Precondition: \emph{at(robot, X)}. We also include the precondition \emph{is\_open(Z)} if \emph{in(X, Z)} is true in the starting set of predicates.
      \item Postcondition: \emph{holding(X)}
      \item Skill: Pick
    \end{itemize}
  \item \emph{place(X, Y)}: Place object X at location Y (\Cref{fig:supp:place}): 
    \begin{itemize}
      \item Precondition: \emph{at(robot,Y), holding(X)}. We also include the precondition \emph{is\_open(Z)} if \emph{in(X, Z)} is true in the starting set of predicates.
      \item Postcondition: \emph{!holding(X),at(X,Y)}
      \item Skill: Place
    \end{itemize}
  \item \emph{open(X)}: Open articulated object X (\Cref{fig:supp:open_fridge,fig:supp:open_drawer}):
    \begin{itemize}
      \item Precondition: \emph{at(robot, X), is\_closed(X), !holding(Z), $\forall$ Z}
      \item Postcondition: \emph{is\_open(X)}
      \item Skill: If X is the fridge entity, then \opendoor, if X is the drawer entity then \opendrawer. 
    \end{itemize}
  \item \emph{close(X)}: Close articulated object X (\Cref{fig:supp:close_fridge,fig:supp:close_drawer}):
    \begin{itemize}
      \item Precondition: \emph{at(robot, X), is\_open(X), !holding(Z), $\forall$ Z}
      \item Postcondition: \emph{is\_closed(X)}
      \item Skill: If X is the fridge entity, then \closedoor, if X is the drawer entity then \closedrawer. 
    \end{itemize}
\end{itemize}

Each task defines the initial set of predicates and the goal set of predicates. 
We use a STRIPS planner to find a set of actions to transform the starting predicates into the goal predicates. 
Since we only deal with object rearrangement problems, the goal predicates of each task are of the form \emph{at(obj\_X, obj\_goal\_X)} for each object \emph{X} to be rearranged.
\cleanhouse and \settable includes \emph{is\_closed(fridge),is\_closed(drawer)} into the goal and starting predicates while \stockfridge includes \emph{is\_open(fridge),is\_open(drawer)} into the goal and starting predicates.
The starting predicates which specify containement and are listed below:
\begin{itemize}
  \item \settable: \emph{in(bowl,drawer),in(fruit,fridge),in(fruit\_goal,bowl\_goal)}
  \item \stockfridge: No containement specified for this task (everything starts open).
  \item \cleanhouse: No containement in this task.
\end{itemize}
We run the STRIPS planner once per task and save the minimum length solution. 
The saved plan is used as the sequence of agent-environment interactions for partial evaluation in \Cref{sec:hab_rearrang_results}.

\subsection{RL Skill Training}
\label{sec:supp:rl_skills}

Each skill is trained to perform a desired interaction.
To facilitate better transitions between skills, skills must reset the robot arm to a ``resting position" with the end-effector at a certain position in front of the robot.
Since the agent has access to proprioceptive state, this also serves as a termination signal for the skill. 

For all skills $\Delta_{arm}^{o}$ is the change in distance between the end-effector and object (if $d_t$ is the distance between the two at timestep $t$ then $\Delta_{arm}^o = d_{t-1} - d_t$). 
$\Delta_{arm}^{rest}$ is the change in distance between the arm end-effector and resting position, 
$\Delta_{arm}^{g}$ is the change in distance between the object and object goal, 
$\Delta_{arm}^h$ is the change in distance between the end-effector and articulated object interaction point, 
and $\Delta_a^g$ is the distance between the articulated object state and its goal state. 
Skills incorporate a collision penalty with $C_t$, the collision force in Newtons at time $t$. 
Episodes terminate if they exceed a collision threshold $C_{max}$ in Newtons.
By default in each skill training episode the robot base angle varies by 0.3 radians, base $x,y$ position by 0.1 meters, and end-effector $x,y,z$ by 0.05 meters.
For each skill, the maximum episode length is 200 steps.

Arm control refers a 3D relative offset of the end-effector and a 1D gripper value.
Base control is a 2D linear and angular velocity. 
For all rewards, implicitly for time $t$, $\mathbb{I}_{holding}$ is the indicator for if the robot is holding an object, $\mathbb{I}_{force}$ is the indicator for if the force collision threshold was exceeded.

For training each skill, we utilize 5,000 training configurations. 
The full task where these skills are deployed are in unseen scene configurations and unseen object placements. 
We also show evaluation for each skill on an evaluation set of 500 configurations in \Cref{sec:supp:skill_exps}. 

\begin{itemize}
  \item \pick$(s^0_i)$ Pick the object at starting state $s_i^0$:  
    \begin{itemize}
      \item Starting state: Objects and clutter is randomly spawned on one of 6 receptacles (sofa, fridge, counter left, counter right, light wood table, dark wood table). 
        Robot is facing the object with default noise applied to the base, orientation, and end-effector. 
      \item Success: Robot holding the target object and in the resting position. 
      \item Failure: $C_{max} = 5000$. The episode also terminates if the robot picks up the wrong object.
      \item Reward: $ \mathbb{I}_{wrong}$  indicates the policy picked up the wrong object.
        \begin{align*}
          r_t = 20\mathbb{I}_{success} + 5\mathbb{I}_{pickup} + 20\Delta_{arm}^{o} \mathbb{I}_{!holding} + 20\Delta_{arm}^{r}\mathbb{I}_{holding} - \max(0.001 C_t, 1.0) \\
          - 10 \mathbb{I}_{force} - 5 \mathbb{I}_{wrong} - 5 \mathbb{I}_{dropped}
        \end{align*} 
      \item Agent action space: Arm control. 
        Once an object is picked, the gripper scalar action is masked out until the skill terminates to prevent dropping the object. 
      \item Agent observation space: Arm+Head depth camera with relative position between object starting position and end-effector.
    \end{itemize}
  \item \place$(s_i^*)$ Place the currently held object at goal state $s_i^*$: 
    \begin{itemize}
      \item Starting state:  An object goal position and clutter is randomly spawned on one of 7 receptacles (sofa, fridge, counter left, counter right, light wood table, dark wood table, sink). 
        The robot is facing the object goal with default noise applied to the base, orientation, and end-effector.
        The object to place starts in the robot's grasp.
      \item Failure: $C_{max} = 7500$. 
      \item Success: The episode is a success if the object is at the goal and the arm is at the resting position. 
      \item Reward: $ \mathbb{I}_{wrong}$  indicates the policy picked up an object.
        \begin{align*}
          r_t = 20\mathbb{I}_{success} + 5\mathbb{I}_{place} + 20\Delta_{o}^{g} \mathbb{I}_{holding} + 20\Delta_{arm}^{r} \mathbb{I}_{!holding} + 20\Delta_{arm}^{r}\mathbb{I}_{holding} - \max(0.001 C_t, 1.0)  \\
          - 10 \mathbb{I}_{force} - 5 \mathbb{I}_{wrong}
        \end{align*} 
      \item Agent action space: Arm control. 
        Once an object is placed, the gripper scalar action is masked out until the skill terminates.
      \item Agent observation space: Arm+Head depth camera with relative position between object goal position and end-effector.
    \end{itemize}
  \item \opendoor $(s_i)$ open the door of the fridge containing object or goal position $s_i$: 
    \begin{itemize}
      \item Starting state: The fridge door starts closed.  
        The robot spawns in a $0.9m \times 1.2m$ square in front of the fridge, facing the fridge handle with default noise applied to the base, orientation, and end-effector. 
      \item Reward: $ \mathbb{I}_{out}$ indicates the robot base left the spawn region. 
        \begin{align*}
          r_t = 10\mathbb{I}_{success} + 5\mathbb{I}_{grabbed} + 1\Delta_{arm}^{h} + 1\Delta_{a}^{g} - 10\mathbb{I}_{out} 
        \end{align*} 
      \item Failure: There is no collision force threshold. The episode terminates with failure if the robot leaves the spawn region.
      \item Success: The episode is a success if the fridge is open more than 90 degrees and the robot is in the resting position. 
      \item Agent action space: Arm and base control.
      \item Agent observation space: Arm+Head depth camera with relative position between end-effector and a target object starting or goal position in the fridge. 
    \end{itemize}
  \item \closedoor$(s_i)$ close the door of the fridge containing object or goal position $s_i$:
    \begin{itemize}
      \item Starting state: The fridge door starts open with a fridge door angle in $[\pi/4-2\pi/3]$ radians. 
        The robot spawns in a $0.9m \times  1.2m$ square in front of the fridge, facing the fridge handle with default noise applied to the base, orientation, and end-effector.
      \item Reward: 
        \begin{align*}
          r_t = 10\mathbb{I}_{success} + 1\Delta_{arm}^{h} + 1\Delta_{a}^{g} 
        \end{align*} 
      \item Failure: There is no collision force threshold. The episode terminates with failure if the robot leaves the spawn region. 
      \item Success: The episode is a success if the fridge is closed with angle within 0.15 radians of closed. and the robot is in the resting position. 
      \item Agent action space: Arm and continuous base control.
      \item Agent observation space: Arm+Head depth camera with relative position between end-effector and a target object starting or goal position in the fridge. 
    \end{itemize}
  \item \opendrawer$(s_i)$  open the drawer containing object or goal position $s_i$:
    \begin{itemize}
      \item Starting state: The drawer starts completely closed.
        A random subset of the other drawers are selected and opened between 0-100\%. 
        The robot spawns in a $0.15m \times 0.75m$ rectange in front of the drawer to be opened, facing the drawer handle with default noise applied to the base, orientation, and end-effector.
      \item Reward: 
        \begin{align*}
          r_t = 10\mathbb{I}_{success} + 5\mathbb{I}_{grabbed} + 1\Delta_{arm}^{h} + 1\Delta_{a}^{g} 
        \end{align*} 
      \item Failure: There is no collision force threshold. 
      \item Success: The episode is a success if the drawer is between 90-100\% open and the arm is at the resting position.
      \item Agent action space: Arm control.
      \item Agent observation space: Arm+Head depth camera with relative position between end-effector and a target object starting or goal position in the drawer. 
    \end{itemize}
  \item \closedrawer$(s_i)$ close the drawer containing object or goal position $s_i$:
    \begin{itemize}
      \item Starting state: The target drawer starts between 80-100\% open. 
        A random subset of the other drawers are selected and opened between 0-100\%. 
        The robot spawns in a $0.15m \times 0.75m$ rectangle in front of the drawer to be closed, facing the drawer handle with default noise applied to the base, orientation, and end-effector.
      \item Reward: 
        \begin{align*}
          r_t = 10\mathbb{I}_{success} + 1\Delta_{arm}^{h} + 1\Delta_{a}^{g} 
        \end{align*} 
      \item Failure: There is no collision force threshold. 
      \item Success: The episode is a success if the fridge is closed and the arm is at the resting position. 
      \item Agent action space: Arm control.
      \item Agent observation space: Arm+Head depth camera with relative position between end-effector and a target object starting or goal position in the drawer. 
    \end{itemize}
  \item \navigate: Navigates to the start of other skills. 
    Importantly, the agent is only provided the 3D coordinate of the start or goal location to navigate to, for instance an object in the fridge or a location to place an object on the counter. 
    This is different from the goal position the agent actually needs to navigate to which is on the floor in front of the object.
    The target on the floor is calculated based on the start state distribution of other skills. 
    The agent does not have access to this privaledged information about the navigation goal position.
    Furthermore, the agent not only needs to navigate to a particular location but also face the correct direction (notated as $\theta^*$). 
    \begin{itemize}
      \item Starting State: A random base position and rotation in the scene. 
        The state of the fridge, drawers, and object configurations are randomly sampled from one of the previous 6 skill training setups. 
      \item Reward:
        \begin{align*}
          r_t = 10 \mathbb{I}_{success} + 20\Delta_{agent}^{goal} + \Delta_{\theta}^{\theta^*} I_{\Delta_{agent}^{goal} < 0.9}
        \end{align*}
        Where $\Delta_{agent}^{goal}$ is the change in geodesic distance to the goal, $\theta$ is the current agent rotation, $\theta^*$ is the target orientation, and $\Delta_{\theta}^{\theta^*}$ is the change in L1 norm between the current agent angle and the target angle. 
      \item Failure: There is no collision force threshold. The episode horizon is $500$ steps. 
      \item Success: The agent is within 0.3 meters of the goal, 0.5 radians of the target angle, and has called the stop action at the current time step. 
      \item Agent action space: Similarily to \cite{yokoyama2021success}, the navigation is handeled by a discrete action space which is then translated into continuous actions. 
        Specifically, the linear velocity from -0.5 to 1 is discretized into 4 cells and the angular velocity from -1 to 1 is discretized into 5 cells, giving 20 cells in total.
        The action corresponding to 0 linear and angular velocity is the stop action. 
      \item Agent observation space: The Head depth camera with the relative position between the robot end-effector and object. 
    \end{itemize}
    We find that learning the termination condition is difficult for the navigation skill as demonstrated by \reffig{supp:fig:nav_skill} which demonstrates that learned termination results in a 20\% drop in success rate.
\end{itemize}

\subsection{\monolithic}
\begin{figure}
    \centering
    \begin{subfigure}[t]{0.9\linewidth}
    	\centering
    	\includegraphics[width=\textwidth]{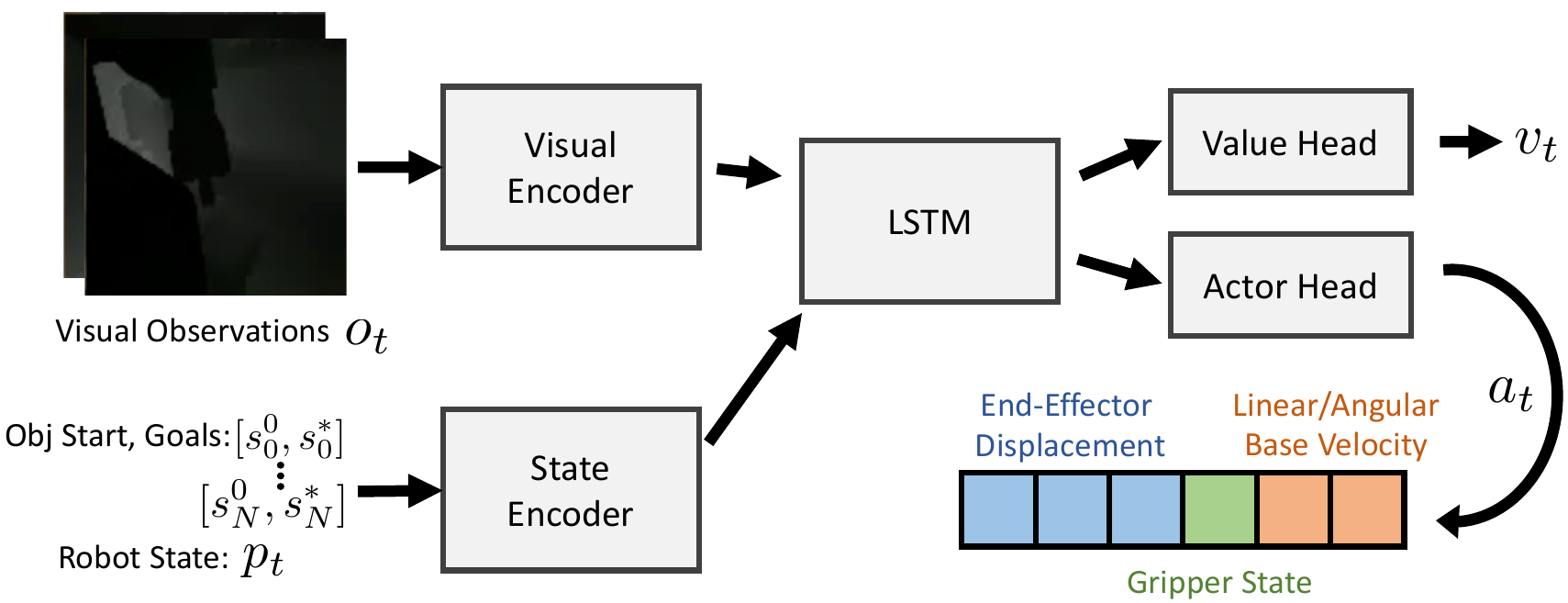}
    \end{subfigure} \\    
    \caption{
      The \monolithic policy architecture for the \tasksuitename task. 
      The policy maps egocentric visual observations $o_t$, the task-specification in the form of a series of geometric object goals $[b^1, g^1, \dots, b^N, g^N$ where $N$ is the number of objects to rearrange, and the robot proprioceptive state $s_t$ into an action which controls the arm, gripper, and base velocity. 
      A value output is also learned for the PPO update. 
    }
    \label{fig:skill_arch}
\end{figure}

The \monolithic approach for the main task follows a similar setup as \Cref{sec:supp:mono} but with a different action space and reward structure. 
The agent maps the egocentric visual observations, task-specification, and proprioceptive state into an action which controls the arm, gripper, and base velocity (policy architecture visualized in \reffig{fig:skill_arch}). 
The arm actions are the same as described in \Cref{sec:exp:analysis}. 

A challenge of the \monolithic approach is learning a long complicated task structure.
We therefore train with a dense reward guiding the robot to complete each part of the task. 
Using a pre-specified ordering of the skills from \Cref{sec:supp:rl_skills}, we infer which skill the robot is currently at. 
We start with the first skill in the pre-specified skill ordering, when that skill terminates we progress to the next skill, etc. 
This current inferred skill only provides the reward to the \monolithic approach. 
The termination, starting state distribution, and transition function all still come from the actual task. 
We utilize a training set of 5000 configurations for the task.
The evaluation set of task configurations consist of new objects placements.

\section{\tasksuitenamefull Further Experiments}
\label{sec:supp:hab_exp}

\subsection{\SPA Failure Analysis}
\label{sec:supp:spa_failures} 
In this section we analyze the source of errors for the \SPA approaches for the \tasksuitename results from \reffig{fig:all_full_task}. 
Specifically, we analyze which part of the sense-plan-act pipeline fails. 
We categorize the errors into three categories.
The first category (`Target Plan') is errors finding a collision free joint configuration which reaches the goal to provide as a goal state for the motion planner. 
The second category (`Motion Plan') is errors with the motion plan phase timing out (both \TPSPA and \TPSPAoracle use a 30 second timeout). 
The third category (`Execution') is if the planned sequence of joint angles is unable to be executed. 
Failures for motion planning the pick, place and arm resets are grouped into these categories. 
These categories do not account for the learned navigation failure rates. 

\begin{figure}
    \centering
    \begin{subfigure}[t]{0.24\linewidth}
    	\includegraphics[width=\textwidth]{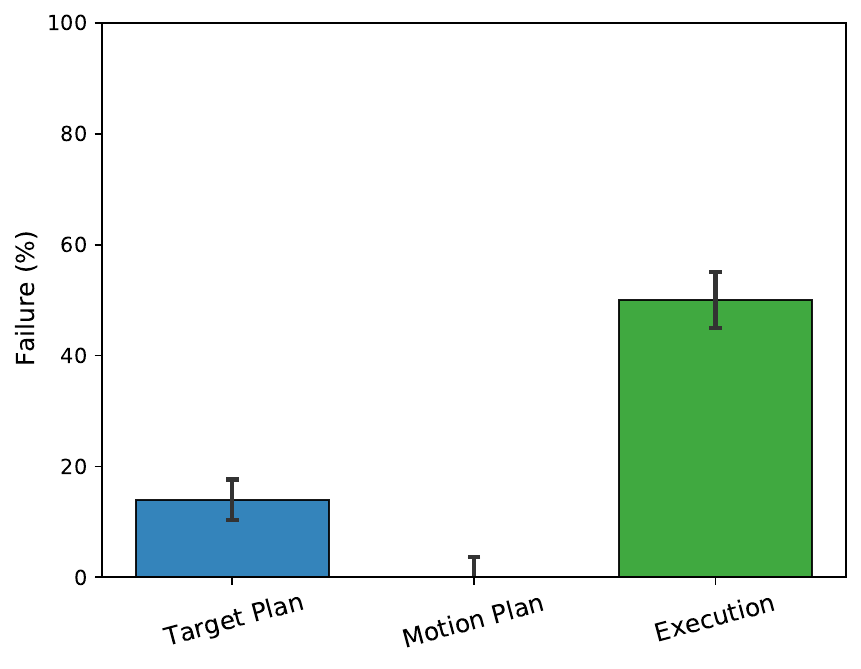}
      \caption{\TPSPA \\\cleanhouse}
    \end{subfigure} 
    \begin{subfigure}[t]{0.24\linewidth}
    	\includegraphics[width=\textwidth]{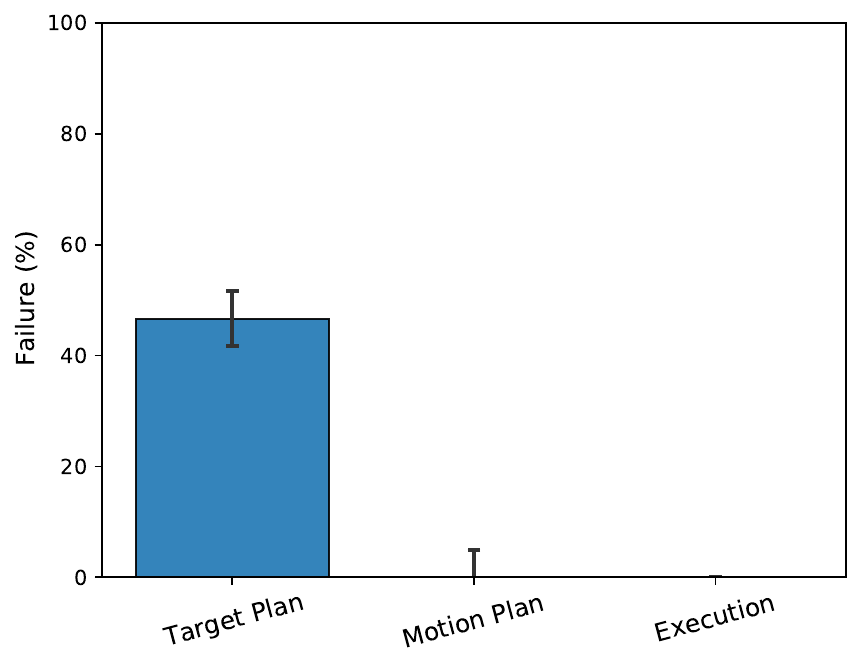}
      \caption{
        \TPSPAoracle \\\cleanhouse \\ 
      }
    \end{subfigure} 
    \begin{subfigure}[t]{0.24\linewidth}
    	\includegraphics[width=\textwidth]{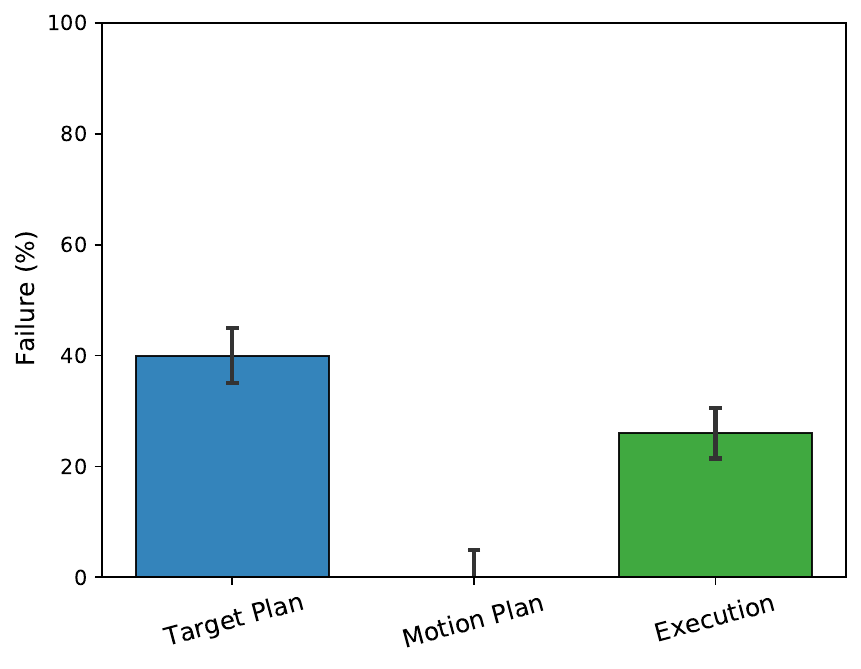}
      \caption{\TPSPA \\\stockfridge}
    \end{subfigure} 
    \begin{subfigure}[t]{0.24\linewidth}
    	\includegraphics[width=\textwidth]{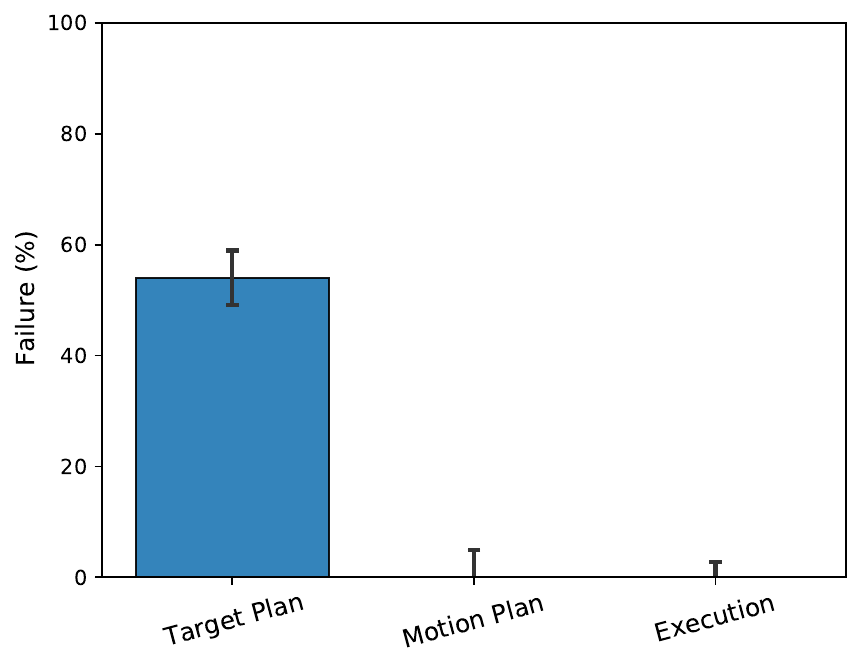}
      \caption{\TPSPAoracle \\\stockfridge}
    \end{subfigure} 
    \caption{
      Motion planner failure rates for \reffig{fig:all_full_task}. 
      Numbers indicate the percent of the 100 evaluation episodes the failure category occurs. 
      `Target Plan' is failures in finding a valid target joint configuration, `Motion Plan' is the motion planning timing out, and `Execution' is the planned sequence of joint angles failing to execute. 
    }
    \label{supp:fig:mp_err}
\end{figure}

We analyze these sources of errors for \TPSPA and \TPSPAoracle with learned navigation in \reffig{supp:fig:mp_err}. 
`Target Plan' fails due to the sampling based algorithm timing out to find the collision free target joint state which accomplishes the goal. 
Methods therefore have a higher `Target Plan' failure rate on \stockfridge where the agent must reach into the fridge to grab and place objects. 
\TPSPAoracle has a higher `Target Plan' failure rate because it has complete information about the geometry in the scene. 
This results in more obstacles being included in the collision check and therefore makes the target sampling harder. 
On the other hand, obstacles do not exist outside the perception of \TPSPA such as behind other objects or outside the field of view making the target sampling easier. 
Next, we see that all methods have a zero `Motion Plan' failure rate. 
This means that when the algorithm is able to find a valid target joint state, the motion planning algorithm is able to find a valid series of joint configurations from the current joint state to the target joint state.
Finally, the `Execution' failure rates for \TPSPAoracle is zero since this method uses a perfect controller. 
On the other hand, \TPSPA can fail to execute due to the imperfect execution controller and planning from incomplete information.
A planned path returned as successful from the motion planner can fail due to unperceived obstacles.

\subsection{Learning Curves}
\label{sec:supp:skill_exps}

\begin{figure}
    \centering
    \begin{subfigure}[t]{0.32\linewidth}
    	\includegraphics[width=\textwidth]{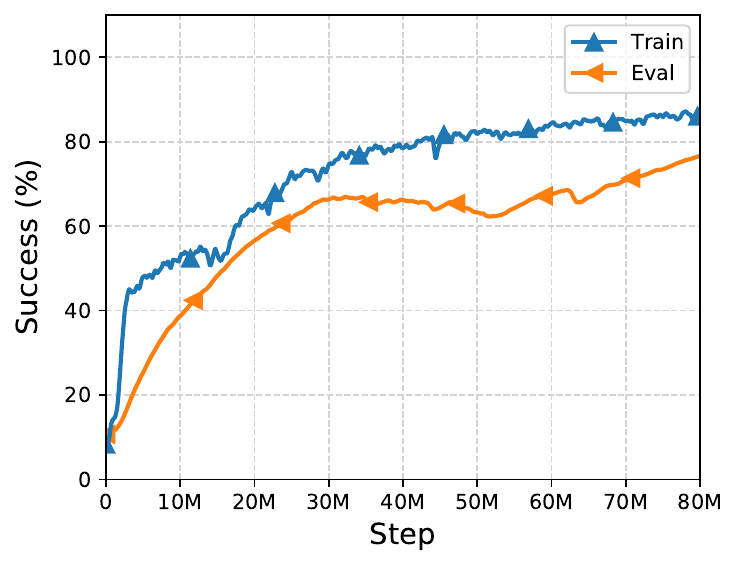}
      \caption{Pick Skill}
    \end{subfigure} 
    \begin{subfigure}[t]{0.32\linewidth}
    	\includegraphics[width=\textwidth]{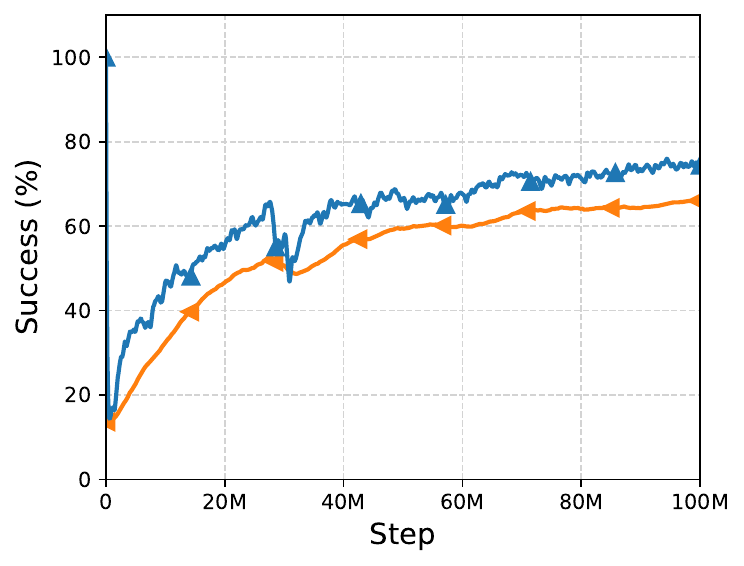}
      \caption{Place Skill}
    \end{subfigure} 
    \begin{subfigure}[t]{0.32\linewidth}
    	\includegraphics[width=\textwidth]{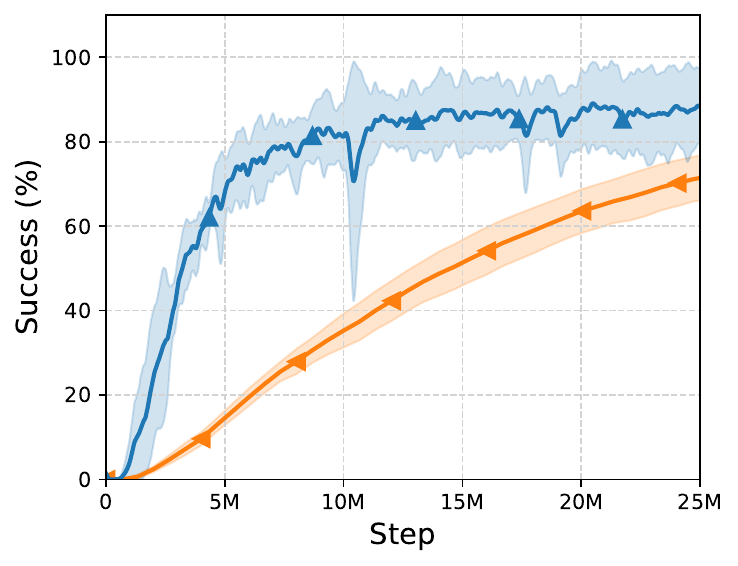}
      \caption{Open Drawer Skill}
    \end{subfigure} 
    \begin{subfigure}[t]{0.32\linewidth}
    	\includegraphics[width=\textwidth]{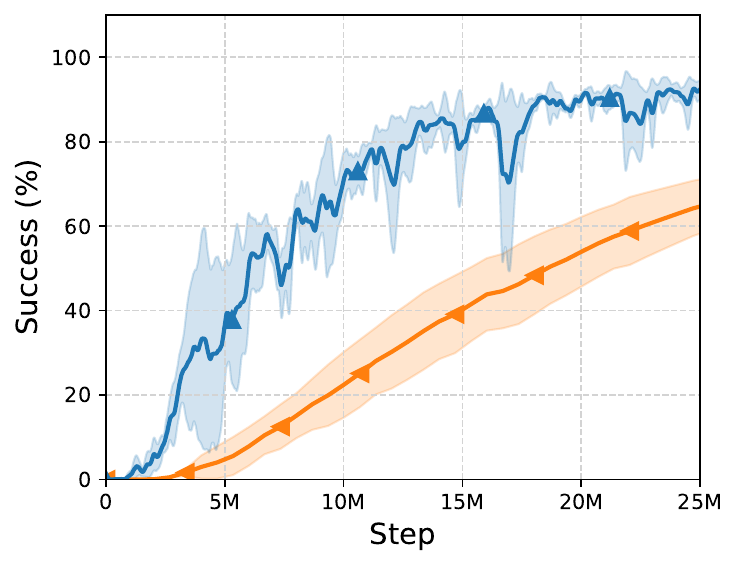}
      \caption{Close Drawer Skill}
    \end{subfigure} 
    \begin{subfigure}[t]{0.32\linewidth}
    	\includegraphics[width=\textwidth]{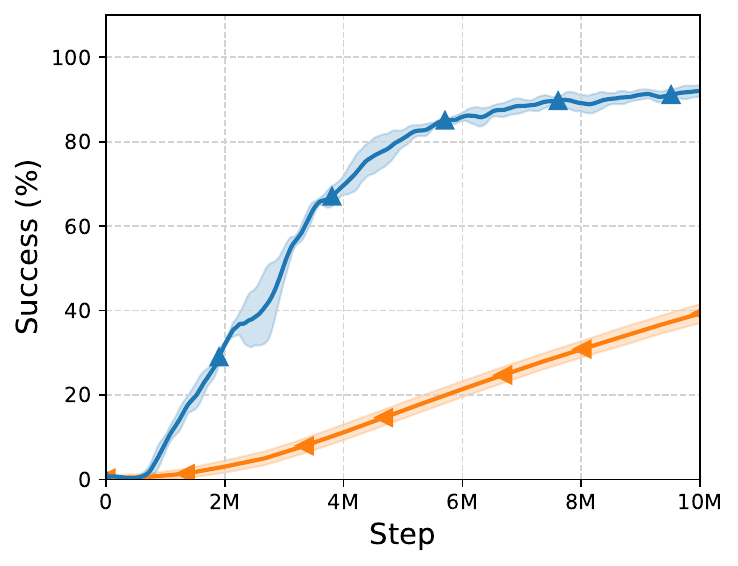}
      \caption{Open Fridge Skill}
    \end{subfigure} 
    \begin{subfigure}[t]{0.32\linewidth}
    	\includegraphics[width=\textwidth]{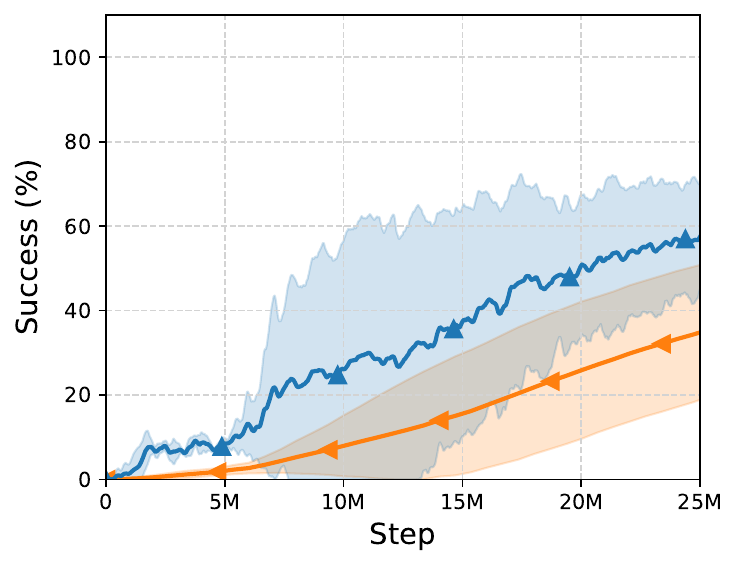}
      \caption{Close Fridge Skill}
    \end{subfigure} 
    \caption{
      Training and evaluation curves for the skills with averages and standard deviations across 3 seeds (except for the Pick and Place skills which are only for 1 seed).
    }
    \label{supp:fig:skills}
\end{figure}

All methods except for \monolithic utilize a set of skills. 
For \TPS these skills are learned with RL described in \Cref{sec:supp:rl_skills}. 
The learning curves showing the success rate as a function of the number of samples is illustrated in \Cref{supp:fig:skills}.
We include both the success rates from training and the results on a held out set of $100$ evaluation episodes. 
\SPA approaches use the robotics pipeline described in \Cref{sec:supp:mp} and do not require any learning. 

\begin{figure}
    \centering
    \includegraphics[width=0.5\linewidth]{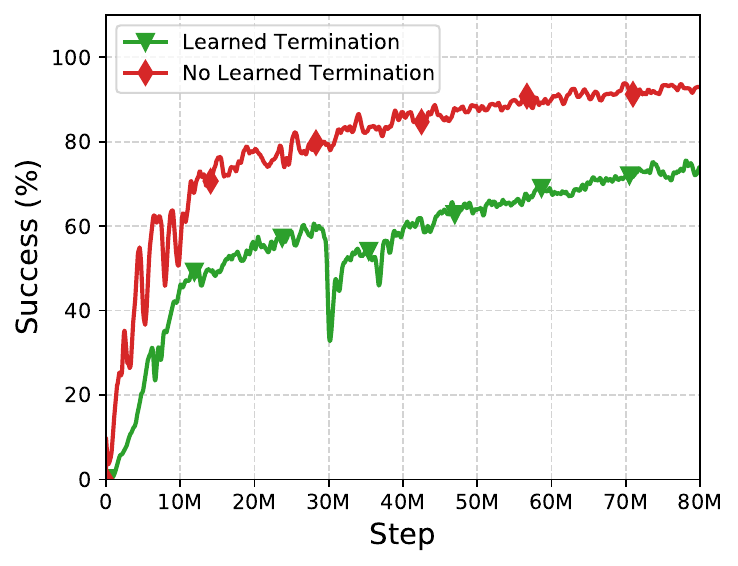}
    \caption{
      Training learning curve for the Navigation skill with and without the learned termination skill for 1 seed. 
    }
    \label{supp:fig:nav_skill}
\end{figure}

Since we found the Navigation skill difficult to train, we separately show the learning curves for the Navigation skill in \reffig{supp:fig:nav_skill}.
There we highlight the difficulty of learning the termination action by comparing to with and without the learned termination condition. 

\begin{figure}
    \centering
    \begin{subfigure}[t]{0.32\linewidth}
    	\includegraphics[width=\textwidth]{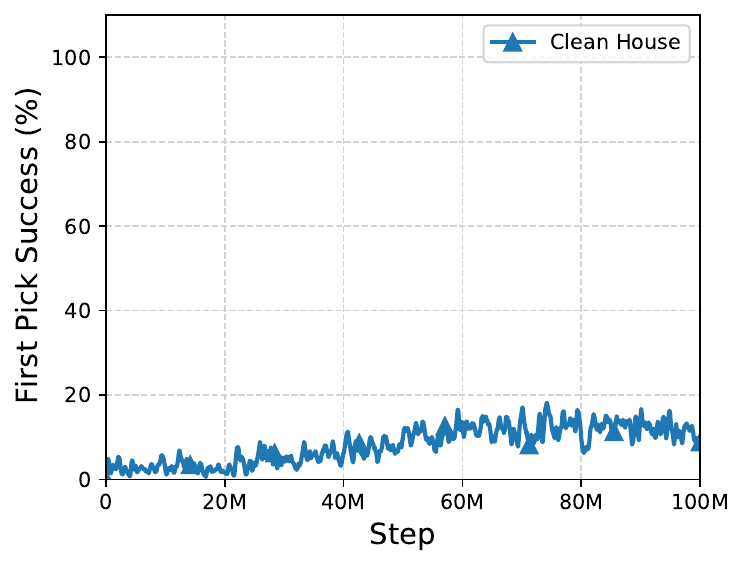}
      \caption{\cleanhouse}
    \end{subfigure} 
    \begin{subfigure}[t]{0.32\linewidth}
    	\includegraphics[width=\textwidth]{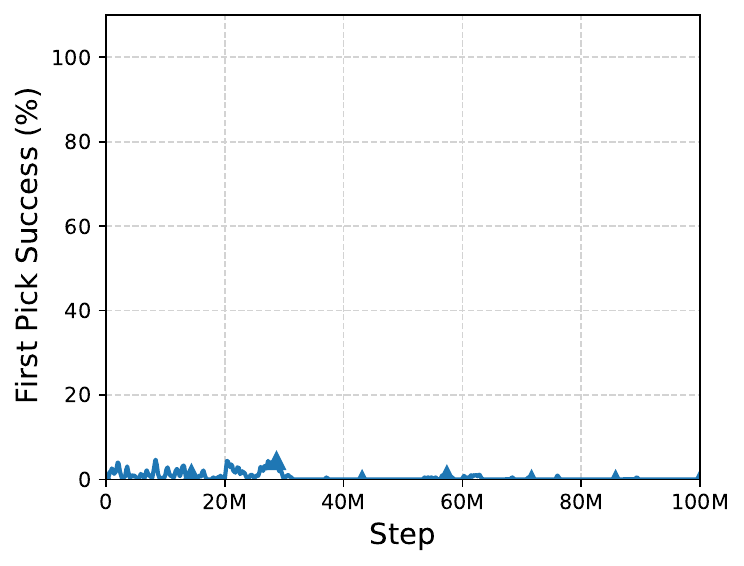}
      \caption{\stockfridge}
    \end{subfigure} 
    \begin{subfigure}[t]{0.32\linewidth}
    	\includegraphics[width=\textwidth]{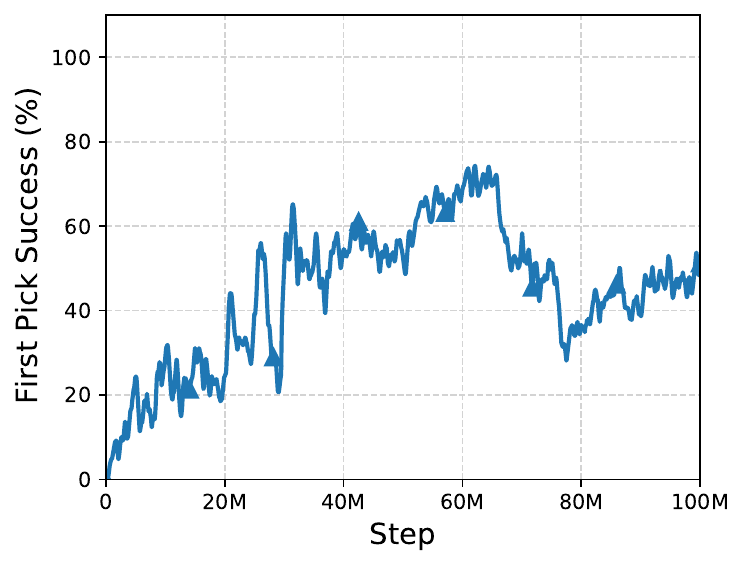}
      \caption{\settable}
    \end{subfigure} 
    \caption{
      Training curves for the \monolithic approach for all tasks for a single seed. 
      Y-axis shows success rates on picking the first object, in the case of \cleanhouse this requires navigating to and picking an object from an unobstructed random receptacle, for \stockfridge this is navigating to and picking an object from the fridge, and for \settable this is navigating to the drawer, opening it and then picking the object inside. 
    }
    \label{supp:fig:mono_learn}
\end{figure}

Likewise, we show the learning curves for the \monolithic approaches in \reffig{supp:fig:mono_learn}.
The success rate for picking the first object in \settable is higher than \cleanhouse since the object always starts in the same drawer for \settable. 
Likewise, \settable requires picking objects from an open drawer whereas \stockfridge requires picking objects from a tight fridge space.

\end{document}